%% file: main_arxiv.tex
\documentclass[lettersize,journal]{IEEEtran}

\usepackage{subcaption}

\usepackage{amsmath,amsfonts}
\usepackage{algorithmic}
\usepackage{algorithm}
\usepackage{array}
\usepackage{textcomp}
\usepackage{stfloats}
\usepackage{url}
\usepackage{verbatim}
\usepackage{graphicx}
\usepackage{cite}
\input{math_commands.tex}

\usepackage{microtype}
\usepackage{graphicx}
\usepackage{booktabs} 

\usepackage{hyperref}

\usepackage{amsmath}
\usepackage{amssymb}
\usepackage{mathtools}
\usepackage{amsthm}

\usepackage[capitalize,noabbrev]{cleveref}

\usepackage{booktabs}       
\usepackage{amsfonts}       
\usepackage{nicefrac}       
\usepackage{microtype}      
\usepackage{xcolor}         
\usepackage{array}

\usepackage{amsmath}
\usepackage{algorithm}
\usepackage{algorithmic}
\usepackage{booktabs}
\usepackage{newfloat}
\usepackage{listings}
\usepackage{mathtools}
\usepackage{multirow}
\usepackage{bbm}
\usepackage{amsfonts,amssymb}
\usepackage{enumitem}
\usepackage{subcaption}
\usepackage[T1]{fontenc}
\usepackage{cleveref}
\usepackage{cancel}

\usepackage{wrapfig}
\usepackage{lineno}
\usepackage{colortbl}
\usepackage{tcolorbox}

\newtcbox{\mybox}[1][red]{on line,
  arc=0pt, outer arc=0pt, 
  colback=#1!30!white,    
  colframe=#1!30!white,   
  boxrule=0pt,            
  boxsep=0pt, left=1pt, right=1pt, top=2pt, bottom=2pt 
}

\newcommand{\coloredline}[2]{\begin{tcolorbox}[colback=#1!30,colframe=#1!30,boxrule=0mm,arc=0mm,auto outer arc,left=0pt,right=0pt,top=0pt,bottom=0pt,boxsep=0pt]
#2
\end{tcolorbox}
}
\theoremstyle{plain}
\newtheorem{theorem}{Theorem}[section]

\theoremstyle{definition}
\newtheorem{definition}[theorem]{Definition}

\newtheorem{hypothesis}[theorem]{Hypothesis}
\theoremstyle{remark}
\newtheorem{remark}[theorem]{Remark}

\definecolor{darkgrey}{rgb}{0.53,0.53,0.53}
\definecolor{middlegrey}{rgb}{0.65,0.65,0.65}
\definecolor{mygrey}{rgb}{0.9,0.9,0.9}
\definecolor{mydarkblue}{rgb}{0,0.08,0.45}
\definecolor{darkdarkblue}{rgb}{0.0,0.0,0.3}
\definecolor{darkblue}{rgb}{0.0,0.0,0.8}
\definecolor{darkred}{rgb}{0.85, 0.1, 0.1}

\definecolor{coral}{rgb}{1.0, 0.5, 0.3}
\definecolor{tomato}{rgb}{1.0, 0.39, 0.28}
\definecolor{orangered}{rgb}{1.0, 0.27, 0}
\definecolor{gold}{rgb}{1.0, 0.84, 0}
\definecolor{orange}{rgb}{1.0, 0.64, 0}
\definecolor{darkorange}{rgb}{1.0, 0.55, 0}
\definecolor{someorange}{rgb}{0.9, 0.35, 0.17}

\definecolor{green}{rgb}{0, 0.5, 0}
\definecolor{limegreen}{rgb}{0.15, 0.65, 0.15}
\definecolor{forestgreen}{rgb}{0.13, 0.54, 0.13}
\usepackage{tikz}

\definecolor{lightblue}{RGB}{106, 159, 181}  
\definecolor{lightgreen}{RGB}{102, 205, 102} 
\definecolor{lightcyan}{RGB}{102, 204, 204}  

\begin{document}

\title{Improving Model Fusion by Training-time Neuron Alignment with Fixed Neuron Anchors}


\author{Zexi~Li,
        Zhiqi~Li,
        Jie~Lin,
        Tao~Shen,
        Jun~Xiao,
        Yike~Guo,~\IEEEmembership{Fellow,~IEEE},
        Tao~Lin,
        and Chao~Wu
\IEEEcompsocitemizethanks{\IEEEcompsocthanksitem Zexi Li, Jie Lin, Tao Shen, Jun Xiao, and Chao Wu are with Zhejiang University, Hangzhou, China.
E-mail: \{zexi.li, lj7674, tao.shen, junx, chao.wu\}@zju.edu.cn. 
\IEEEcompsocthanksitem Zhiqi Li is with Georgia Institute of Technology, Atlanta, USA. E-mail: zli3167@gatech.edu.
\IEEEcompsocthanksitem Tao Lin is with Westlake University, Hangzhou, China. E-mail: lintao@westlake.edu.cn.
\IEEEcompsocthanksitem Yike Guo is with Hong Kong University of Science and Technology, Hong Kong SAR, China. E-mail: yikeguo@ust.hk.
\IEEEcompsocthanksitem Corresponding Author: Chao Wu, Tao Lin, and Yike Guo.}
}

\markboth{IEEE Transactions on Pattern Analysis and Machine Intelligence}%
{Shell \MakeLowercase{\textit{et al.}}: A Sample Article Using IEEEtran.cls for IEEE Journals}


\maketitle

\begin{abstract}
Model fusion aims to integrate several deep neural network (DNN) models' knowledge into one by fusing parameters, and it has promising applications, such as improving the generalization of foundation models and parameter averaging in federated learning. However, models under different settings (data, hyperparameter, etc.) have diverse neuron permutations; in other words, from the perspective of loss landscape, they reside in different loss basins, thus hindering model fusion performances. To alleviate this issue, previous studies highlighted the role of permutation invariance and have developed methods to find correct network permutations for neuron alignment after training. 
Orthogonal to previous attempts, this paper studies training-time neuron alignment, improving model fusion without the need for post-matching. Training-time alignment is cheaper than post-alignment and is applicable in various model fusion scenarios.
Starting from fundamental hypotheses and theorems, a simple yet lossless algorithm called TNA-PFN is introduced. TNA-PFN utilizes partially fixed neuron weights as anchors to reduce the potential of training-time permutations, and it is empirically validated in reducing the barriers of linear mode connectivity and multi-model fusion. It is also validated that TNA-PFN can improve the fusion of pretrained models under the setting of model soup (vision transformers) and ColD fusion (pretrained language models). Based on TNA-PFN, two federated learning methods, FedPFN and FedPNU, are proposed, showing the prospects of training-time neuron alignment. FedPFN and FedPNU reach state-of-the-art performances in federated learning under heterogeneous settings and can be compatible with the server-side algorithm. 
\end{abstract}


\section{Introduction}\label{sec:intro}
Deep neural networks (DNNs) have shown great powers in various machine learning tasks; for instance, large language models~\cite{gpt-2,gpt-3,llama-2}, represented by GPT-4~\cite{openai2024gpt4}, represent human-level intelligence in question answering, and diffusion models can generate images or videos that cannot be distinguished from reality~\cite{dalle_2,stable_diffusion,sora}. Towards more powerful models, model fusion~\cite{fusionbench,deepfusion_survey} aims to integrate the knowledge and powers of several DNNs into one model by fusing the model parameters, and it has wide and promising applications~\cite{mergekit,cold_fusion,wortsman2022model}. One line of research finds that model fusion in pretrained-finetuned paradigm can improve generalization, such as model soup~\cite{wortsman2022model} and ColD fusion~\cite{cold_fusion}. Model soup~\cite{wortsman2022model}, an application of model fusion, effectively enhances generalization performance by fusing the weights of multiple fine-tuned models, which significantly improves prediction accuracy across various tasks without increasing computational demands during inference. ColD fusion~\cite{cold_fusion} finds that model fusion of fine-tuned language models can be recycled to continually improve the pretrained model they are based upon. In addition, federated learning~\cite{mcmahan2017communication,wang2020federated,pmlr-v202-li23s} adopts weighted model fusion on the server to generate global models by integrating the knowledge of local data in a privacy-preserving and communication-efficient manner.

\begin{figure}[t]
  \centering
    \includegraphics[width=0.99\linewidth]{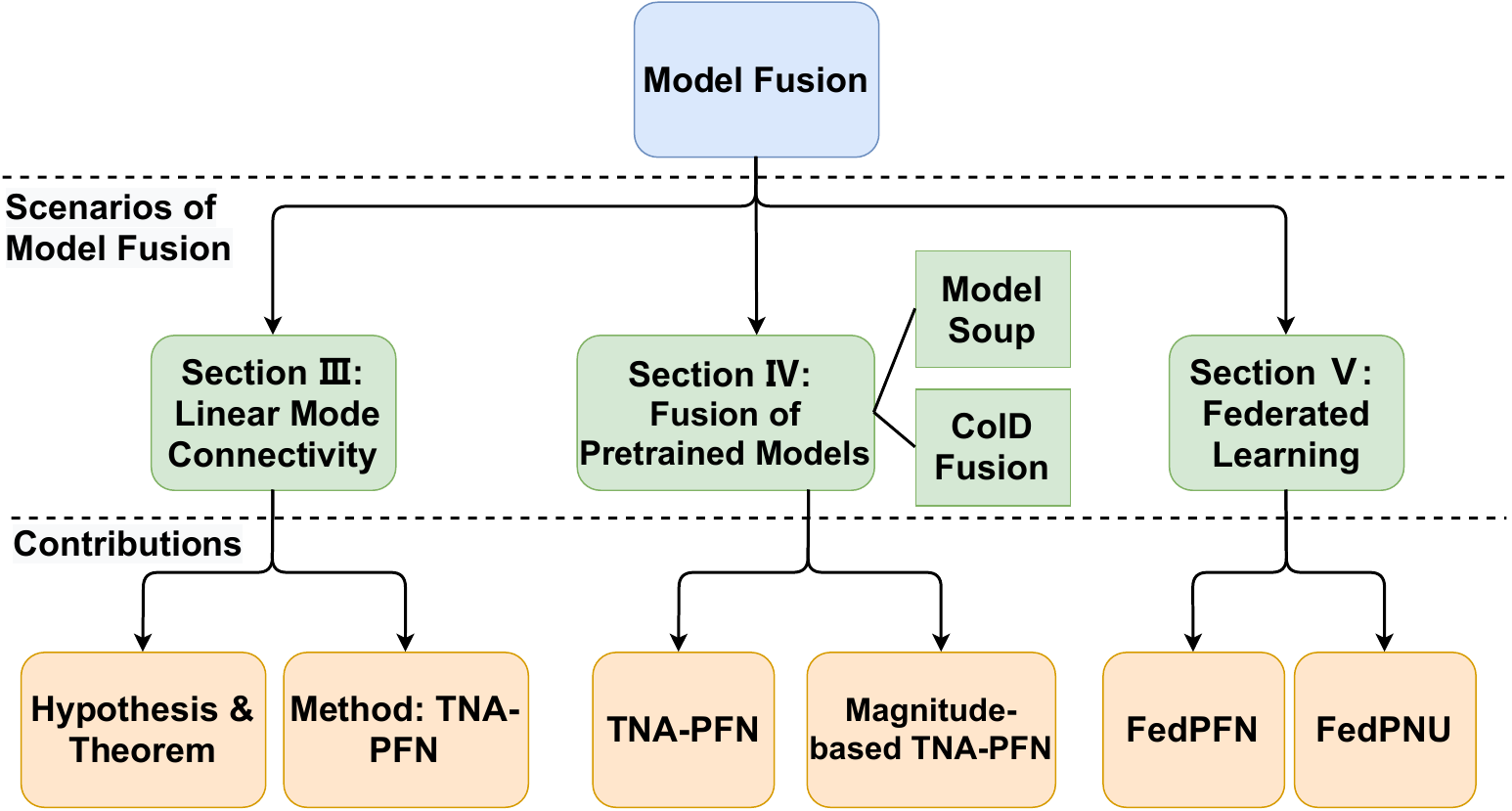}
    \vspace{-0.2cm}
    \caption{\textbf{Organization overview of the paper.}}\label{fig:paper_overview}
\end{figure}

Despite the prospects of model fusion, there exist barriers when directly fusing model parameters due to the properties of DNNs. Linear mode connectivity (LMC) studies the fundamental aspects of connectivity in loss landscape between different stochastic gradient descent (SGD) solutions~\cite{frankle2020linear,draxler2018essentially,entezari2021role,ainsworth2022git}, serving as basic theories and foundations for model fusion. The word "connectivity" refers to the landscape of linearly fusing two models as $\mathbf{w} = \alpha \mathbf{w}_1 + (1 - \alpha)\mathbf{w}_2, \text{s.t.}~ \alpha \in [0, 1]$. It is found that even if two trained SGD solutions have the same initialization and trainset but have different SGD random seeds (i.e., batch orders), there will be a loss barrier in the LMC~\cite{entezari2021role,ainsworth2022git}, not to mention in realistic model fusion cases, models are trained on disjoint and heterogeneous/multi-task datasets~\cite{wortsman2022model,ties_merge}. Recent studies find that the barrier in LMC is mainly due to permutation symmetry (also known as permutation invariance) properties of DNNs~\cite{entezari2021role}. Because of permutation symmetry, neurons between models are not aligned, and it is suggested that the function of the network can remain the same while changing the permutations of neurons, which can result in many functionally same but geometrically different solutions. From the perspective of the loss landscape, if neurons are not aligned, different solutions reside in distinct loss basins~\cite{ainsworth2022git}; directly averaging these models' parameters will result in a high-loss plateau with poor generalization. 

Previous works in LMC try to find the right permutations for post-training neuron alignment. In Entezari et al.~\cite{entezari2021role}, it is conjectured that if taking all permutations into account, all SGD solutions can be mapped into the same loss basin where no barrier in LMC. Git Re-Basin~\cite{ainsworth2022git} further validates this conjecture by proposing three algorithms to find such permutations in a more efficient manner. However, post-training matching has limitations, especially for model fusion with large models. First is the computation complexity that post-hoc neuron alignment is a hard combinatorial optimization problem. As stated in Ainsworth et al. \cite{ainsworth2022git}, even for a three-layer MLP with 512 widths, the number of permutation symmetries is nearly 10\textasciicircum3498. When comes to larger foundation models, especially vision transformers~\cite{vit_first_paper} and large language models~\cite{gpt_first_paper}, finding such an appropriate permutation to align neurons is more challenging and expensive. 
For the scenarios where alignment and fusion among multiple models are needed, especially federated learning~\cite{wang2020federated,mcmahan2017communication}, the cost of the post-hoc alignment methods increases as the number of models. 
Additionally, post-hoc matching requires tailored designs for different modules of DNNs, and one method cannot be applied to all architectures. The early methods, such as simulated annealing~\cite{entezari2021role}, git re-basin~\cite{ainsworth2022git}, and OTFusion~\cite{singh2020model}, can be applied to MLP and convolutional layers but are not applicable to attention layers and layer normalizations in the transformer blocks. Though new methods are proposed for transformer fusion~\cite{imfeld2023transformer} later on, whether they are applicable to more new architectures (KAN~\cite{kan}, Mamba~\cite{mamba}, etc.) still remains questioned. 

This paper, instead of post-training neuron alignment, explores whether training-time neuron alignment can be achieved in an efficient and principal way, which can improve model fusion across various scenarios and model architectures. Also, training-time alignment is orthogonal to post-hoc methods, so it can reduce the cost of post-hoc matching or further improve post-matching if jointly used. In addition, studying training-time neuron alignment can help to understand the behind mechanisms of model parameters and training dynamics, serving as foundations for future research of model fusion~\cite{fusionbench} and model editing~\cite{knowledge_editing_survey_1}.

Towards training-time neuron alignment, in this paper, it is hypothesized that the key is to \textbf{break}\textit{ permutation \textbf{symmetry}} via \textit{permutation \textbf{asymmetry} subspace}. During training, if the potential permutation symmetries are reduced, the neurons will be more aligned in a subspace so that the LMC barriers will decrease and model fusion will improve. It is found that pruning at initialization supports the hypothesis, but pruning will impair individual model performances. Instead of pruning, this paper introduces TNA-PFN, a simple yet lossless algorithm for training-time alignment. TNA-PFN partially fixes some neuron weights as anchors for reducing the potential of permutation symmetries and training in a subspace. Due to the overparameterization property of DNNs, the neurons are so redundant that fixing a proportion of weights may not impair training performances and may preserve the learned features/knowledge. The hypothesis and TNA-PFN are first proposed and verified in LMC with both theoretical and empirical support since LMC is the simplest form and basic foundation of model fusion. Then, TNA-PFN and the methods derived from TNA-PFN are validated under various model fusion scenarios, including model soup \cite{wortsman2022model} for vision transformers, ColD fusion \cite{cold_fusion} for pretrained language models, and federated learning under heterogeneous datasets~\cite{Li_2023_ICCV,pmlr-v202-li23s,acar2020federated}. In a nutshell, this paper has the following contributions. 

\begin{itemize}
    \item We discover the neuron alignment problem from the perspective of training time, which provides new insights. We hypothesize that learning in permutation subspaces can reach better LMC and model fusion.
    \item Under the hypothesis, we first find pruning at initialization can improve LMC. Then, we propose a simple yet more lossless training-alignment method---TNA-PFN, which is validated under both theoretical and empirical analysis. 
    \item TNA-PFN excels in wide model fusion applications. TNA-PFN can boost model soup and ColD fusion for free, showcasing its prospects for pretrained foundation models. Also, we extend TNA-PFN in federated learning and devise two algorithms, FedPFN and FedPNU, both of which are validated effective for improving the global model's generalization under extensive experiments.\looseness=-1
\end{itemize}

We study training-time neuron alignment under comprehensive scenarios of model fusion, and the main organization of the paper is in \autoref{fig:paper_overview}. First, in \autoref{sec:lmc}, training-time neuron alignment is studied in linear mode connectivity, where hypothesis and preliminary findings are shown. TNA-PFN is proposed to reduce the barriers in LMC and is also verified by theoretical insights. Second, in \autoref{sec:pretrained_fusion}, TNA-PFN is extended into practice---model fusion under the pretrained-finetuned paradigm, where two settings are considered: model soup of vision transformers and ColD fusion of pretrained language models. Third, in \autoref{sec:TNA4FL}, two federated learning algorithms are proposed, which are derived from TNA-PFN. It is validated that the two algorithms can reduce the model drifts caused by data heterogeneity and improve the generalization of global models under extensive settings, such as different datasets, data heterogeneity, local epochs, and client numbers.

\section{Related Works} \label{sec:related_works}
\subsection{Model Fusion}
Model fusion is an emerging technique in deep learning that unifies the knowledge of several DNNs into one single model in a cost-effective and data-efficient manner~\cite{fusionbench,deepfusion_survey}. Broadly speaking, model fusion includes model ensemble~\cite{fusionbench,deepfusion_survey}, which only ensembles the predictions of models instead of parameters~\cite{garipov2018loss,benton2021loss}, while in this paper, we focus on a more focused definition of model fusion that studies the fusion of model parameters~\cite{singh2020model,imfeld2023transformer}, also known as model merging~\cite{mergekit}. Linear mode connectivity studies the fusion of two SGD solutions (i.e., the modes)~\cite{frankle2020linear,draxler2018essentially}, which provides theoretical foundations of model fusion. Besides, in federated learning, local models are trained on heterogeneous data, and the server generates a global model in each communication round by fusion of local models' parameters, where the original model fusion algorithm is called FedAvg~\cite{mcmahan2017communication}. Recently, model fusion techniques have been proposed to improve foundation models under a pretrained-finetuned paradigm. The notion of task arithmetic aims to use model fusion as a technique of model editing by adding or removing task-specific model weight vectors~\cite{ilharco2023editing,ortiz2023task}. In addition, this paper focuses on two scenarios of model fusion under the pretrained-finetuned paradigm, namely, model soup and ColD fusion. Model soup~\cite{wortsman2022model} enhances the generalization performance of foundation models by fusing the weights of multiple fine-tuned models, which significantly improves prediction accuracy across various tasks without increasing computational demands during inference. ColD fusion~\cite{cold_fusion} finds that model fusion of fine-tuned language models can be recycled to continually improve the pretrained model they are based upon. 

\subsection{Linear Mode Connectivity}
Linear mode connectivity refers to the phenomenon that there exists a loss (energy) barrier along the linear interpolation path of two networks, in the cases where i) the two networks have the same initialization and are trained on the same dataset but with different random seeds (data shuffles and augmentations)~\cite{ainsworth2022git}; ii) the two networks are with different initializations but are trained on the same dataset~\cite{entezari2021role}; iii) the two networks are the initial network and the final trained network~\cite{vlaar2022can}. Specifically, Adilova et al.~\cite{adilova2023layerwise} examines the linear mode connectivity of different layers. Valaar et al.~\cite{vlaar2022can} studies the relationship between generalization and the initial-to-final linear mode connectivity. Frankle et al.~\cite{frankle2020linear} connects linear mode connectivity with the lottery ticket hypothesis and finds better connectivity can result in better pruning performances. Zhao et al.~\cite{zhao2020bridging} bridges mode connectivity and adversarial robustness. Some works try to extend mode connectivity beyond ``linear'', e.g., searching for a non-linear low-loss path~\cite{draxler2018essentially} or studying mode connectivity under spurious attributes~\cite{lubana2023mechanistic}.\looseness=-1

Permutation invariance (a.k.a. permutation symmetry) refers to the property of neural networks that the positions of neurons can be permuted without changing its function~\cite{simsek2021geometry,hecht1990algebraic}, and it is believed to be the primary cause of loss barrier in linear mode connectivity~\cite{entezari2021role,ainsworth2022git}. \cite{entezari2021role} hypothesizes that if taking the permutation invariance into consideration, all solutions can be mapped into the same low-loss basin with connectivity. Further, Ainsworth~et al.~\cite{ainsworth2022git} validates this hypothesis by using ``re-basin'' which aims to find the appropriate permutation matrices to map the networks into the same basin. Other methods are also utilized to match the neurons for better model fusion, such as optimal transport~\cite{singh2020model}, Bayesian nonparametric technique~\cite{yurochkin2019bayesian,wang2019federated}, Hungarian algorithm~\cite{tatro2020optimizing}, graph matching~\cite{liu2022deep}, and implicit Sinkhorn differentiation~\cite{pena2023re}. We note that all these methods are for post-matching after training, while we focus on training-time neuron alignment. \looseness=-1

\subsection{Model Fusion in Federated Learning}
Federated learning (FL) is a collaborative training paradigm in which several clients collaboratively train a global model without sharing their data in a communication-efficient and privacy-preserving way~\cite{mcmahan2017communication}. In FL, a central server, as the role of coordinator, collects clients' local models and generates the global model via model fusion. FedAvg~\cite{mcmahan2017communication} is the typical model fusion method in FL, where the fusion weights of local models are set by the sizes of local datasets. In FedLAW~\cite{pmlr-v202-li23s} and FedDisco~\cite{ye2023feddisco}, weighted fusion is improved by seeking data-heterogeneity-aware fusion weights or setting the norm of weights smaller than 1 for global weight decay regularization. 

In FL, clients have heterogeneous data and conduct several local epochs of training before model fusion, causing model drift of local models. To tackle model drift and improve model fusion, some server-side methods have been proposed in FL. FedDF~\cite{lin2020ensemble} uses ensemble distillation on the server to improve the model fusion of FedAvg. PFNM~\cite{yurochkin2019bayesian} and FedMA~\cite{wang2020federated} use Bayesian nonparametric methods for neuron alignment of local models before model fusion. While this paper focuses on training-time alignment methods that are orthogonal to the server-side model fusion techniques. 
Note that the previous work of PAN~\cite{li2022federated} shares a similar motivation with this paper's FL methods. PAN uses position-aware encoding on the data input for aligning intermediate activations, which is orthogonal to the neuron weight anchors in this paper. In addition, PAN only focuses on FL, while this paper studies wide applications of model fusion, such as LMC, model soup, and FL.

\subsection{Works about the Relationship between Pruning and LMC}
In Frankle et al.~\cite{frankle2020linear}, the authors use linear mode connectivity to study the performances of lottery-ticket-hypothesis-based pruning and find that the sparse pruned model with good connectivity will be more likely to reach the full accuracy after pruning. While in this paper, it is found that random pruning (not necessarily lottery tickets) can improve linear mode connectivity. Though the two papers both discuss the relationship between pruning and linear mode connectivity, they have different focuses and contributions: Frankle et al.~\cite{frankle2020linear} finds LMC indicates better results of pruning, whereas we find pruning can improve LMC, and the causal logic is different. It is notable that some concurrent works~\cite{mcdermott2023linear,choisparse} also find that pruning can improve LMC, but according to the dates of release, this paper may be the first to present the finding, and it is non-trivial.

\begin{figure*}[t]
  \centering
    \includegraphics[width=0.45\linewidth]{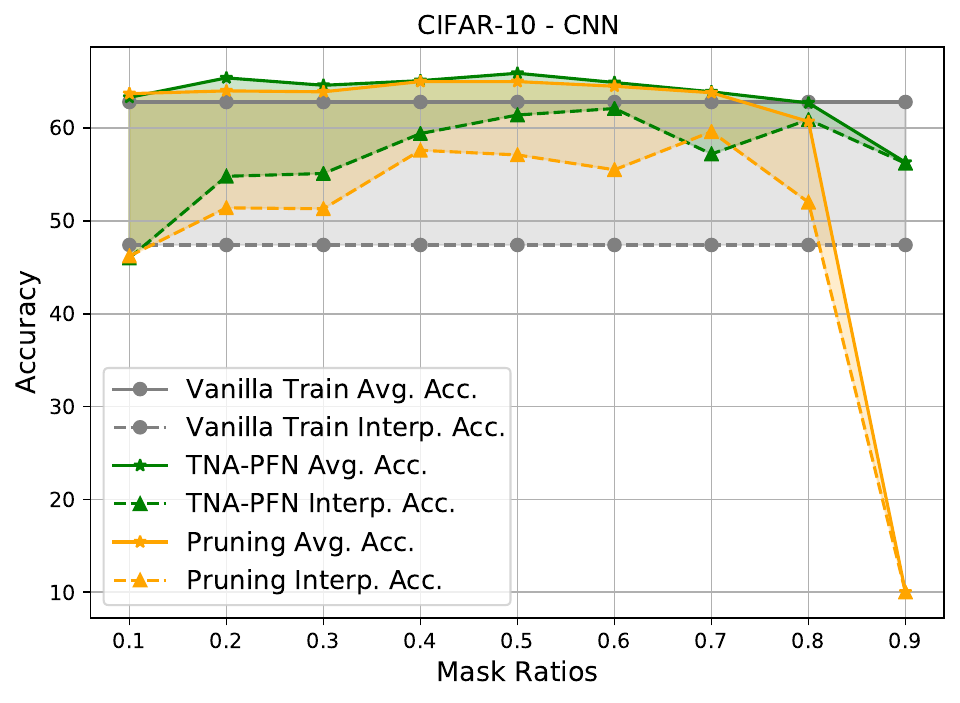}
    \hspace{0.6cm}
    \includegraphics[width=0.45\linewidth]{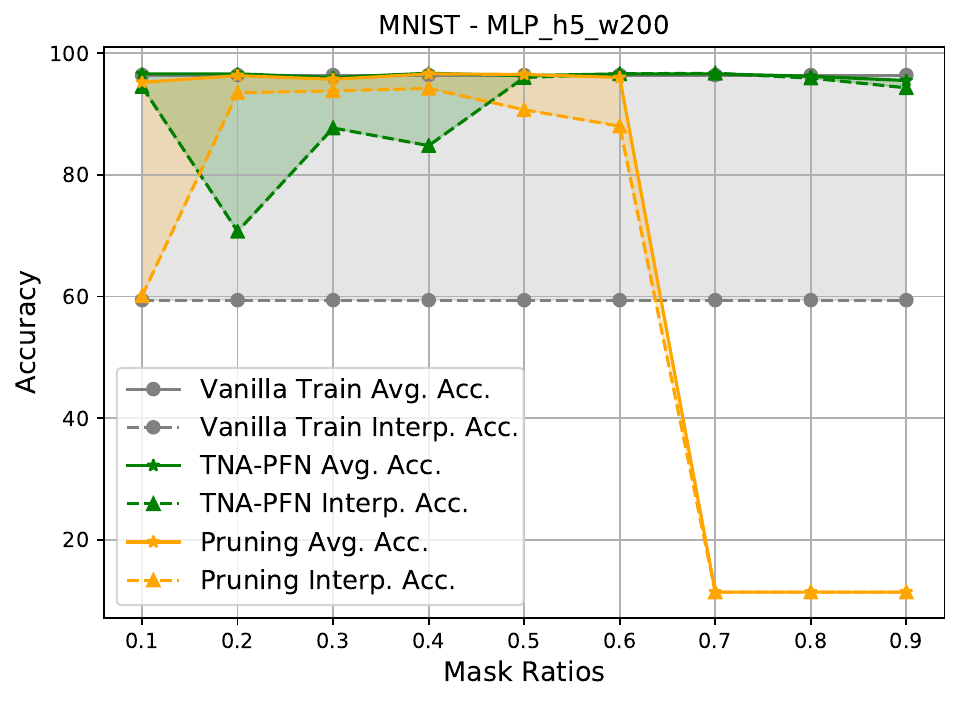}
    \vspace{-0.1cm}
    \caption{\textbf{LMC of random pruning at initialization and TNA-PFN under different mask ratios.} For pruning, the mask ratio is the pruning ratio. ``Avg. Acc.'' means averaged accuracies of individual models, and ``Interp. Acc.'' means the accuracy of the interpolated model ($\alpha=0.5$) of two modes. The shadow areas mean the accuracy barriers in LMC, the smaller the better.}
    \label{fig:lmc:pruning_mask_ratios}
\end{figure*}

\section{Hypothesis, Theory, and Empirical Finding of Training-time Neuron Alignment: a Preliminary Study in Linear Mode Connectivity}\label{sec:lmc}
 As LMC provides basic theories and foundations of model fusion, this section gives a preliminary study about training-time neuron alignment in LMC. 

\subsection{Preliminary of Linear Mode Connectivity}
This paper considers the linear mode connectivity of two SGD solutions, which have the same initialization but different data orders\footnote{It is noted that there are other forms of LMC, such as the LMC from the initialization and the trained model~\cite{vlaar2022can}, and the LMC between two models with different initializations~\cite{entezari2021role}. In this paper, LMC servers are preliminaries for model fusion with applications such as federated learning, where different models have the same initialization, so the LMC cases with the same initialization are considered.}. The definitions of loss barrier and accuracy barrier are listed below.\looseness=-1

\begin{definition}\label{def:loss_barrier}
    (\textbf{Loss barrier}~\cite{entezari2021role}) 
    Let $f_\mathbf{w}(\cdot)$ be a function represented by a neural network with parameter vector $\mathbf{w}$ that includes all parameters and $\mathcal{L}(\mathbf{w})$ be the any given loss (e.g., train or test error) of $f_\mathbf{w}(\cdot)$. Given two independently trained networks $\mathbf{w}_1$ and $\mathbf{w}_2$, let $\mathcal{L}(\alpha \mathbf{w}_1 + (1 - \alpha)\mathbf{w}_2)$, for $\alpha \in [0, 1]$ be the loss of the linearly interpolated network. 
The loss barrier $B_{loss}(\mathbf{w}_1, \mathbf{w}_2)$ along the linear path between $\mathbf{w}_1$ and $\mathbf{w}_2$ is defined as the highest difference between the loss of the interpolated network and linear interpolation of the loss values of the two networks:\looseness=-1
\begin{align} \label{eqn:loss_barrier}
  B_{loss}(\mathbf{w}_1, \mathbf{w}_2) =&
  \sup_\alpha \left\{[\mathcal{L}(\alpha \mathbf{w}_1 + (1 - \alpha)\mathbf{w}_2)]\right.\notag \\ &\left.-[ \alpha\mathcal{L}(\mathbf{w}_1)+(1 - \alpha) \mathcal{L}(\mathbf{w}_2)]\right\}.
\end{align}
\end{definition}

The loss barrier of the above definition is not bounded. To better depict and compare the barrier changes, a definition of the accuracy barrier, which is bounded within $[0, 1]$, is given below. \looseness=-1

\begin{definition} \label{def:acc_barrier}
    (\textbf{Accuracy barrier}) 
    Let $\mathcal{A}(\mathbf{w})$ be the accuracy (e.g., train or test accuracy) of $f_\mathbf{w}(\cdot)$. Let $\mathcal{A}(\alpha \mathbf{w}_1 + (1 - \alpha)\mathbf{w}_2)$, for $\alpha \in [0, 1]$ be the accuracy of the linearly interpolated network. 
The accuracy barrier $B_{acc}(\mathbf{w}_1, \mathbf{w}_2)$ along the linear path between $\mathbf{w}_1$ and $\mathbf{w}_2$ is defined as the highest ratio of the interpolated network's accuracy drop to the averaged accuracy:

\small{
\begin{align} \label{eqn:acc_barrier}
  B_{acc}(\mathbf{w}_1, \mathbf{w}_2) = 
  \sup_\alpha \left[1 - \frac{\mathcal{A}(\alpha \mathbf{w}_1 + (1 - \alpha)\mathbf{w}_2)}{\alpha\mathcal{A}(\mathbf{w}_1)+(1 - \alpha) \mathcal{A}(\mathbf{w}_2)}\right].
\end{align}
}
\end{definition}

The above definition maps the barrier into $[0, 1]$. If the accuracy barrier is 0, it means no barrier exists along the linear interpolation path; else if the barrier is nearly 1, it means the generalization of the interpolated model is nearly zero, and its prediction is no better than random guessing.

\textit{Permutation invariance.} Permutation invariance refers to the property that the positions (i.e., permutations) of neurons of a given network can be changed without changing the network's function, and it is also known as permutation symmetry~\cite{ainsworth2022git}. We take a multi-layer MLP as an example to demonstrate the property. \looseness=-1

Assume an MLP network has $L+1$ layers (containing input and output layer), and each layer contains $J_{l}$ neurons, where $l \in \{0,1,\cdots,L\}$ is the layer index. $J_0$ and $J_L$ are input and output dimensions. We denote the parameters of each layer as the weight matrix $\mathbf{W}_{l} \in \mathbb{R}^{J_{l} \times J_{l-1}}$ and the bias vector $\mathbf{b}_{l} \in \mathbb{R}^{J_{l}}$, $l\in \{1,2,\cdots,L\}$. 
The input layer does not have parameters. We use $\mathbf{h}_l \in \mathbb{R}^{J_l}$ as the outputs of the $l$-th layer. We have $\mathbf{h}_{l}=\sigma_{l}(\mathbf{W}_l\mathbf{h}_{l-1} + \mathbf{b}_l)$, where $\sigma_l(\cdot)$ is the element-wise activation function, e.g., ReLU. We use $\mathbf{\Pi} \in \{0,1\}^{J\times J}$ as a permutation matrix that satisfies $\sum_{j}\mathbf{\Pi}_{\cdot,j}=1$ and $\sum_{j}\mathbf{\Pi}_{j,\cdot}=1$. By applying the permutation matrices to the layers, the network function remains unchanged. For the $l$-th layer, the layer-wise permutation process is
\begin{equation}
    \mathbf{h}_l = \sigma_l(\mathbf{\Pi}_l \mathbf{W}_l \mathbf{\Pi}_{l-1}^T \mathbf{h}_{l-1} + \mathbf{\Pi}_l \mathbf{b}_l), \label{eq:permute}
\end{equation}
where $\mathbf{\Pi}_0=\mathbf{I}$ and $\mathbf{\Pi}_L=\mathbf{I}$, meaning that the input and output are not shuffled. We note that the permutation matrices have the following properties: 
\begin{align}\label{equ:perm_property}
    \mathbf{\Pi}^T\mathbf{\Pi}=\mathbf{I}, \mathbf{\Pi} \mathbf{a} + \mathbf{\Pi} \mathbf{b} = \mathbf{\Pi}(\mathbf{a} + \mathbf{b}),\notag\\ \mathbf{\Pi} \mathbf{a} \odot \mathbf{\Pi} \mathbf{a} = \mathbf{\Pi} (\mathbf{a} \odot \mathbf{b}), \sigma(\mathbf{\Pi} \vecx) = \mathbf{\Pi} \sigma(\vecx),
\end{align}
where $\mathbf{I}$ is the identity matrix, $\odot$ denotes Hadamard product, and $\sigma(\cdot)$ is an element-wise function.

\subsection{Hypothesis and Preliminary Finding}
Previous works find that permutation symmetry is the main cause of LMC barriers. Due to the numerous parameters and permutation symmetries, SGD will find solutions far from each other in the landscapes during training. Therefore, we think a possible solution for training-time neuron alignment is breaking permutation symmetries in training via the same asymmetric subspace across models, so we make the following hypothesis.
\begin{hypothesis}[Informal]\label{subspace_hypothesis}
    If we can reduce the potential number of permutation symmetries by learning different models in the same permutation subspace, the linear mode connectivity will be improved.
\end{hypothesis}

\textit{Network pruning improves LMC.} One straight-forward method which fits Hypothesis~\ref{subspace_hypothesis} is network pruning at initialization. We apply random weight pruning to an initialized model, and different neurons will have different pruning structures (i.e., pruning ratios and pruning positions), so the symmetric structure of the network is broken, and the permutations of neurons are limited. From this pruned initialization, different training runs will learn in the same subspace where pruned weights are always zeros and permutations of neurons are constrained. 

It is validated in \autoref{fig:lmc:pruning_mask_ratios} that pruning can actually improve LMC compared with vanilla training, which supports our hypothesis. But when the pruning ratio $\rho$ (the proportion of eliminated neuron weights) is high (i.e., 0.8 and 0.9), pruning will result in an untrainable network with nearly zero generalization. It is more severe when the initialized model is pretrained transformers, from \autoref{fig:model_soup_fig}, it can be seen that even if the pruning ratio is mild ($\rho = 0.2$), random pruning is fatal to ViT's network generalization. Thus, to overcome pruning's drawbacks, we propose a simple yet lossless (in terms of generalization) training-time alignment method in \autoref{subsec:tna_pfn}.

\subsection{Training-time Neuron Alignment with Partially Fixed Neurons as Anchors}\label{subsec:tna_pfn}

\subsubsection{Method Formulation} \label{subsec:PFN_formulation}
\begin{figure}[t]
  \centering
    \includegraphics[width=0.99\linewidth]{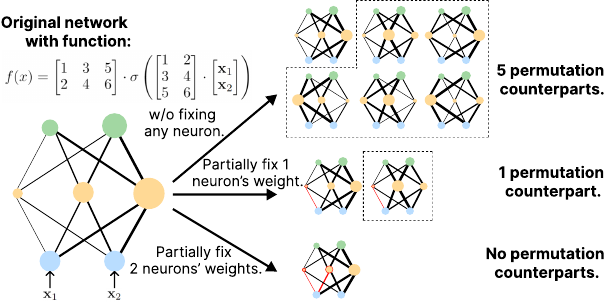}
    \vspace{-0.2cm}
    \caption{\textbf{A simple demonstration of reducing permutation symmetries via fixed neuron anchors.} There are 6 networks (1 original + 5 counterparts) which are functionally identical but with different permutations. The number of permutation symmetries can be reduced by asymmetrically fixing some weights (in \textcolor{red}{red}). Though this demonstration presents a simple static network permutation, we will show in the main paper that this kind of method can realize better neuron alignment under training dynamics.}\label{fig:PFN_demo}
\end{figure}

Pruning sets the fixed weights as zeros, and multiplying these zero weights will let intermediate features be zeros as $\mathbf{h}_{l} = \mathbf{W}_{l-1}\mathbf{h}_{l-1}$, missing important data features and losing expressiveness of DNNs if pruning is severe.
Instead of setting the weights as zeros like pruning, we propose to fix some neurons' weights, which will also break the network symmetry to reduce the permutations while preserving the expressiveness power of networks. An intuitive demonstration is in \autoref{fig:PFN_demo}, by fixing some weights of neurons, the number of potential permutations decreases. The fixed neuron weights can serve as anchors for reducing the potential number of network permutations. The proposed method is called \textbf{T}raining-time \textbf{N}euron \textbf{A}lignment with \textbf{P}artially \textbf{F}ixed \textbf{N}eurons as Anchors, dubbed as TNA-PFN. Due to the overparameterization and redundancy properties of DNNs, partially fixing some weights as TNA-PFN will not impede the optimization, instead, serving as regularization for better generalization.

We give the detailed implementation of TNA-PFN via unified gradient mask. To keep the weights frozen, we use fixed and unified gradient masks to keep certain neuron weights from updating. 
Specifically, given an initial network parameterized by a weight vector $\mathbf{w}^0\in \mathbb{R}^{d}$. 
For $\mathbf{w}^0$, we randomly generate a mask for each layer according to the mask ratio $\rho$ (refers to the proportion of zeros in the mask $\mathbf{m}^0$, i.e., the proportion of fixed weights), and the whole mask is $\mathbf{m}^0 \in \{0,~1\}^d$. In $\mathbf{m}^0$, $0$ for fixed and $1$ for updated, indicating the parameter update status. We individually train $n$ models with different batch orders or datasets. We set each model's initialization as $\mathbf{w}_{i} \leftarrow \mathbf{w}^{0}, i \in [n]$. 
Each model $\mathbf{w}_i, i \in [n]$ conducts the following updates in every SGD iteration using the identical mask $\mathbf{m}^0$ for neuron alignment:
\begin{equation} \label{equ:PFN_sgd}
    \mathbf{w}_{i} \leftarrow \mathbf{w}_{i}-\eta(\mathbf{m}^0\odot\mathbf{g}_i(\mathbf{w}_{i})),
\end{equation}
where $\odot$ denotes the element-wise Hadamard product, $\eta$ refers to the learning rate, and $\mathbf{g}_i$ is its gradients of the optimizer, such as SGD or Adam. After training for $E$ epochs, we validate the LMC with respect to the loss or accuracy barriers in Definitions \ref{def:loss_barrier} and \ref{def:acc_barrier}. The method is notated as TNA-PFN. 
\looseness=-1

We also present how TNA-PFN improves LMC under different $\rho$ in \autoref{fig:lmc:pruning_mask_ratios}, which shows TNA-PFN's effectiveness in reducing the barriers and its advantages over pruning. Additionally, a \emph{connectivity-accuracy tradeoff} is observed for both TNA-PFN and pruning that when the mask ratio is higher, the accuracy barriers diminish along with the decrease in the averaged accuracies of independently trained models. However, when $\rho$ is set appropriately (e.g., 0.4-0.6 for the CIFAR-10 and CNN setting), both the averaged accuracy and LMC can be improved. 

\textit{Discussions on differences between TNA-PFN and previous gradient-mask-based methods.} Applying gradient masks is discovered in previous gradient compression literature of distribution optimization, but our method is different from the previous works in the aspects as follows. 
\emph{i) Motivation difference:} Gradient compression is proposed for communication efficiency of distributed optimization while we study the training-time neuron alignment problem in LMC and model fusion. 
\emph{ii) Implementation difference:} Gradient compression uses different random top-k gradient masks at each worker and changes the mask per communication iteration~\cite{lin2018deep,vogels2019powersgd}; whereas, TNA-PFN uses the same random gradient masks at each model, fixes the mask, and independently trains the models without any communications; and FedPFN/FedPNU (presented in \autoref{sec:TNA4FL}) uses the same masks at each client's local training and changes the mask per global communication round. 
\emph{iii) Effect difference:} Since the masks of workers are different and changing, previous gradient compression methods cannot learn in a consistent subspace of parameters, while we learn in a subspace by the same gradient mask so that some neuron weights are not updated. 
\looseness=-1

\subsubsection{Theoretical Analysis}

We make a theoretical analysis about how TNA-PFN can improve LMC, shown in \autoref{thm:1} (proof is in the Appendix). The main idea is to treat the linear interpolated landscape of the barrier as a function of parameter $\alpha$, and the connectivity can be depicted by the first and second derivatives of the function.\looseness=-1

\begin{theorem} \label{thm:1}
We define a two-layer neural network with ReLU activation, and the function is $f_{\vv,\mU}(\vx)=\vv^\top\sigma(\mU\vx)$ where $\sigma(\cdot)$ is the ReLU activation function. $\vv\in \mathbb{R}^h$ and $\mU\in \sR^{h\times d}$ are parameters\footnote{For simplicity and without loss of generality, the bias terms are omitted.} and $\vx \in \sR^d$ is the input which is taken from $\sX=\{\vx\in \sR^d| \Vert\vx\Vert_2<b\}$ uniformly. Consider two different networks parameterized with $\{\mU,\vv\}$ and $\{\mU',\vv'\}$ respectively, and for arbitrarily chosen masks $\mM_\vv \in \{0,1\}^h$ and $\mM_{\mU}\in \{0,1\}^{h\times d}$, each element of $\mU$ and $\mU'$, $\vv$ and $\vv'$ is i.i.d. sampled from a sub-Gaussian distribution $\mathcal{\text{sub-G}}(0,\sigma_\mU^2)$ and $\mathcal{\text{sub-G}}(0,\sigma_\vv^2)$ respectively with setting $\evv_{i}=\evv'_{i}$ when $\emM_{\vv,i}=0$ and $\emU_{i,j}=\emU'_{i,j}$ when $\emM_{\mU,ij}=0$. 
We consider the linear mode connectivity of the two networks and define the difference function between interpolated network and original networks as $z_{\vx}(\alpha)=(\alpha\vv + (1-\alpha)\vv')^\top\sigma((\alpha\mU+(1-\alpha)\mU')\vx)-\alpha\vv^\top\sigma(\mU\vx)-(1-\alpha){\vv'}^\top\sigma(\mU'\vx)$, $\alpha \in [0, 1]$. The function over all inputs is defined as $z(\alpha)=\frac{1}{|\sX|}\int_\sX z_{\vx}(\alpha)d\vx$. We use $\left|z(\alpha)\right|$, $\left|\frac{dz(\alpha)}{d\alpha}\right|$ and $\left|\frac{d^2z(\alpha)}{d\alpha^2}\right|$ to depict the linear mode connectivity, showing the output changes along the $\alpha$ path.  With probability $1-\delta$, it has,
\small{
\begin{align}
&\left|z(\alpha)\right|\le \sqrt{2} b\sigma_\vv\sigma_\mU\log(8h/\delta) \sqrt{h}\sqrt{1-\rho_\mU},\\
&\left|\frac{dz(\alpha)}{d\alpha}\right| \leq 4\sqrt{2} b\sigma_\vv\sigma_\mU \log{(24h/\delta)}\sqrt{h}(\sqrt{1-\rho_\vv}+\sqrt{1-\rho_\mU}),\\
&\left|\frac{d^2z(\alpha)}{d\alpha^2}\right| \leq 8 b\sigma_\vv \sigma_\mU \log(4h/\delta) \sqrt{h}\sqrt{(1-\max\{\rho_\mU,\rho_\vv\})},
\end{align}
}
where $\rho_\vv$ and $\rho_\mU$ refer to the mask ratios (the proportion of zeros in the mask) of masks $\mM_{\vv}$ and $\mM_{\mU}$ respectively. 
\end{theorem}

\begin{remark} \label{rmk:1}    
    $\left|z(\alpha)\right|$ is the barrier given $\alpha$. $\left|\frac{dz(\alpha)}{d\alpha}\right|$ demonstrates the barrier function changes along the interpolation path $\alpha\in[0,1]$, and the smaller value means smaller changes. If $\left|\frac{dz(\alpha)}{d\alpha}\right|\to 0$, it means that $z(\alpha)$ is a constant, but it does not mean $z(\alpha)$ is a linear function of $\alpha$. 
    $\left|\frac{d^2z(\alpha)}{d\alpha^2}\right|$ reflects the linearity of function $z(\alpha)$, and if $\left|\frac{d^2z(\alpha)}{d\alpha^2}\right|\to 0$, it means that $z(\alpha)$ is linear w.r.t. $\alpha$.

\end{remark}

\begin{figure*}[!t]
    \centering
\includegraphics[width=0.245\linewidth]{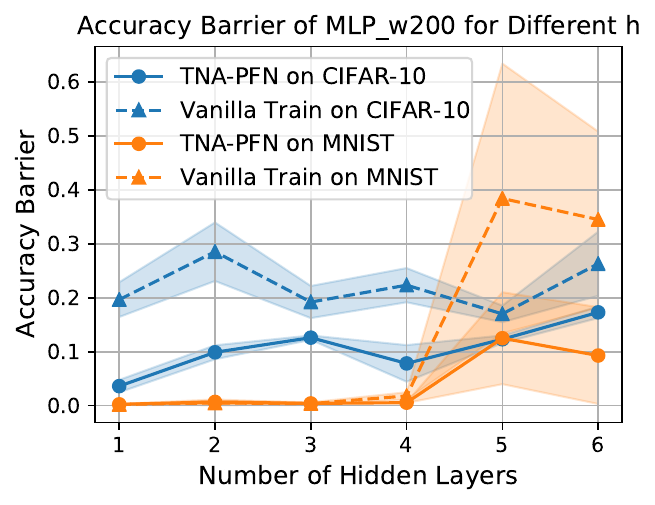}
\includegraphics[width=0.245\linewidth]{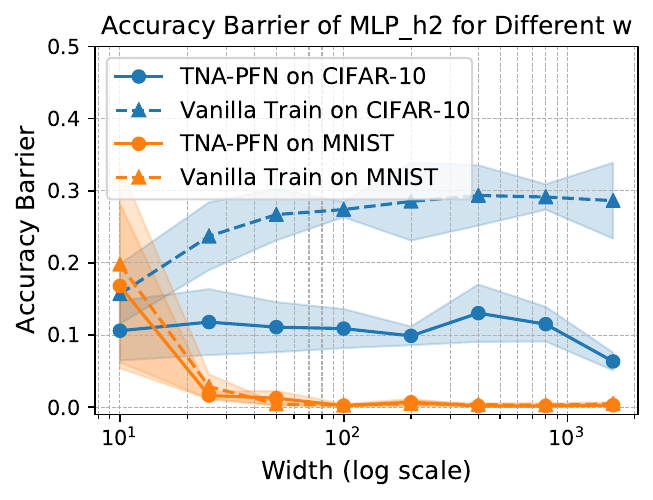}
\includegraphics[width=0.245\linewidth]{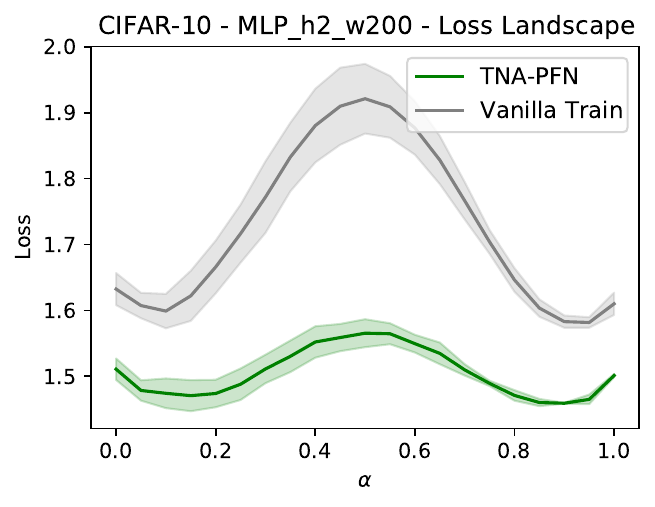}
\includegraphics[width=0.245\linewidth]{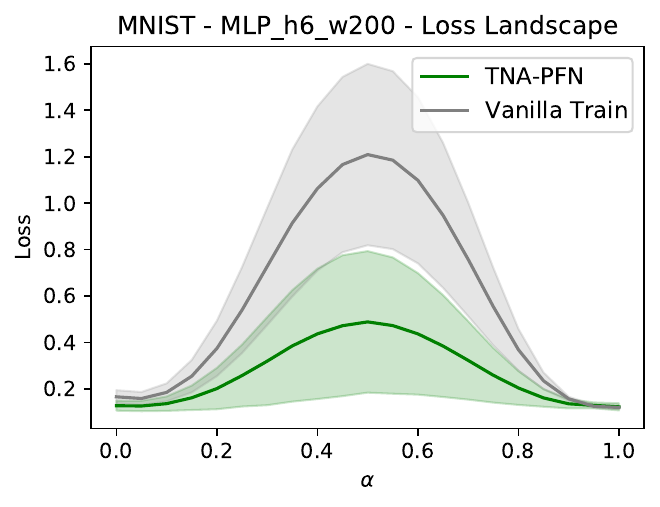}
\vspace{-0.35cm}
    \caption{\textbf{Left two: Accuracy barriers of MLP under different hidden layers ($h$) and widths ($w$). Right two: Loss landscapes of MLP.} For MLPs, if the barriers exist, TNA-PFN can reduce them. The shadow areas refer to the standard deviations.}
    \label{fig:lmc:mlp_barrier_landscape1}
\end{figure*}

\begin{table*}[t]
\centering
\caption{\textbf{The performances of post-matching methods after TNA-PFN.} Interpolated Accuracy (Interp. Acc.) means the accuracy of the linearly interpolated model, i.e., $\mathcal{A}(0.5 \mathbf{w}_1 + 0.5\mathbf{w}_2)$. ``Iter.'' refers to the number of iterations in the post-matching methods, reflecting the computation costs.}
\resizebox{\linewidth}{!}{
\begin{tabular}{@{}l|c|c|c|c|c|c|c|c}
\toprule
\multirow{2}{*}{} &\multicolumn{4}{c}{\textbf{CIFAR-10}} &\multicolumn{4}{c}{\textbf{MNIST}} \\
\cmidrule(lr){2-5} \cmidrule(lr){6-9}
 &\multicolumn{2}{c}{\textbf{MLP\_h2\_w200}} &\multicolumn{2}{c}{\textbf{ResNet20}} &\multicolumn{2}{c}{\textbf{MLP\_h5\_w200}} &\multicolumn{2}{c}{\textbf{MLP\_h6\_w200}} \\
\midrule
\textbf{Metrics\textbackslash Methods} &\textbf{Vanilla train} &\textbf{TNA-PFN }&\textbf{Vanilla train} &\textbf{TNA-PFN} &\textbf{Vanilla train} &\textbf{TNA-PFN} &\textbf{Vanilla train} &\textbf{TNA-PFN} \\
\midrule
Interp. Acc. w/o Post-matching &31.9±2.4 &\textbf{43.7±0.4} &36.1±4.3 &\textbf{46.2±4.7} &59.4±24.2 &\textbf{84.8±8.2} &63.7±15.6 &\textbf{87.5±8.9} \\
\midrule
Interp. Acc. after 10 Iter. of SA &32.2±2.2 &\textbf{43.7±0.4} &36.7±3.4 &\textbf{46.2±4.7} &59.7±24.2 &\textbf{85.4±8.0} &64.9±14.4 &\textbf{87.7±9.1} \\
Interp. Acc. after 100 Iter. of SA &31.9±2.4 &\textbf{43.7±0.4} &36.1±4.3 &\textbf{46.2±4.7} &60±24.1 &\textbf{86.9±7.6} &64.2±15.1 &\textbf{88.2±7.9} \\
\midrule
Interp. Acc. after WM &44.7±1.3 &\textbf{48.5±0.9} &\textbf{53.7±2.9} &53.6±2.5 &96.9±0.3 &\textbf{97.1±0.2} &96.8±0.3 &\textbf{96.9±0.4} \\
Required Iter. in WM &5.2±1.0 &\textbf{4.8±1.5} &4.6±0.5 &\textbf{2.5±0.2} &10.4±1.2 &\textbf{7.6±3.8} &11.2±1.8 &\textbf{7.33±4.2} \\
\bottomrule
\end{tabular}}
\label{tab:lmc:post-match}
\end{table*}

\autoref{thm:1} shows how TNA-PFN improves LMC by masking some weights from updating (setting $\rho_\vv$ and $\rho_\mU$ larger than zeros), where the networks learn in a unified permutation subspace.

\subsubsection{Empirical Results on Linear Mode Connectivity}

In this subsection, we will conduct experiments to validate the effectiveness of TNA-PFN in improving LMC. If not mentioned otherwise, the mask ratio $\rho$ of TNA-PFN is 0.4 (the hyperparameter which is mild across various settings).

\textit{Permutation symmetry: an intuitive example.} To provide neuron-level evidence, we design a demonstration example that showcases and visualizes permutation symmetry during standard training and the permutation asymmetry encouraged by TNA-PFN. We use the synthetic data and MLP in the 2nd Polynomial dataset of \autoref{tab:lmc:more_dl_tasks}, and we use an initialization of ordered weights for better visualization and validation. The neuron weight changes from initialization to the final model are visualized, and it is found that TNA-PFN can anchor the unfrozen weights. 
Results are shown in Figures 1 and 2 of the Appendix.\looseness=-1

\begin{figure}[t]
  \centering
    \includegraphics[width=0.6\linewidth]{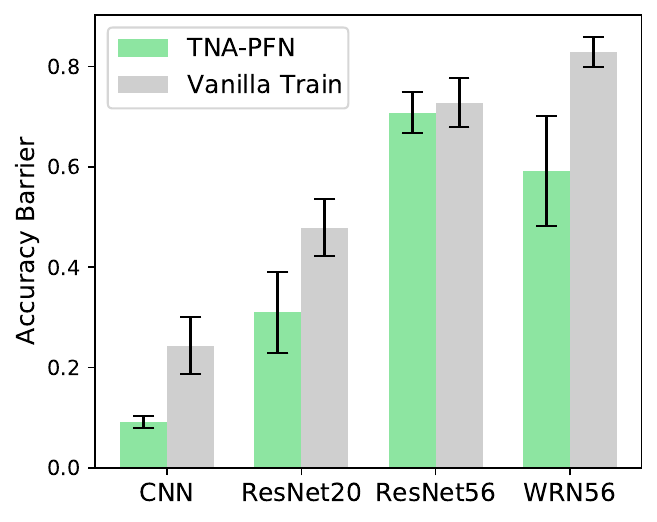}
    \caption{\textbf{Accuracy barriers under different model architecture.} WRN56 abbreviates for WideResNet56. CIFAR-10.}
    \label{fig:lmc:model_archs_barrier}
\end{figure}

\textit{Different model depths, widths, and architectures.} In \autoref{fig:lmc:mlp_barrier_landscape1}, we conduct experiments on MLP with different hidden layers and widths. For MNIST~\cite{lecun-mnisthandwrittendigit-2010}, we find shallower and wider networks will not cause barriers, which is consistent with the previous observations~\cite{entezari2021role}. For CIFAR-10~\cite{krizhevsky2009learning}, the barriers always exist under various depths and widths. Our proposed TNA-PFN can obviously reduce the accuracy barriers from 0.3-0.4 to 0.1, and we also visualize the loss landscapes, which illustrate the barrier reductions.


We study the LMC of simple CNN and ResNets and present the results in \autoref{fig:lmc:model_archs_barrier}. ResNets~\cite{he2016deep} have higher barriers than simple CNN, and the barriers are exacerbated when the networks are deeper or wider. It is suggested that TNA-PFN can lower the barriers under different architectures. \looseness=-1

Generally, we observe that TNA-PFN has more dominant advantages when the models are wider, and the observations are: (1) the second figure in \autoref{fig:lmc:mlp_barrier_landscape1}: for CIFAR-10 when width increases, the barriers of vanilla training go up while the barriers of TNA-PFN go down; (2) \autoref{fig:lmc:model_archs_barrier}: the barrier reduction of TNA-PFN is more obvious for WRN56 compared with ResNet56. This observation is consistent with previous literature about permutation symmetry and linear mode connectivity~\cite{entezari2021role}.

\begin{table*}[!t] 
	\centering
 \begin{minipage}{0.448\linewidth}
    \centering
        \caption{\textbf{Linear mode connectivity of non-random initializations.} Dataset: CIFAR-10. \textit{Semi-trained CNN:} The initialization is first trained on a disjoint subset of CIFAR-10 for 0.5 epoch. \textit{Pretrained ResNet18:} Pretrained on ImageNet.}
        \vspace{-0.08cm} 
        \resizebox{\linewidth}{!}{
            \begin{tabular}{@{}l|c|c|c}
            \toprule
            \textbf{Models} &\textbf{Metrics} &\textbf{Vanilla Train} &\textbf{TNA-PFN} \\
            \hline
            \multirow{4}{2cm}{Semi-trained \\CNN} 
            &Avg. Acc. &64.4±0.7 &\textbf{65.8±0.2} \\
            &Interp. Acc. &52.8±2.3 &\textbf{63.3±1.2} \\
            \cmidrule{2-4}
            &Acc. Barrier &0.181±0.027 &\textbf{0.0413±0.018} \\
            &Loss Barrier &0.306±0.058 &\textbf{0.0762±0.038} \\
            \midrule
            \multirow{4}{2cm}{Pretrained \\ ResNet18} 
            &Avg. Acc. &65.3±1.2 &\textbf{69.1±0.5} \\
            &Interp. Acc. &11.9±0.39 &\textbf{54.9±8.2} \\
            \cmidrule{2-4}
            &Acc. Barrier &0.817±0.003 &\textbf{0.205±0.12} \\
            &Loss Barrier &0.957±0.45 &\textbf{0.395±0.2} \\
            \bottomrule
            \end{tabular}}
        \label{tab:lmc:nonrandom_init}
    \end{minipage}
        \hfill
	\begin{minipage}{0.545\linewidth}\centering
            \caption{\textbf{Linear mode connectivity of multi-model fusion.} The number of models is 5.}
            \vspace{0.05cm} 
        \resizebox{\linewidth}{!}{
            \begin{tabular}{@{}l|c|c|c}
            \toprule
            \textbf{Datasets / Models} &\textbf{Metrics} &\textbf{Vanilla Train} &\textbf{TNA-PFN} \\
            \midrule
            \multirow{3}{*}{CIFAR-10 / CNN} &Avg. Acc. &63.1±0.6 &\textbf{65.5±0.3} \\
            &Interp. Acc. &21.3±9.1 &\textbf{48.3±7.2} \\
            \cmidrule{2-4}
            &Acc. Barrier &0.663±0.14 &\textbf{0.264±0.11} \\
            \midrule
            \multirow{3}{*}{CIFAR-10 / MLP\_h2\_w200} &Avg. Acc. &44.2±0.5 &\textbf{48.4±0.5} \\
            &Interp. Acc. &21.6±1.9 &\textbf{36.6±1.4} \\
            \cmidrule{2-4}
            &Acc. Barrier &0.511±0.043 &\textbf{0.245±0.035} \\
            \midrule
            \multirow{3}{*}{MNIST / MLP\_h5\_w200} &Avg. Acc. &96.5±0.3 &\textbf{96.5±0.2} \\
            &Interp. Acc. &34.5±15.5 &\textbf{87.1±9.4} \\
            \cmidrule{2-4}
            &Acc. Barrier &0.643±0.16 &\textbf{0.0974±0.096} \\
            \bottomrule
            \end{tabular}}
            \label{tab:lmc:multi_model_fusion}
	\end{minipage}
\end{table*}

\textit{The role of post-hoc neuron alignment methods after training-time alignment.} We consider simulated annealing (SA)~\cite{entezari2021role} and weight matching (WM)~\cite{ainsworth2022git} after TNA-PFN in \autoref{tab:lmc:post-match}. 
SA requires large computations, and we notice the improvements are also marginal. Under limited computation budgets (10 or 100 iterations), we find that TNA-PFN can reach a higher result than vanilla training after SA. For WM, it is indicated that after TNA-PFN, the required iterations are shortened while the interpolated accuracies are similar. The results reveal that training-time neuron alignment can reduce the costs of post-matching and remain similar or even better post-matched LMC.

\begin{table}[t]
\caption{\textbf{Results of loss barriers on more deep learning tasks.}}
    \centering
\resizebox{0.95\columnwidth}{!}{
\begin{tabular}{l|cc|c}
\toprule
\textbf{Methods\textbackslash Datasets} &\textbf{2nd Polynomial} &\textbf{3rd Polynomial} & \textbf{IMDb} \\
\midrule
Vanilla Train & 0.268±0.061 &0.0554±0.047 & 0.710±0.17\\
TNA-PFN & \textbf{0.0381±0.0096} &\textbf{0.0355±0.023} &\textbf{0.375±0.17}\\
\bottomrule
\end{tabular}}
    \label{tab:lmc:more_dl_tasks}
\end{table}

\begin{figure}
    \centering
    \includegraphics[width=1.0\linewidth]{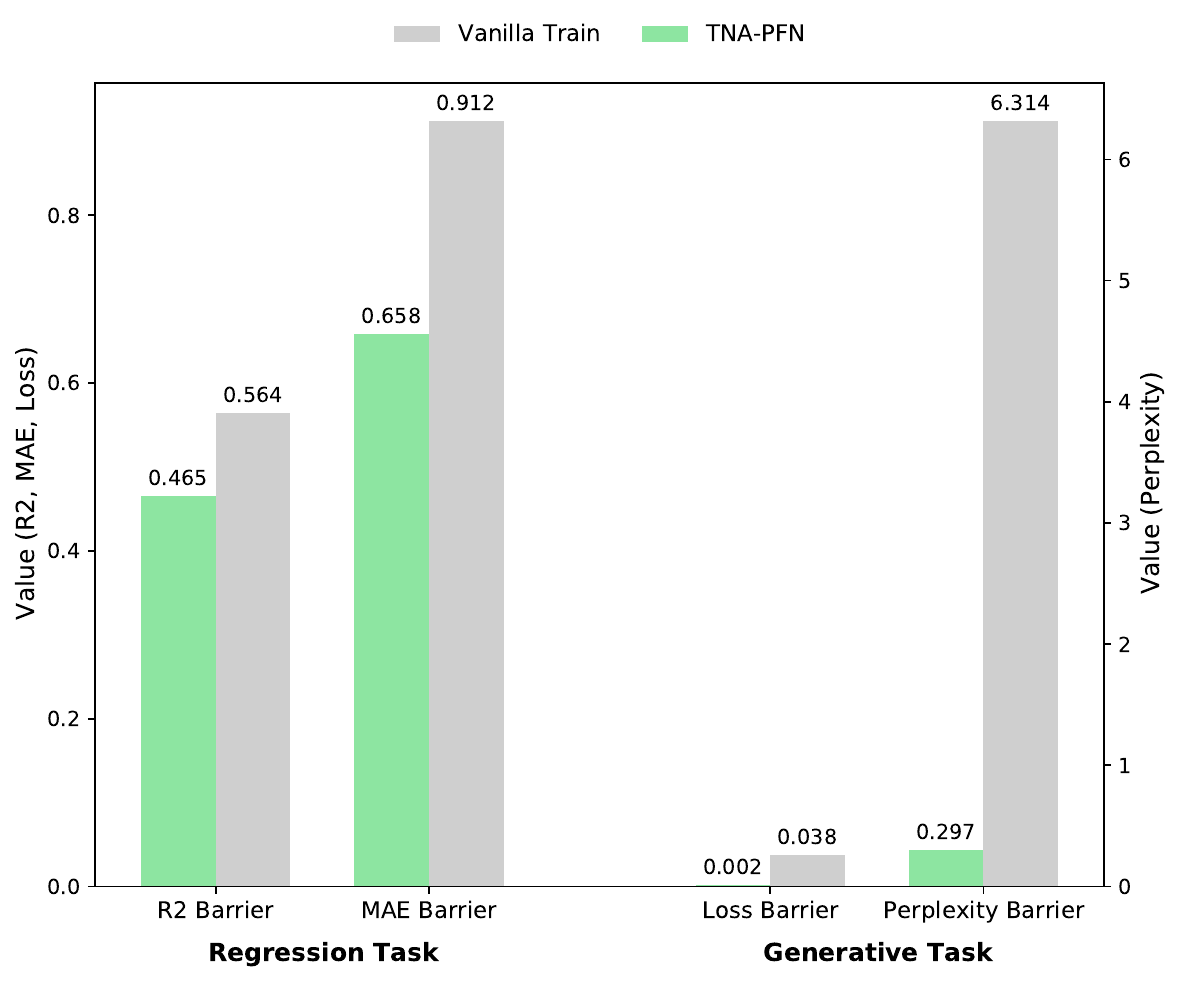}
    \caption{\textbf{The connectivity barriers of vanilla training and TNA-PFN under regression and generative tasks.} The regression task is training MLP models on the California Housing Dataset with $\rho=0.3$ for TNA-PFN, and the generative task is finetuning GPT-2 (small) on WikiText-2 with $\rho=0.5$ for TNA-PFN.}
    \label{fig:regression_and_generative_tasks}
\end{figure}

\textit{More deep learning tasks.} We conduct more experiments beyond vision tasks and display the results in \autoref{tab:lmc:more_dl_tasks}. \textit{i) Polynomial approximation task}~\cite{pena2023re,von2019continual}: we use an MLP with one hidden layer to approximate the second and third polynomial functions: $y = 2x^2-1, y=(x-3)^3$. \textit{ii) Sentiment analysis of text}~\cite{liu2022omnigrok}: we use an LSTM~\cite{graves2012long} to predict the sentiment of IMDb reviews~\cite{maas2011learning}. It can be seen that the loss barriers are decreased by training-time alignment under both polynomial approximation and sentiment analysis tasks. \textit{iii) Large-scale dataset}: We also implement the experiments on a large-scale dataset, the subset of ImageNet~\cite{deng2009imagenet,tinyimagenet}. The result is shown in the Appendix. In addition, we have conducted the experiments under regression tasks and generative language tasks in \autoref{fig:regression_and_generative_tasks}. We trained MLP regression models with one hidden layer and a width of 16 on the California Housing Dataset~\cite{paces1997sparse} and evaluated $R^2$ and MAE barriers across runs. We fine-tuned GPT-2 (small)~\cite{radford2019language} on WikiText-2~\cite{merity2016pointer} under multiple seeds and mild hyperparameter variations. The results in \autoref{fig:regression_and_generative_tasks} show that TNA-PFN can effectively reduce the connectivity barriers under these tasks.

\textit{Results under non-random initializations.} We also examine whether training-time alignment can help when the initializations are not random, which commonly occurs in the pretrained-finetuned paradigm and federated learning. 
As presented in \autoref{tab:lmc:nonrandom_init}, TNA-PFN is more advantageous to improve LMC under non-random initializations. TNA-PFN nearly clears the barriers under the setting of CNN and CIFAR-10. For the pretrained ResNet18, we observe that TNA-PFN reduces the barrier by a large margin while improving individual accuracies.

\textit{Layer-wise analysis.} We conduct a layer-wise analysis of TNA-PFN to see which layer matters most in improving LMC in the Appendix.

\textbf{Note:} Completely eliminating LMC barriers in training time is impossible because training will introduce stochasticity, and the stochasticity will result in network permutations. Though we introduce TNA-PFN to reduce the barrier by reducing the stochasticity, the stochasticity cannot be completely eliminated, nor the barrier. Also, the proposed training-time methods are orthogonal to the post-training matching methods, and both of them have their advantages. However, instead of LMC, this paper focuses more on model fusion applications and shows how they can benefit from training-time neuron alignment.\looseness=-1

\section{Training-time Neuron Alignment for Fusion of Pretrained Models}\label{sec:pretrained_fusion}
In this section, we verify TNA-PFN in model fusion with multiple models, especially model fusion of pretrained models. 

\subsection{LMC of Multi-model Fusion}\label{subsec:vanilla_model_fusion}
Vanilla LMC studies the connectivity between two SGD solutions. To bridge LMC to multi-model fusion, we conduct experiments on LMC with multiple models and the results are shown in \autoref{tab:lmc:multi_model_fusion}.
We consider the connectivity of 5 independently trained models by assigning a uniformly weighted fusion after training. We test the generalization of the fused model as interpolated accuracy (Interp. Acc.), compare it with the averaged accuracy of independent models (Avg. Acc.), and compute the accuracy barrier (Acc. Barrier = (Avg. Acc. - Interp. Acc.) / Avg. Acc.). It is evident that after the training-time alignment, the interpolated accuracies are largely promoted by up to 152\% and the barriers are much lower with a maximal reduction of 84.9\%, showing TNA-PFN's prospects in broader model fusion applications like federated learning. It is also intriguing to observe that the averaged accuracies also increase after TNA-PFN. We explain this phenomenon that partially fixing some weights may play the role of regularization, and for some models with redundant neurons, this regularization can also help in generalization. 

The experimental results show the prospects of TNA-PFN in multi-model fusion, and then we will test it in more practical scenarios---improving multi-model fusion of pretrained transformer models under pretrained-finetuned paradigm, where model soup and ColD fusion are taken as applications. 

\subsection{Model Soup}

\begin{table}
\centering
    \caption{\textbf{Results of model soup \cite{wortsman2022model} for finetuning pretrained transformers.} The initialized model is ViT-B/32 pretrained on CLIP. The number of models per experiment is 5. "Avg. Acc." refers to the averaged accuracy of individual models. "Diff\_lr" ("Diff\_seed") refers to model soup with different learning rates (random seeds).} 
\resizebox{\linewidth}{!}{
    \begin{tabular}{@{}l|c|c|c}
    \toprule
    \textbf{Dataset / Type of Soup} &\textbf{Metrics} &\textbf{Vanilla Train} &\textbf{TNA-PFN} \\
    \hline
    \multirow{3}{*}{CIFAR-10 / Diff\_lr} 
    &Avg. Acc. &\textbf{80.3} &79.9 (\textcolor{someorange}{0.4$\downarrow$}) \\
    &Uniform Soup&\textbf{91.6} &91.2 (\textcolor{someorange}{0.4$\downarrow$}) \\
    &Greedy Soup&96.7 &\textbf{96.9 (\textcolor{limegreen}{0.2$\uparrow$})} \\
    \cmidrule{2-4}
    \multirow{3}{*}{CIFAR-10 / Diff\_seed} 
    &Avg. Acc. &96.3 &\textbf{96.6 (\textcolor{limegreen}{0.3$\uparrow$})} \\
    &Uniform Soup&97.4 &\textbf{97.6 (\textcolor{limegreen}{0.2$\uparrow$})} \\
    &Greedy Soup&97.5 &\textbf{97.6 (\textcolor{limegreen}{0.1$\uparrow$})} \\
    \midrule
    \multirow{3}{*}{CIFAR-100 / Diff\_lr} 
    &Avg. Acc. &55.5 &\textbf{56.6 (\textcolor{limegreen}{1.1$\uparrow$})} \\
    &Uniform Soup&66.6 &\textbf{67.6 (\textcolor{limegreen}{1.0$\uparrow$})} \\
    &Greedy Soup&80.9 &\textbf{82.6 (\textcolor{limegreen}{1.7$\uparrow$})} \\
    \cmidrule{2-4}
    \multirow{3}{*}{CIFAR-100 / Diff\_seed} 
    &Avg. Acc. &64.8 &\textbf{67.8 (\textcolor{limegreen}{3.0$\uparrow$})} \\
    &Uniform Soup&62.1 &\textbf{67.2 (\textcolor{limegreen}{5.1$\uparrow$})} \\
    &Greedy Soup&67.7 &\textbf{73.3 (\textcolor{limegreen}{5.6$\uparrow$})} \\
    \bottomrule
    \end{tabular}
    }
    \label{tab:model_soup_vit}
\end{table}

\begin{figure*}
    \centering
    \includegraphics[width=0.48\linewidth]{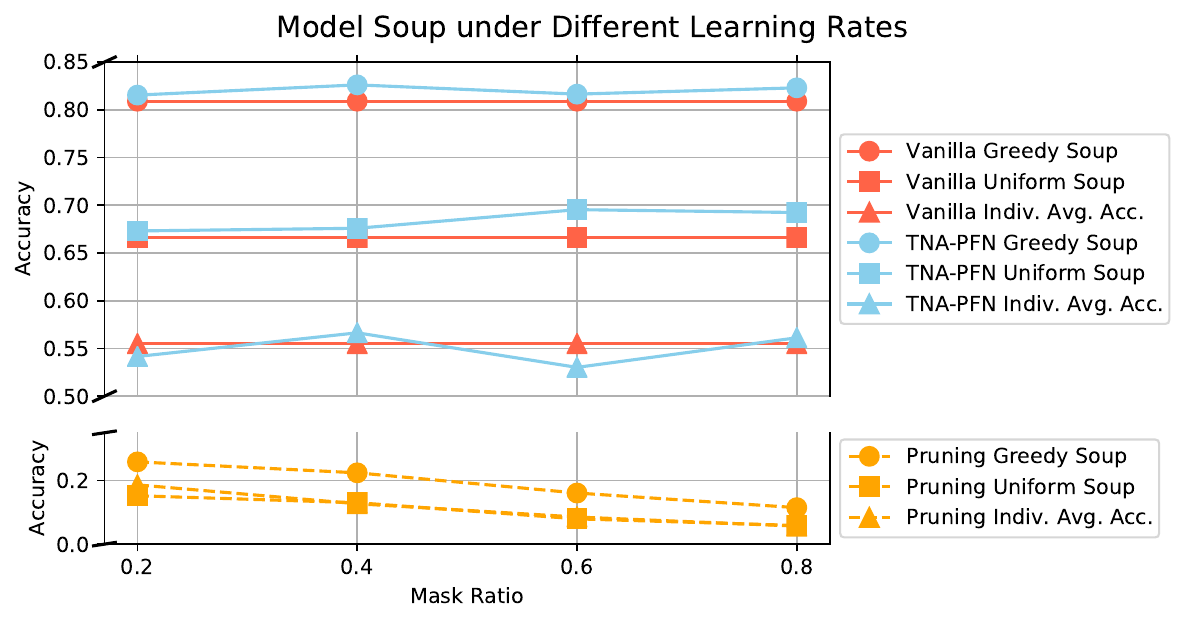}
    \includegraphics[width=0.48\linewidth]{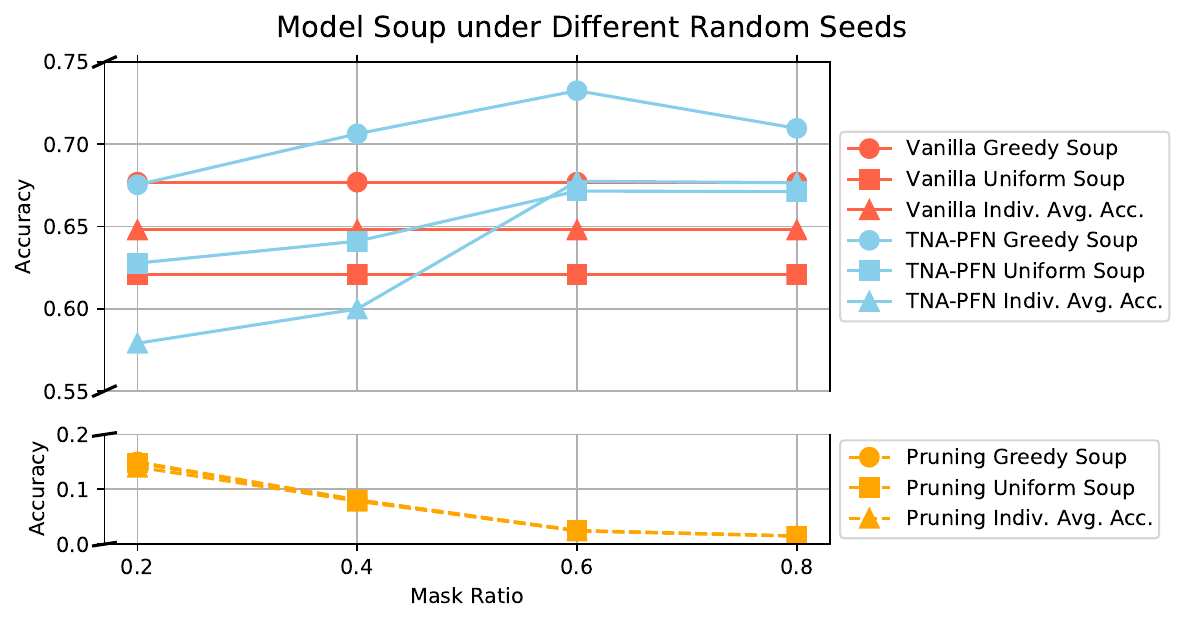}
    \caption{\textbf{Results of model soup under different mask ratios.} The initialization is ViT-B/32 and the finetuned dataset is CIFAR-100.}
    \label{fig:model_soup_fig}
\end{figure*}

\begin{figure}[t]
  \centering
    \includegraphics[width=0.93\linewidth]{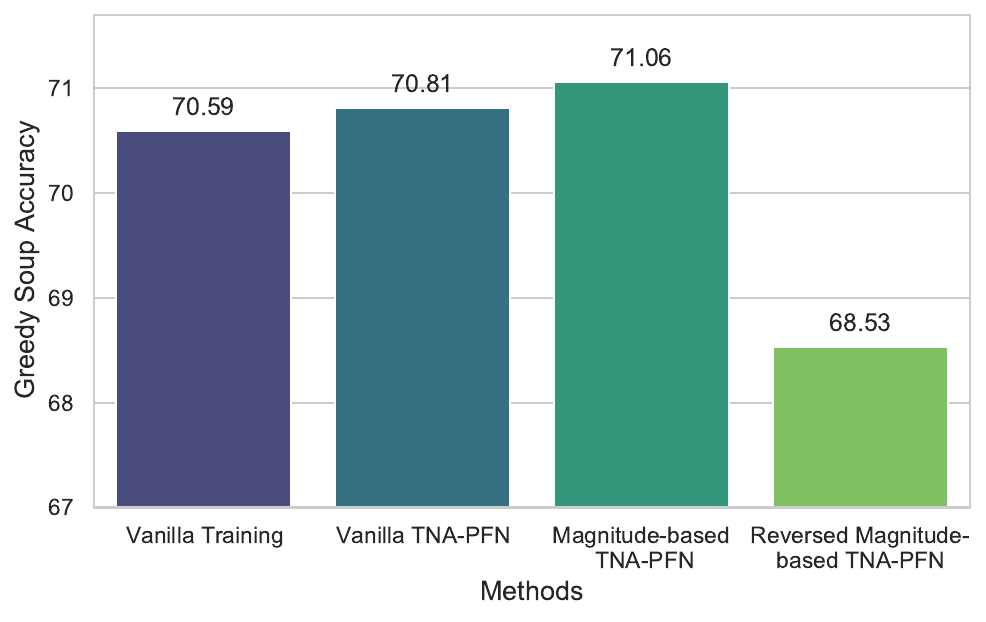}
    \caption{\textbf{Study of magnitude-based TNA-PFN for model soup.} ViT-B/32, CIFAR-100, $\rho=0.4$, greedy soup with different random seeds.}
    \label{fig:model_soup_magnitude_based}
\end{figure}

\textit{Model soup} is a technique for improving the generalization of finetuned foundation models via model fusion~\cite{wortsman2022model}. Conventional finetuning uses grid search to train several models with different hyperparameters and selects the one with the best generalization. However, model soup utilizes these model checkpoints via model fusion, obtaining a generalized model better than the one with the best hyperparameters. Specifically, model soup incorporates finetuning several models from the same initialization under different hyperparameters (e.g., batch sizes or learning rates), data augmentation, and random seeds and then greedily (greedy soup) or uniformly (uniform soup) fusing them into one for boosting the generalization on the finetuned task. 

In \autoref{tab:model_soup_vit}, we conduct model soup experiments using ViT-B/32~\cite{vit_first_paper} pretrained on CLIP~\cite{pmlr-v139-radford21a}. The results show greedy soup can realize the maximal generalization gain compared with uniform soup and vanilla training, which is consistent with the literature. 
It demonstrates that TNA-PFN can consistently promote greedy soups' generalization. We also show the results of TNA-PFN using different mask ratios in~\autoref{fig:model_soup_fig}. It can be seen that pruning will result in generalization failure even if the pruning ratio is low (e.g. 0.2) for pretrained ViT. For individual model accuracies, TNA-PFN has similar or lower accuracies compared with vanilla training, but TNA-PFN outperforms the vanilla in greedy and uniform soups. This shows the effectiveness of training-time alignment in TNA-PFN when model fusion. In addition, it is found that for pretrained ViT, even freezing most model parameters (e.g., right of \autoref{fig:model_soup_fig}, mask ratio 0.6 and 0.8 for TNA-PFN), the generalization will not be lost, and TNA-PFN can largely improve model soup. It suggests that pretrained models have a large redundancy in model parameters.

\textit{Magnitude-based TNA-PFN.} In vanilla TNA-PFN, fixed neuron weights are randomly selected since, in LMC, the models are randomly initialized. While in model soup, pretrained initialization is used for finetuning; therefore, we explore how to select the fixed weights instead of being random. Inspired by magnitude-based pruning~\cite{han2015learning,han2015deep}, we hypothesize that weights with higher magnitude (high absolute value) are more suitable to be anchors that are frozen/fixed during training. The results are in \autoref{fig:model_soup_magnitude_based}, where ``vanilla TNA-PFN'' denotes the method of randomly selecting fixed weights in TNA-PFN, ``magnitude-based TNA-PFN'' involves fixing the top $\rho$ proportion of the highest absolute weights in each layer, and ``reversed magnitude-based TNA-PFN'' refers to fixing the weights with the lowest absolute values. It shows that the magnitude-based TNA-PFN reaches a better result than the vanilla TNA-PFN, whereas the reserved magnitude-based method is even worse than vanilla training.

\subsection{ColD Fusion} \label{subsec:model_soup_cold_fusion}

\begin{table}[t]
\centering
    \caption{\textbf{Results of ColD fusion \cite{cold_fusion} for pretrained language models.} The model is RoBERTa-base and the three datasets are from GLUE~\cite{wang2018glue}. ``Sum'' refers to the joint testset combining the three datasets.} 
\resizebox{0.93\linewidth}{!}{
    \begin{tabular}{@{}l|ccc|c}
    \toprule
    \textbf{Methods} &\textbf{COLA} &\textbf{RTE} &\textbf{MRPC} &\textbf{Sum}\\
    \midrule
    FedAvg &77.66  &82.31  &77.33  &77.66 \\
    \midrule
    \rowcolor{gray!20}FedAvg+TNA-PFN &80.73  &83.75  &\textbf{81.86}  &81.43 \\
    \rowcolor{gray!20}FedPFN &\textbf{82.74}  &\textbf{85.56}  &81.16  &\textbf{82.66} \\
    \rowcolor{gray!20}FedPNU &81.50  &83.39  &79.48  &80.11 \\
    \bottomrule
    \end{tabular}
    }
    \label{tab:pretrained_lm}
\end{table}

\textit{ColD fusion}, abbreviated from Collaborative Descent Fusion, is proposed to improve the pretrained language models' multi-task generalization via collaborative model fusion~\cite{cold_fusion}. It finds that model fusion of fine-tuned language models can be recycled to continually improve the pretrained model they are based upon. In this subsection, we verify whether training-time neuron alignment methods can improve ColD fusion performances.

The original procedure of ColD fusion resembles FedAvg~\cite{mcmahan2017communication}, where models are trained on different task datasets, and iterative model fusion is conducted after several epochs; therefore, we use FedAvg to proxy for vanilla ColD fusion. The experimental results are shown in \autoref{tab:pretrained_lm}, where three variants of TNA-PFN: FedAvg+TNA-PFN, FedPFN, and FedPNU, are shown. FedPFN and FedPNU are two federated learning algorithms derived from TNA-PFN, and they will be introduced in \autoref{sec:TNA4FL}. In \autoref{tab:pretrained_lm}, it is shown that three training-time alignment methods all surpass vanilla ColD fusion (i.e., FedAvg) in three GLUE \cite{wang2018glue} datasets and overall performance, further demonstrating the potential of training-time neuron alignment methods in improving model fusion of pretrained models.\looseness=-1

Additionally, our TNA-PFN-based methods only apply a sparse gradient mask during training, which only adds little computation and is flexible and applicable across any model architecture. Compared with the post-hoc neuron alignment methods, our methods can boost the generalization of finetuning foundation models for free.

\noindent\textbf{Discussions on hyperparameter $\rho$:} In general, TNA-PFN is not sensitive to $\rho$; it can work well under a high range without losing accuracy, but improving connectivity. For instance, across CNN/MLP on CIFAR-10 and MNIST, TNA-PFN improves linear mode connectivity (LMC) for a broad range (e.g., $\rho \in [0.3,0.6]$ on CIFAR-10/CNN), with the default $\rho=0.4$ working well in most cases.
\textit{When to use smaller $\rho$:} In small-data or tight-capacity settings (e.g., some CIFAR-10 model-soup variants), too large a $\rho$ can act as strong regularization and slightly reduce single-model accuracy. In those cases, \emph{start with $\rho=0.2\sim0.3$}, which preserves capacity while still creating useful asymmetry for alignment.
\textit{Recommended default and tuning recipe:} We recommend a simple recipe: begin at $\rho=0.4$ (our default in Sec.~III), evaluate both single-run accuracy and LMC (or fusion accuracy), then adjust by $\pm0.1$ if single-model accuracy drops or barriers persist. For pretrained transformers, a modest $\rho$ combined with magnitude-based anchors (Sec.~IV.B, Fig.~7) is especially effective.\looseness=-1

\section{Training-time Neuron Alignment for Model Fusion in Federated Learning}
\label{sec:TNA4FL}
Federated learning (FL) is a privacy-preserving and collaborative training paradigm, and it utilizes model fusion for aggregating multiple local models into a global one. FL is an important application of model fusion. In this section, whether training-time neuron alignment methods can improve the model fusion of federated learning is studied. Derived from TNA-PFN, two FL methods, namely FedPFN and FedPNU, are proposed.

\subsection{Preliminary of Federated Learning}
Federated learning usually involves a server and $n$ clients to jointly learn a global model without data sharing, which is originally proposed in Mcmahan et al.~\cite{mcmahan2017communication}. 
Denote the set of clients by $\mathcal{S}$, the labeled data of client $i$ by $\mathcal{D}_i=\{(x_j, y_j)\}_{j=1}^{N_i}$ , and the parameters of the current global model by $\mathbf{w}_{g}^{t}$. 
FL starts with client training in parallel, initializing each clients' model $\mathbf{w}_{i}^{t}$ with $\mathbf{w}_{g}^{t}$.

FL is more communication-efficient than conventional distributed training, in that it assumes the clients train the models for epochs (the full data) instead of iterations (the mini-batch data) between the communications to the server. The number of local epochs is denoted as $E$.

In each local epoch, clients conduct SGD update with a local learning rate $\eta_l$, each SGD iteration shows as
\begin{small}
\begin{align}
\textbf{Client training: } \mathbf{w}_{i}^{t} \leftarrow \mathbf{w}_{i}^{t} - \eta_l \nabla \ell(B_k, \mathbf{w}_{i}^{t}), \text{ for }  k=1,2,\cdots,K, \label{eq_client}
\end{align}
\end{small}
where $\ell$ is the loss function and $B_k$ is the mini-batch sampled from $\mathcal{D}_i$ at the $k$th iteration. 
After the client local updates, the server samples $m$ clients for mode fusion. 
Then, the server conducts FedAvg to aggregate the local models into a new global model, which is weighted by the proportions of data sizes.

\begin{small}
\begin{align}
\textbf{FedAvg model fusion: }  \mathbf{w}_{g}^{t+1} = \sum_{i=1}^{m} \lambda_{i}\textbf{w}_{i}^{t}, ~\lambda_{i}=\frac{|\mathcal{D}_{i}|}{|\mathcal{D}|}, \forall i \in [m]. \label{eq_FedAvg}
\end{align}
\end{small}

The sum of clients' data is denoted as $\mathcal{D} = \bigcup_{i\in\mathcal{S}} \mathcal{D}_i$. With the updated global model $\mathbf{w}_{g}^{t+1}$, it then starts the next round of client training. 
The procedure of FL, therefore iterates between \autoref{eq_client} and \autoref{eq_FedAvg}, for $T$ communication rounds.

The IID data distributions of clients refer to each client's distribution $\mathcal{D}_i$ is IID sampled from $\mathcal{D}$. However, in practical FL scenarios, data heterogeneity exists among clients whose data are non-IID with each other. Each client may have different data distributions in the input (e.g. image distribution) or output (e.g. label distribution).

Due to data heterogeneity of clients and large local training epochs, local models in FL meet the model drift problem~\cite{wang2020tackling,karimireddy2020scaffold,pmlr-v202-li23s}, resulting in neuron misalignment during model fusion~\cite{wang2020federated,li2022federated,yurochkin2019bayesian}.

\begin{table*}[t]
    \footnotesize
    \centering
    \caption{ \textbf{Top-1 test accuracy (\%) achieved by comparing the FL methods on three datasets with different model architectures} ($E=5$). 
    \textbf{Bold} fonts highlight the best two methods in each setting. 
    }
    \resizebox{\linewidth}{!}{
    \begin{tabular}{c|cc|cc|cc|cc|cc|cc}
    \toprule
    \textbf{Datasets}&\multicolumn{4}{c}{\textbf{FashionMNIST}}&\multicolumn{4}{c}{\textbf{CIFAR-10}}&\multicolumn{4}{c}{\textbf{CIFAR-100}}\\
    \cmidrule(lr){1-5}
    \cmidrule(lr){6-9}
    \cmidrule(lr){10-13}
    \textbf{dir} &\multicolumn{2}{c}{100}&\multicolumn{2}{c}{0.1}&\multicolumn{2}{c}{100}&\multicolumn{2}{c}{0.1}&\multicolumn{2}{c}{100}&\multicolumn{2}{c}{0.1}\\
    \midrule
    \textbf{Methods\textbackslash Models} &MLP&LeNet&MLP&LeNet&CNN&ResNet&CNN&ResNet&CNN&ResNet&CNN&ResNet\\
    \midrule
    FedAvg &88.7±0.5	&90.5±0.1	&81.7±2.7	&83.5±3.8	&65.4±1.2	&73.4±2.1	&57.5±1.3	&50.9±1.8	&18.9±0.9	&26.4±0.4	&22.6±0.9	&28.5±1.3\\
    \midrule
    FedProx &88.0±0.1	&90.0±0.2	&82.6±0.9	&85.8±0.7	&65.4±0.9	&65.5±0.8	&59.7±1.1	&49.9±2.1	&\textbf{27.7±0.5}	&26.7±0.4	&24.7±0.0	&23.0±1.5\\
    FedDyn &88.2±0.2 &85.3±7.7 &65.2±0.4 &68.9±5.7 &33.2±1.6 &19.2±1.2 &20.5±2.1 &19.0±0.9 &23.7±4.7 &18.6±1.1	&21.1±0.9 &1.0±0.0 \\
    FedRoD &87.9±1.3 &89.9±0.4 &81.7±1.4 &85.5±1.2 &64.4±0.8 &74.0±1.4 &58.6±1.5 &52.2±1.3 &20.3±0.5 &26.4±0.9 &19.5±0.5 &26.5±0.5 \\
    FedNH &86.6±1.9 &87.5±0.6 &69.2±6.8 &72.3±4.6 &66.5±1.3 &71.6±0.7 &49.1±6.1 &28.9±12.7 &21.5±0.5 &26.1±0.1 &20.1±0.2 &25.6±0.8 \\
    FedETF &88.3±0.4 &89.9±0.3 &81.9±1.1 &85.7±1.3 &65.0±1.0 &71.8±0.4 &53.6±0.9 &\textbf{59.8±2.4} &22.1±0.2 &28.4±0.7 &20.1±0.4 &27.7±0.5 \\
    \midrule
    FedDF &\textbf{89.1±0.1}	&90.3±0.2	&81.3±2.8	&86.0±1.9	&66.3±0.8	&\textbf{75.6±3.3}	&57.6±3.0	&55.2±1.4	&21.4±0.3	&28.5±1.0	&24.2±0.2	&31.2±1.2\\
    \midrule
    \rowcolor{gray!20}FedPFN &88.8±0.1	&\textbf{90.6±0.1}	&81.8±1.7	&84.9±2.8	&66.9±0.6	&73.7±1.3	&\textbf{62.2±0.5}	&51.4±0.7	&20.9±0.7	&27.3±0.3	&24.9±1.2	&34.5±2.4\\
    \rowcolor{gray!20}FedPNU &88.7±0.2	&90.4±0.2	&\textbf{83.2±0.8}	&\textbf{86.6±1.0}	&\textbf{67.5±0.3}	&73.5±2.5	&\textbf{61.3±0.4}	&55.9±1.6	&22.1±0.5	&29.4±0.0	&24.8±0.2	&\textbf{35.3±1.5}\\
    \midrule
    \rowcolor{gray!20}FedDF+FedPFN &\textbf{88.9±0.2}	&90.5±0.1	&80.7±3.3	&\textbf{86.4±2.0}	&\textbf{67.9±0.4}	&73.0±1.2	&59.3±3.6	&54.4±5.1	&\textbf{27.8±0.8}	&\textbf{31.0±1.3}	&\textbf{27.0±0.2}	&31.1±1.2\\
    \rowcolor{gray!20}FedDF+FedPNU &88.7±0.1	&\textbf{90.7±0.2}	&\textbf{82.1±2.4}	&\textbf{86.4±1.8}	&66.4±0.8	&\textbf{74.1±1.3}	&60.0±2.4	&\textbf{57.1±4.1}	&22.2±0.9	&\textbf{30.8±1.2}	&\textbf{25.8±0.7}	&\textbf{35.2±1.2}\\
    \bottomrule
    \end{tabular}
    }
    \label{table:fl:first_table}
\end{table*}

\subsection{Proposed Methods} \label{subsec:fl_methods}
In this subsection, we propose two variants of TNA-PFN in FL, the first is called Federated Learning with Partially Fixed Neurons (FedPFN) and the second is Federated Learning with Progressive Neuron Updating (FedPNU). 

\definecolor{green_plus_red}{rgb}{0.5, 0.7, 0.0} 
\begin{algorithm}[th] 
  \caption{\mybox[lightgreen]{\textbf{FedPFN}} and \mybox[lightblue]{\textbf{FedPNU}}}
  \textbf{Input}: {clients $\{1,\dots,n\}$, mask ratio $\rho$, comm. round $T$, local epoch $E$, initial global model $\textbf{w}_{g}^{1}$, initial neuron mask $\mathbf{m}^{1}$;}\\
  \textbf{Output}: final global model $\textbf{w}_{g}^{T}$;
  \begin{algorithmic}[1]
    \FOR{each round $t=1,\dots, T$}
        \STATE \texttt{\# Client updates}
        \FOR{each client $i, i\in[n]$ \textbf{in parallel}}
        \STATE Receive global model $\textbf{w}_g^{t}$ and neuron mask $\mathbf{m}^{t}$;
        \STATE \coloredline{lightblue}{Compute the reverse mask $\mathbf{\hat{m}}^{t}$ of $\mathbf{m}^{t}$;}
        \STATE Set local model $\textbf{w}_i^{t} \leftarrow \textbf{w}_g^{t}$;
        \STATE Compute \mybox[lightgreen]{$E$} / \mybox[lightblue]{int($\frac{E}{2}$)} epochs of client local training by \autoref{equ:fl_pfn}:
        \STATE \coloredline{lightcyan}{\qquad$\mathbf{w}_{i}^{t} \leftarrow \mathbf{w}_{i}^{t}-\eta_l(\mathbf{m}^t\odot\mathbf{g}_i(\mathbf{w}_{i}^{t}))$;}
        \STATE \coloredline{lightblue}{{Compute $E$ - int($\frac{E}{2}$) epochs of client local training by \autoref{equ:fl_pfn}}:}
        \STATE \coloredline{lightblue}{\qquad$\mathbf{w}_{i}^{t} \leftarrow \mathbf{w}_{i}^{t}-\eta_l(\mathbf{\hat{m}}^t\odot\mathbf{g}_i(\mathbf{w}_{i}^{t}))$;}
        \ENDFOR
        \STATE \texttt{\# Server updates}
        \STATE The server samples $m$ clients and receive their models $\{\mathbf{w}_{i}^{t}\}_{i=1}^{m}$;
        \STATE Obtain the global model by FedAvg:
        \STATE \qquad $ \textbf{w}_g^{t+1} \leftarrow \sum_{i=1}^{m}\lambda_i\textbf{w}_{i}^{t}$, where $\lambda_i$ is the aggregation weight of client $i$;
        \STATE Randomly generate the new neuron mask $\mathbf{m}^{t+1}$ according to the ratio $\rho$.
    \ENDFOR
    \STATE Obtain the final global model $\textbf{w}_{g}^{T}$.
  \end{algorithmic}
\label{fedpfn_and_pnu_pseudo-code}
\end{algorithm}

\textit{FedPFN} (pseudo-code in \textcolor{lightgreen}{Algorithm \ref{fedpfn_and_pnu_pseudo-code}}). There are $T$ communication rounds in FL. During FL training, in communication round $t \in [T]$, the central server generates a new random mask $\mathbf{m}^t \in \{0,~1\}^d$ according to the masking ratio $\rho$. Also, the central server generates the global model $\mathbf{w}^t$ by the global aggregation scheme (e.g., FedAvg~\cite{mcmahan2017communication}) and sends $\mathbf{m}^t$ and $\mathbf{w}^t$ to the clients.

Clients initialize their local models as the received global model, $\mathbf{w}_i^t \leftarrow \mathbf{w}^t$. Client $i$ conducts SGD updates with mask $\mathbf{m}^t$, so that the masked neuron weights are fixed at this round. The SGD updates are as follows for $E$ epochs,
\begin{equation} \label{equ:fl_pfn}
    \mathbf{w}_{i}^{t} \leftarrow \mathbf{w}_{i}^{t}-\eta_l(\mathbf{m}^t\odot\mathbf{g}_i(\mathbf{w}_{i}^{t})),
\end{equation}
where $\odot$ denotes the element-wise (Hadamard) product and $\eta_l$ refers to the local learning rate. By applying FedPFN, during local training, all clients learn in the same effective permutation subspace so model drifts and permutation invariance issues can be relieved. Besides, the neuron mask $\mathbf{m}^t$ changes from round to round, so all the neurons can be evenly trained, and it will break the connectivity-accuracy tradeoff observed in LMC (\autoref{fig:lmc:pruning_mask_ratios}).

\textit{FedPNU} (pseudo-code in \textcolor{lightblue}{Algorithm \ref{fedpfn_and_pnu_pseudo-code}}). In FedPNU, we additionally consider a reversed mask $\hat{\mathbf{m}}^t$ of $\mathbf{m}^t$. During training, the clients first train with mask $\mathbf{m}^t$ according to \autoref{equ:fl_pfn} for the half local training, i.e., $\text{int}(\frac{E}{2})$ epochs; then, they train with the reversed mask $\hat{\mathbf{m}}^t$ for the remaining $E-\text{int}(\frac{E}{2})$ epochs. 
In FedPNU, the clients progressively train the networks in a subspace and the accordingly complementary subspace, by which the neurons of local models are more aligned. FedPNU progressively trains the two complementary subspaces, and it is less sensitive to hyperparameter $\rho$ (\autoref{fig:fl:mask_ratios}), reducing the efforts of hyperparameter tuning.

We note that FedPFN and FedPNU are lightweight and flexible since they only add a gradient mask before the optimizer's updates, so they are orthogonal to current FL algorithms (especially server-side global model fusion schemes). We will show they can be incorporated into existing FL methods for further improving performances. 

\subsection{Experiments} \label{subsec:fl_exp}

\textit{Settings and baselines.} We use three datasets to verify the algorithms: FashionMNIST~\cite{xiao2017fashion}, CIFAR-10, and CIFAR-100~\cite{krizhevsky2009learning}. We adopt the Dirichlet sampling to generate non-IID data (a.k.a., heterogeneous data) for each client, which is widely used in FL literature \cite{lin2020ensemble,chen2021bridging}. The Dirichlet sampling considers a class-imbalanced data heterogeneity, controlled by the hyperparameter ``dir'', the smaller, the more heterogeneous. We vary ``dir'' in range $[100, 0.5, 0.1]$, which respectively means IID, moderately non-IID, and extremely non-IID data. 
For FashionMNIST, the models are MLP\_h2\_w200 and LeNet5~\cite{lecun1998gradient}; for CIFAR-10 and CIFAR-100, the models are simple CNN~\cite{pmlr-v202-li23s} and ResNet20~\cite{he2016deep}. We consider the state-of-the-art FL methods as the baselines. For client-side methods, we consider vanilla training, FedProx~\cite{li2020federated}, FedDyn~\cite{acar2020federated}, FedRoD~\cite{chen2021bridging}, FedNH~\cite{dai2022tackling}, FedETF~\cite{Li_2023_ICCV}; for the server-side algorithms, we consider FedAvg~\cite{mcmahan2017communication} and FedDF~\cite{lin2020ensemble}. If not mentioned otherwise, the number of clients in the experiments is 20 and full client selection is applied. 
For more implementation details, please refer to the Appendix.

\textit{Different datasets and models.} In \autoref{table:fl:first_table}, we demonstrate the results under different datasets, data heterogeneity, and models. Since our methods are client-side algorithms, we note that the vanilla FedPFN/FedPNU are actually FedAvg+FedPFN/FedPNU, and we also combine our methods with the server-side approach FedDF, namely FedDF+FedPFN/FedPNU. The results show that FedPFN and FedPNU consistently improve over FedAvg, showing that incorporating the training-time alignment method can boost model fusion in FL. Also, our methods can strengthen FedDF to reach a higher performance.
Generally, our methods and the variants incorporating FedDF always achieve state-of-the-art results across various settings. 

\begin{figure*}[t]
\centering
    \begin{minipage}{0.47\linewidth}\centering
        \centering
        \includegraphics[width=0.78\linewidth]{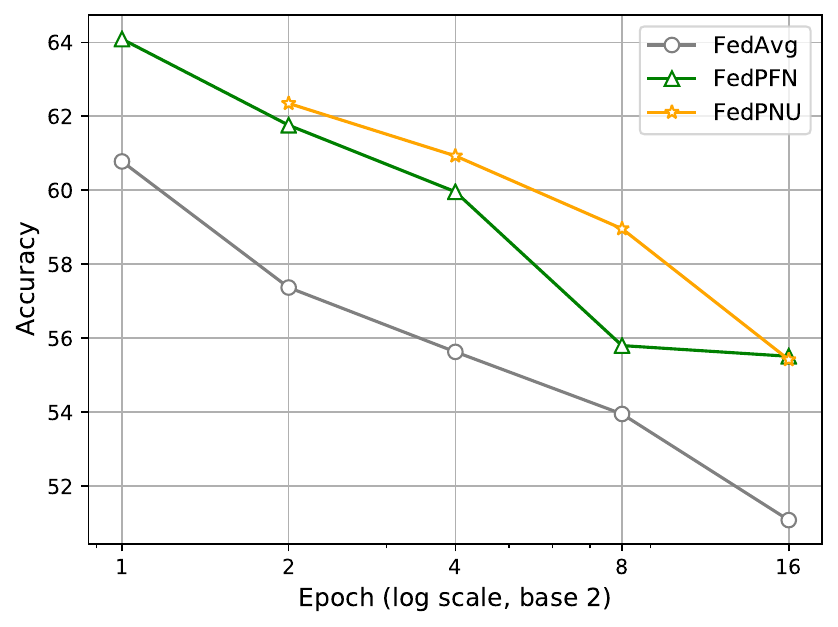}
    \caption{\textbf{Performances of FedPFN and FedPNU under different local epochs.} CIFAR-10 with dir = 0.1 and the model is CNN.}
    \label{fig:fl:local_epoch}
    \end{minipage}
    \hspace{0.1in}
    \begin{minipage}{0.45\linewidth}\centering
        \centering
    \includegraphics[width=0.8\linewidth]{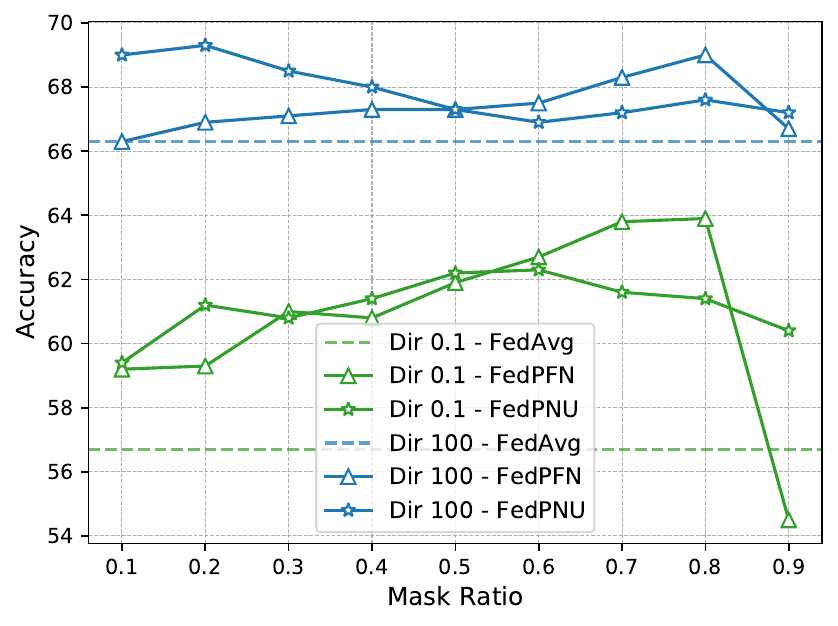}
    \caption{\textbf{Performances of FedPFN and FedPNU under different mask ratios.} CIFAR-10, CNN, and $E=3$.}
    \label{fig:fl:mask_ratios}
    \end{minipage}
\end{figure*}

\begin{table}[t]
\caption{\textbf{Results about different numbers of clients and partial selections.} CIFAR-10 with dir = 0.5, $E=3$, CNN.}
    \centering
\resizebox{0.95\columnwidth}{!}{
\begin{tabular}{l|ccccc}
\toprule
\multirow{2}{*}{\textbf{Methods}}&\multicolumn{5}{c}{\textbf{Number of clients (selection ratio)}}\\
\cmidrule{2-6}
 &30 (1.0) &60 (1.0) &90 (0.4) &90 (0.6) &90 (1.0) \\
\midrule
FedAvg &63.6±1.2 &62.5±0.7 &60.8±0.4 &61.4±0.7 &61.6±0.5\\
\midrule
\rowcolor{gray!20}FedPFN &\textbf{65.6±0.1} &\textbf{64.7±0.3} &\textbf{62.9±0.4} &\textbf{63.6±0.5} &\textbf{64.0±0.4} \\
\rowcolor{gray!20}FedPNU &65.2±0.2 &63.2±0.7 &62.0±0.9 &62.3±0.3 &62.3±0.7 \\
\bottomrule
\end{tabular}}
    \label{tab:fl:client_num}
\end{table}

\textit{Different number of clients.} We scale the number of clients in the range $[30, 60, 90]$ and apply partial client selections when the number is 90. From \autoref{tab:fl:client_num}, it is found that our methods can still improve the global model's generalization when scaling up the clients, which showcases the effectiveness of training-time neuron alignment methods in improving model fusion when the number of models is large.

\begin{table}[t]
\caption{\textbf{Results of random initialization pruning in FL and fixing FedPFN's mask.} CIFAR-10 with dir = 0.3 and $E=3$.}
    \centering
\resizebox{0.95\columnwidth}{!}{
\begin{tabular}{l|cccc}
\toprule
\textbf{Models\textbackslash Methods} &\textbf{FedAvg} &\textbf{FedPFN} & \textbf{FedPFN (fixed)} &\textbf{FedPruning} \\
\midrule
CNN      & 64.8±1.0 &\textbf{65.7±1.0} & 64.9±1.0 & 63.7±1.1 \\
ResNet20 & 72.0±0.7 &\textbf{72.4±0.5} &71.3±1.3 &70.2±1.3 \\
\bottomrule
\end{tabular}}
    \label{tab:fl:pruning_fixmask}
\end{table}

\textit{Different local epochs.} We verify the TNA variants in FL under different local epochs in \autoref{fig:fl:local_epoch}. We find that the improvements are also strong when there are more local updates. It is observed that FedPNU is more robust regarding local epochs, and this is because it learns in the complementary subspaces progressively, reducing the negative effects of subspaces on accuracy. Similar reasons are also for why FedPNU is robust when the mask ratio is as high as 0.9 in \autoref{fig:fl:mask_ratios}.

\textit{The effects of mask ratios for FedPFN and FedPNU.} From \autoref{fig:fl:mask_ratios}, it is shown that FedPFN benefits under smaller subspaces (higher mask ratios) but falls short when the subspace is too small (ratio $\rho=0.9$); whereas FedPNU is robust across all mask ratios due to its progressive learning.

\textit{Comparison with pruning and fixed masks.} We make an ablation study on the design of FedPFN. We compare FedPFN with the TNA-PFN variant denoted as FedPFN (fixed) in which we fix the neuron mask $\mathbf{m}^t$ in every round ($\mathbf{m}^t = \mathbf{m}^{t-1} = \mathbf{m}^0$). We also implement the setting where the random pruning is applied at initialization before FL training, named as FedPruning. \autoref{tab:fl:pruning_fixmask} presents the results. Although we find pruning can improve LMC in \autoref{sec:lmc}, it will cause generalization degradation in FL due to the connectivity-accuracy tradeoff. Also, if we incorporate TNA-PFN by keeping the same neuron mask during FL training, it will have marginal or even negative improvements. The above findings indicate that FL is sensitive in the subspaces and further validate the rationale of our devised methods.\looseness=-1

\textit{More experiments.} We include more results and illustrations in the Appendix.

\section{Conclusion}\label{sec:conclusion}
This paper focuses on training-time neuron alignment for improving model fusion performances. Starting from linear mode connectivity, it is hypothesized that by reducing potential permutation symmetries, the neurons between models can be better aligned. Then, a simple yet lossless algorithm called TNA-PFN is proposed. TNA-PFN uses frozen neuron weights as anchors to make every model training in a permutation subspace.
TNA-PFN is validated both theoretically and empirically in reducing the barriers of linear mode connectivity. We then use training-time alignment methods, TNA-PFN or its variants, in a wide range of model fusion applications. First, TNA-PFN can enhance the generalization of model soup in vision transformers and ColD fusion for pretrained language models. Second, in federated learning, where we propose two algorithms based on TNA-PFN, the proposed training-time alignment methods show effective performances in various settings, such as different data heterogeneity, different numbers of clients, different local epochs, etc. In a nutshell, this paper shows promising prospects for training-time alignment in model fusion, which are supported by theoretical analysis and extensive empirical results.

\section*{Acknowledgements}

This work was supported in part by the National Science and Technology Major Project (No.\ 2022ZD0115101), the Research Center for Industries of the Future (RCIF) at Westlake University, the Westlake Education Foundation, the National Key Research and Development Project of China (2021ZD0110505), the Zhejiang Provincial Key Research and Development Project (2023C01043), Academy Of Social Governance Zhejiang University, the National Natural Science Foundation of China (62441617), Zhejiang Provincial Natural Science Foundation of China (No. LD25F020001), and Fundamental Research Funds for the Central Universities (226-2025-00057).





\bibliographystyle{IEEEtran}
\bibliography{reference}

\begin{IEEEbiography}[{\includegraphics[width=1in,height=1.25in,clip,keepaspectratio]{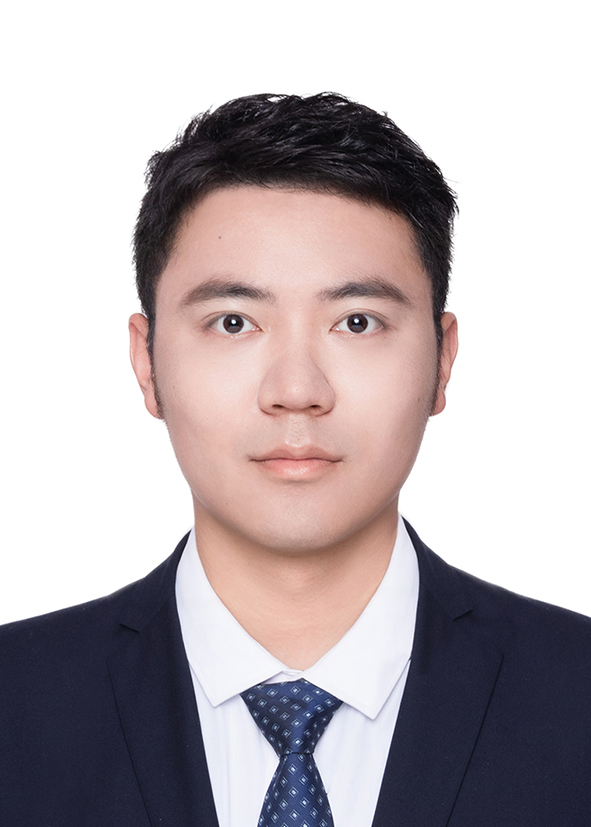}}]{Zexi~Li} is a research scientist at Tongyi Lab, Alibaba Group. He obtained his Ph.D. in Artificial Intelligence from Zhejiang University, and he was a visiting Ph.D. student at St John's College and CamMLSys Lab, the University of Cambridge. He received B.S. degree with honors from Zhejiang University. His main research interests include collaborative learning, federated learning, large language models, and LLM agent systems. He has published papers in top-tier conferences and journals, such as ICML, ICCV, NeurIPS, and Patterns, Cell Press.
\end{IEEEbiography}

\begin{IEEEbiography}[{\includegraphics[width=1in,height=1.25in,clip,keepaspectratio]{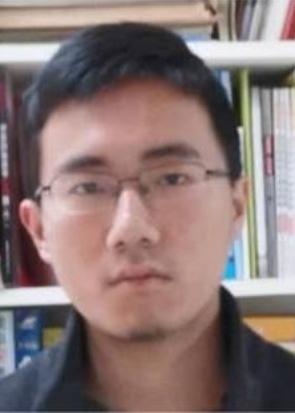}}]{Zhiqi Li} received the B.Eng. degree in Computer Science and Technology and the B.S. degree in Mathematics and Applied Mathematics from Zhejiang University, Hangzhou, in 2023. He is currently pursuing the M.S. degree at the Georgia Institute of Technology, Atlanta, USA. His research areas include federated learning and computer graphics.
\end{IEEEbiography}

\begin{IEEEbiography}[{\includegraphics[width=1in,height=1.25in,clip,keepaspectratio]{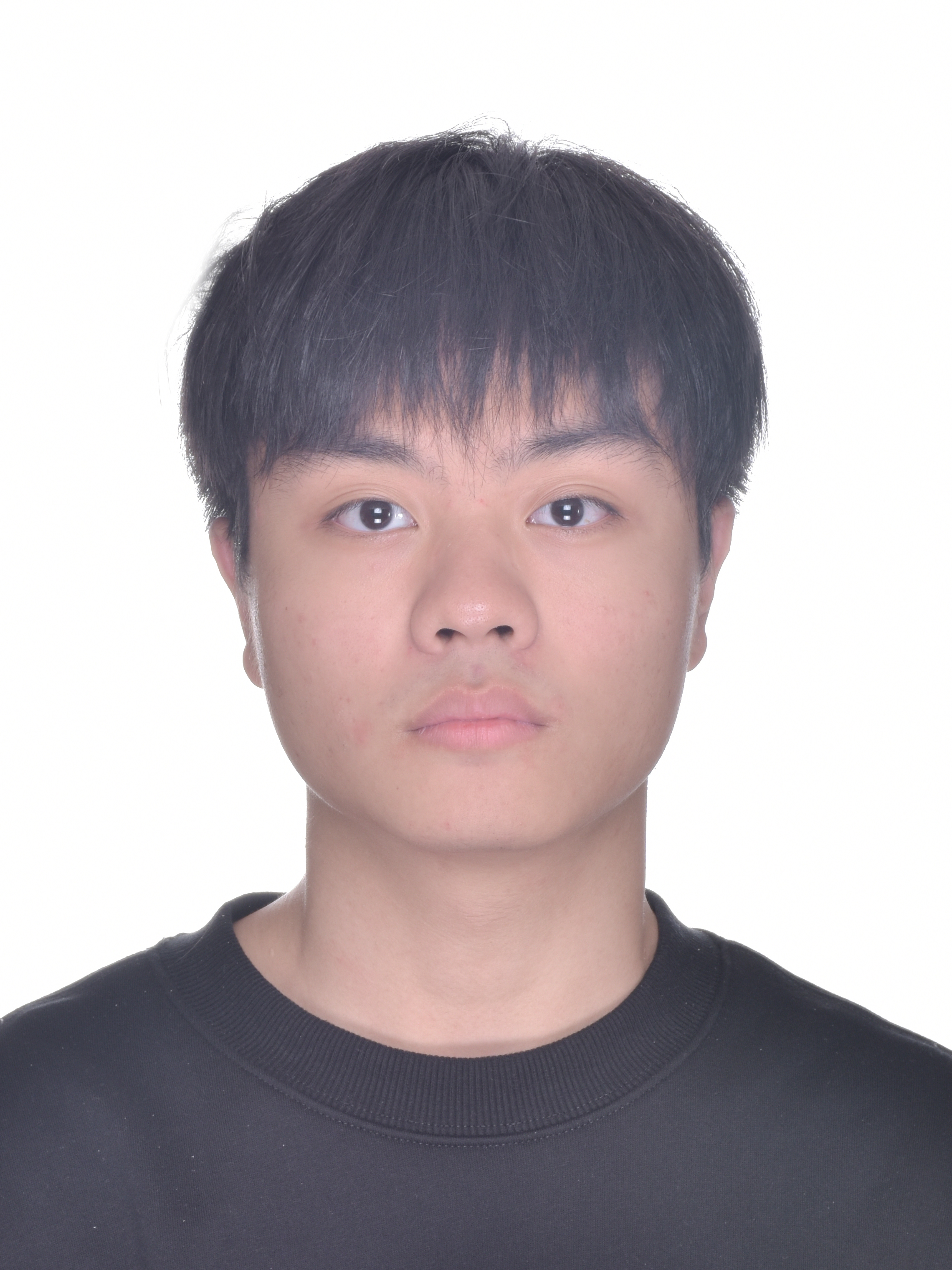}}]{Jie Lin} received the M.S. degree from Zhejiang University, Hangzhou, China. His research interests include federated learning and deep learning.
\end{IEEEbiography}

\begin{IEEEbiography}[{\includegraphics[width=1in,height=1.25in,clip,keepaspectratio]{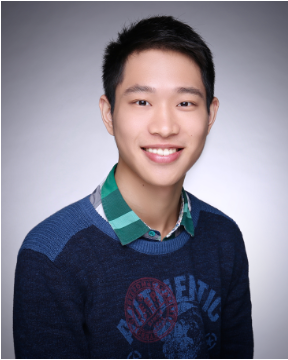}}]{Tao Shen} received the Ph.D. degree in Computer Science from Zhejiang University in 2025. His research interests span large-scale collaborative learning for large language models, distributed optimization under heterogeneous environments, and asynchronous as well as continual/lifelong learning for LLMs. His work focuses on developing efficient on-device training methods and designing heterogeneity-aware learning algorithms, making advanced AI accessible, efficient, and privacy-preserving for all.
\end{IEEEbiography}

\begin{IEEEbiography}[{\includegraphics[width=1in,height=1.25in,clip,keepaspectratio]{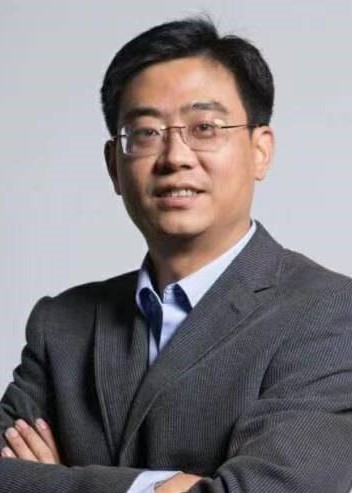}}]{Jun Xiao} received the Ph.D. degree in computer science and technology from the College of Computer Science, Zhejiang University, Hangzhou, China, in 2007. He is currently a professor with the College of Computer Science, Zhejiang University. His current research interests include computer vision, crossmedia understanding, and machine learning.
\end{IEEEbiography}

\begin{IEEEbiography}[{\includegraphics[width=1in,height=1.25in,clip,keepaspectratio]{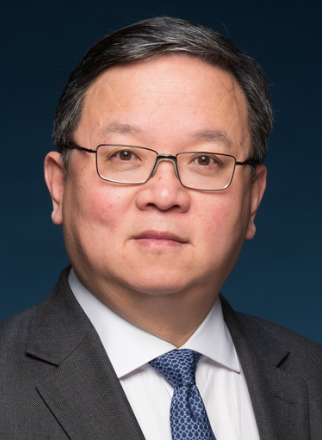}}]{Yike Guo} (Fellow, IEEE) is as the Provost of the Hong Kong University of Science and Technology (HKUST) and is concurrently a Chair Professor in the Department of Computer Science and Engineering of HKUST. 
He was the Founding Director of the Data Science Institute at Imperial College London. 
He is Fellow of IEEE, Fellow of Royal Academy of Engineering (FREng), a Member of Academia Europaea (MAE), Fellow of Hong Kong Academy of Engineering Sciences (FHKEng), Fellow of the Institute of Electrical and Electronics Engineers (FIEEE), Fellow of British Computer Society (FBCS), and Fellow of Chinese Association for Artificial Intelligence (FCAAI). His research interests lie in data mining, machine learning, and large-scale data management.
\end{IEEEbiography}

\begin{IEEEbiography}[{\includegraphics[width=1in,height=1.25in,clip,keepaspectratio]{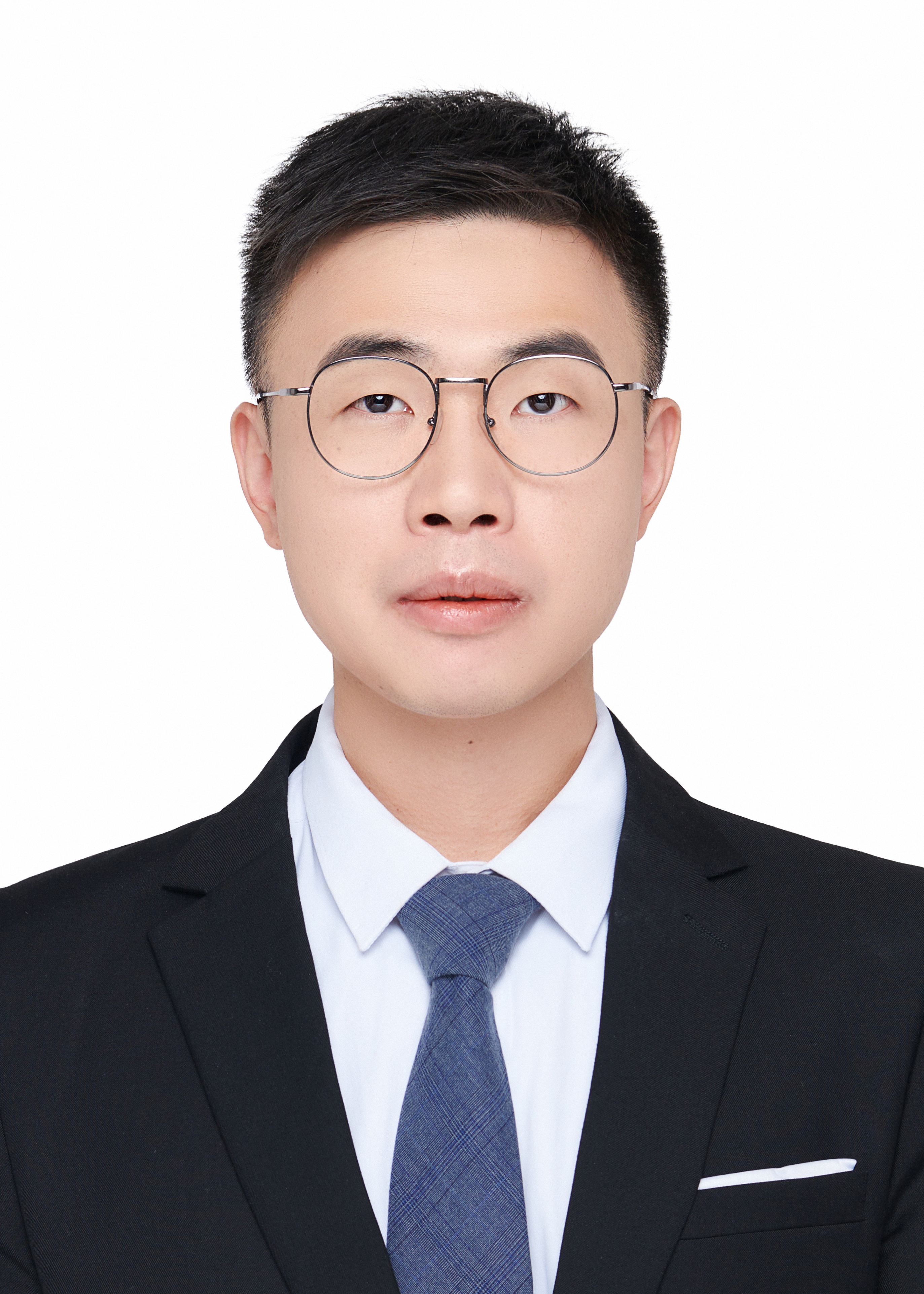}}]{Tao Lin} received the Ph.D. degree from EPFL, Switzerland, in 2022. He received a Master of Science degree from EPFL and a Bachelor of Engineering degree from Zhejiang University, in 2017 and 2014, respectively. 
He is currently a tenure-track assistant professor at Westlake University in China. His research focuses on deep learning and optimization, as well as its application in the distributed deep learning and inference system. He regularly published research papers at top-tier international machine learning conferences like ICML, NeurIPS, and ICLR, and some outcomes have integrated into PyTorch Distributed Training framework. He was a recipient of EPFL IC Outstanding Doctoral Thesis Award.
\end{IEEEbiography}

\begin{IEEEbiography}[{\includegraphics[width=1in,height=1.25in,clip,keepaspectratio]{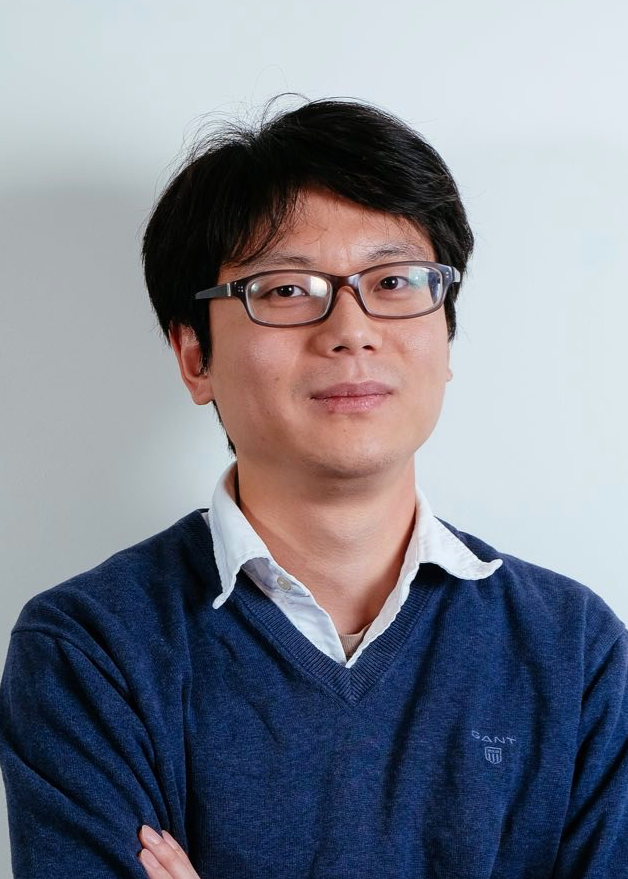}}]{Chao~Wu} is a tenure-track associate professor at the School of Public Affairs, Zhejiang University and the director of Computational Social Science Research Center in Zhejiang University. He is also an honorary Research Fellow at The Department of Computer Science, Imperial College London. His research interests include federated learning and distributed machine learning, data privacy protection and data pricing, and computational social sciences. He has published more than 100 papers in international conferences and journals and presided over many scientific research projects, including the key projects of the National Natural Science Foundation of China.
\end{IEEEbiography}

\newpage
\appendices
\onecolumn

\begin{center}
\Large
\textbf{Appendix}
\end{center}

In this appendix, we provide the details omitted in the main paper and more analyses and discussions.

\section{Implementation Details} \label{app_sec:implem_details}

\subsection{Datasets}
\textbf{MNIST}~\cite{lecun-mnisthandwrittendigit-2010} comprises a collection of 70,000 handwritten digits (0-9), divided into 60,000 training images and 10,000 testing images. Each image is grayscale and has a dimension of 28x28 pixels. \textbf{CIFAR-10}~\cite{krizhevsky2009learning} consists of 60,000 32x32 color images, evenly distributed across 10 different classes or labels, such as airplanes, automobiles, birds, cats, etc., each represented by 6,000 images. The dataset is split into 50,000 training images and 10,000 test images. \textbf{The polynomial approximation dataset}~\cite{pena2023re,von2019continual} is the synthetic dataset of the second and third polynomial functions: $y = 2x^2-1, y=(x-3)^3$. The input of the second polynomial function is uniformly generated from $x \in [-1.0, 1.0]$ with 100 data points, and the input of the third polynomial function is uniformly generated from $x \in [2.0, 4.0]$ with 100 data points. Each $y$ label in both the second and the third polynomial datasets is added by a random Gaussian noise with zero mean and 0.05 std. The \textbf{IMDb} (Internet Movie Database)~\cite{maas2011learning} dataset is a popular dataset used in Natural Language Processing (NLP) and sentiment analysis tasks. It consists of 50,000 movie reviews, evenly split into 25,000 reviews for training and 25,000 reviews for testing, each labeled as either positive or negative. \textbf{FashionMNIST}~\cite{xiao2017fashion} is a dataset designed as a more advanced replacement for the MNIST dataset, suitable for benchmarking machine learning models. It consists of 70,000 images divided into 60,000 training samples and 10,000 test samples. Each image is a 28x28 grayscale representation of fashion items from 10 different classes, including shirts, trousers, sneakers, etc. The \textbf{CIFAR-100} dataset~\cite{krizhevsky2009learning} is similar to the CIFAR-10 dataset but more challenging as it contains 100 different classes grouped into 20 superclasses. It contains 60,000 32x32 color images, with 600 images per class, divided into 50,000 training images and 10,000 test images. This dataset is primarily used for developing and evaluating more sophisticated image classification models.

\subsection{Models}
\textbf{CNN and MLP.} The simple CNN for CIFAR-10 and CIFAR-100 is a convolution neural network model with ReLU activations, which consists of 3 convolutional layers followed by 2 fully connected layers. The first convolutional layer is of size (3, 32, 3), followed by a max pooling layer of size (2, 2). The second and third convolutional layers are of sizes (32, 64, 3) and (64, 64, 3), respectively. The last two connected layers are of sizes (64*4*4, 64) and (64, num\_classes), respectively. 

The MLP model MLP\_h2\_w200 stands for an MLP with 2 hidden layers and a width of 200 in each layer. We vary h and w in Figure 4 to see the barriers in linear mode connectivity. We use MLP\_h2\_w200 for the MLP model in Table VII. 

\textbf{ResNets.} We followed the model architectures used in \cite{li2018visualizing}. The number of the model names means the number of layers of the models. Naturally, the larger number indicates a deeper network. For WRN56 in Figure 5, it is an abbreviation of Wide-ResNet56-4, where "4" refers to four times as many filters per layer. The ResNets used in Table VII are ResNet20 for CIFAR-10 and CIFAR-100. It is notable that since the batch normalization layers will have abnormal effects on model fusion~\cite{li2020fedbn,lin2020ensemble}, following~\cite{adilova2023layerwise}, we remove all the batch normalization layers from the ResNets.

\subsection{Randomness}
In all experiments, we implement the experiments three times with different random seeds and report the averaged results with standard deviations. 

For the experiments in linear mode connectivity, within a set of experiments, we generate an initial model according to the random seed $a$ and conduct training, then, we set the random seed as $a+1$ and load the initial model from random seed $a$ and conduct another independent training; afterward, the linear connectivity of the two models are tested.

For the experiments in federated learning. Given a random seed, we set torch, numpy, and random functions as the same random seed to make the data partitions and other settings identical. To make sure all algorithms have the same initial model, we save an initial model for each architecture and load the saved initial model at the beginning of one experiment. Also, for the experiments with partial participation, the participating clients in each round are vital in determining the model performance, and to guarantee fairness, we save the sequences of participating clients in each round and load the sequences in all experiments. This will make sure that, given a random seed and participation ratio, every algorithm will have the same sampled clients in each round.

\subsection{Evaluation} 
\textbf{Linear mode connectivity.} We validate all the accuracy and loss barriers on the test datasets to indicate the model generalization. 

\textbf{Federated learning.} We evaluate the global model performance on the test dataset of each dataset. The test dataset is mostly class-balanced and can reflect the global learning objective of a federated learning system. Therefore, the performance of the model on the test set can indicate the generalization performance of global models~\cite{pmlr-v202-li23s,lin2020ensemble}. In each experiment, we run 100 rounds and take the average test accuracy of the last 5 rounds as the final test accuracy. 

\subsection{Hyperparameter}
\textbf{Linear mode connectivity.} For CIFAR-10 and MNIST, We set a fixed learning rate of 0.1 and use the SGD optimizer with a weight decay of 5e-4 and momentum of 0.9; the number of learning epochs is 10. For the Polynomial datasets, the learning rate is 0.05 for 100 epochs. For the IMDb dataset, the learning rate is 0.0005 for 20 epochs. 

\textbf{Federated learning.} We set the initial learning rates as 0.08 in CIFAR-10 and FashionMNIST and set it as 0.05 in CIFAR-100. Following~\cite{pmlr-v202-li23s,chen2021bridging}, we set a decaying learning rate scheduler in all experiments; that is, in each round, the local learning rate is 0.99*(the learning rate of the last round). We set the weight decay factor as 5e-4. We set SGD optimizer as the clients' local solver and set momentum as 0.9. 

For the server-side optimizer FedDF, the server-side learning rate is 0.01 and the number of epochs is 20. We set $\mu = 0.001$ for FedProx.

\newpage

\section{Proof of Theorem III.4} \label{app_sec:proof}
We first recap the Theorem III.4 for convenience and provide the proof.
\begin{theorem} \label{app_thm:1}
We define a two-layer neural network with ReLU activation, and the function is $f_{\vv,\mU}(\vx)=\vv^\top\sigma(\mU\vx)$ where $\sigma(\cdot)$ is the ReLU activation function. $\vv\in \mathbb{R}^h$ and $\mU\in \sR^{h\times d}$ are parameters\footnote{For simplicity and without loss of generality, we omit the bias terms.} and $\vx \in \sR^d$ is the input which is taken from $\sX=\{\vx\in \sR^d| \Vert\vx\Vert_2<b\}$ uniformly. Consider two different networks parameterized with $\{\mU,\vv\}$ and $\{\mU',\vv'\}$ respectively, and for arbitrarily chosen masks $\mM_\vv \in \{0,1\}^h$ and $\mM_{\mU}\in \{0,1\}^{h\times d}$, each element of $\mU$ and $\mU'$, $\vv$ and $\vv'$ is i.i.d. sampled from a sub-Gaussian distribution $\mathcal{\text{sub-G}}(0,\sigma_\mU^2)$ and $\mathcal{\text{sub-G}}(0,\sigma_\vv^2)$ respectively with setting $\evv_{i}=\evv'_{i}$ when $\emM_{\vv,i}=0$ and $\emU_{i,j}=\emU'_{i,j}$ when $\emM_{\mU,ij}=0$. 
We consider the linear mode connectivity of the two networks and define the difference function between interpolated network and original networks as $z_{\vx}(\alpha)=(\alpha\vv + (1-\alpha)\vv')^\top\sigma((\alpha\mU+(1-\alpha)\mU')\vx)-\alpha\vv^\top\sigma(\mU\vx)-(1-\alpha){\vv'}^\top\sigma(\mU'\vx)$, $\alpha \in [0, 1]$. The function over all inputs is defined as $z(\alpha)=\frac{1}{|\sX|}\int_\sX z_{\vx}(\alpha)d\vx$. We use $\left|z(\alpha)\right|$, $\left|\frac{dz(\alpha)}{d\alpha}\right|$ and $\left|\frac{d^2z(\alpha)}{d\alpha^2}\right|$ to depict the linear mode connectivity, showing the output changes along the $\alpha$ path.  With probability $1-\delta$, it has,
\begin{align}
&\left|z(\alpha)\right|\le \sqrt{2} b\sigma_\vv\sigma_\mU\log(8h/\delta) \sqrt{h}\sqrt{1-\rho_\mU},\\
&\left|\frac{dz(\alpha)}{d\alpha}\right| \leq 4\sqrt{2} b\sigma_\vv\sigma_\mU \log{(24h/\delta)}\sqrt{h}(\sqrt{1-\rho_\vv}+\sqrt{1-\rho_\mU}),\\
&\left|\frac{d^2z(\alpha)}{d\alpha^2}\right| \leq 8 b\sigma_\vv \sigma_\mU \log(4h/\delta) \sqrt{h}\sqrt{(1-\max\{\rho_\mU,\rho_\vv\})},
\end{align}
where $\rho_\vv$ and $\rho_\mU$ refer to the mask ratios (the proportion of zeros in the mask) of masks $\mM_{\vv}$ and $\mM_{\mU}$ respectively. 
\end{theorem}

\textit{Proof:}
Let's first define $g_{\alpha}(\vx) = (\alpha\mU+(1-\alpha)\mU')\vx$. Then we can express $z_{\vx}(\alpha)$ as:

\begin{equation} \label{proof:equ1}
    z_{\vx}(\alpha) = (\alpha\vv + (1-\alpha)\vv')^\top \sigma(g_{\alpha}(\vx))-\alpha\vv^\top\sigma(\mU\vx)-(1-\alpha){\vv'}^\top\sigma(\mU'\vx).
\end{equation}

The first derivative of $z_{\vx}(\alpha)$ with respect to $\alpha$ will be:
\begin{equation} \label{proof:equ2}
\frac{dz_{\vx}(\alpha)}{d\alpha} = (\vv - \vv')^{\top}\sigma(g_{\alpha}(\vx)) + (\alpha\vv + (1 - \alpha)\vv')^{\top}\sigma'(g_{\alpha}(\vx))-\vv^\top\sigma(\mU\vx)+{\vv'}^\top\sigma(\mU'\vx).
\end{equation}

The second derivative with respect to $\alpha$ will be:
\begin{equation} \label{proof:equ3}
\frac{d^2z_{\vx}(\alpha)}{d\alpha^2} = 2(\vv - \vv')^{\top}\sigma'(g_{\alpha}(\vx)) + (\alpha\vv + (1-\alpha)\vv')^{\top}\sigma''(g_{\alpha}(\vx)).
\end{equation}

We also assume that the number of hidden neurons $h$ is sufficiently large for the convenience of analysis as \cite{entezari2021role} and we use $\#\{\mM_\mU=i\}$ and $\#\{\mM_\vv=i\}$ denote the number of $i$ in $\mM_\mU$ and $\mM_\vv$ respectively, $i=1,2$.  In the following proof, we will make use of Hoeffding's inequality for sub-Gaussian distributions. Here, we state it for reference: Let $X_1, \ldots, X_n$ be $n$ independent random variables such that $X_i \sim \operatorname{sub-G}\left(0, \sigma^2\right)$. Then for any $\va=(a_1,...,a_n) \in \mathbb{R}^n$, we have
$$
\mathbb{P}\left[|\sum_{i=1}^n a_i X_i| >t\right] \leq 2\exp \left(-\frac{t^2}{2 \sigma^2||a||_2^2}\right).
$$
\textbf{1)} For the 0-order difference equation, we have 
\begin{align}
    |z_{\vx}(\alpha)| &= \left| \alpha\vv^\top \left[\sigma(g_{\alpha}(\vx))-\sigma(\mU\vx)\right]+(1-\alpha){\vv'}^\top \left[\sigma(g_{\alpha}(\vx))-\sigma(\mU'\vx)\right]\right|\\
    &\leq \alpha\left| \vv^\top \left[(\sigma(g_{\alpha}(\vx))-\sigma(\mU\vx)\right]\right|+(1-\alpha)\left|{\vv'}^\top \left[\sigma(g_{\alpha}(\vx))-\sigma(\mU'\vx)\right]\right|.\label{proof:final00}
\end{align}
Then we bound the first term and the second term is bounded similarly due to symmetry.  For the \textbf{concentration upper bound} of the first term of \autoref{proof:final00}, we use the Hoeffding’s inequality for elements of $\vv$, with probability $1-\frac{\delta}{k}$ 
\begin{align}
    \alpha\left| \vv^\top \left[(\sigma(g_{\alpha}(\vx))-\sigma(\mU\vx)\right]\right|&\leq \alpha\sigma_v\sqrt{2\log(2k/\delta)}\Vert\sigma(g_\alpha(\vx))-\sigma(\mM\vx)\Vert_2\\
    &\leq \alpha\sigma_v\sqrt{2\log(2k/\delta)}\Vert g_\alpha(\vx)-\mM\vx\Vert_2\label{proof:final01}\\
    &= \alpha(1-\alpha)\sigma_v\sqrt{2\log(2k/\delta)}\Vert (\mU'-\mU)\vx\Vert_2.\label{proof:final011}
\end{align}
\autoref{proof:final01} is due to the fact that the ReLU activation function satisfies the Lipschitz continuous condition with constant $1$.  For the item $\Vert (\mU-\mU')\vx \Vert_2$, notice that $\emU_{ij}=\emU'_{ij}$ when $\emM_{\mU,ij}=0$, and then take a union bound, with probability $1-\frac{\delta}{k}$, we have 
\begin{align}
    \Vert(\mU-\mU')\vx\Vert_2&\leq \sqrt{\sum_{i=1}^h |[\mM_{\mU,i:}\odot (\mU_{i,:}-\mU'_{i,:})]\vx|^2}\\
    &=\sqrt{\sum_{i=1}^h |(\mU_{i,:}-\mU'_{i,:})(\mM_{\mU,i:}\odot \vx)|^2}\\
    &\le\sigma_\mU \sqrt{\sum_{i=1}^h \Vert \mM_{\mU,i:}\odot \vx \Vert_2^2}\sqrt{4\log(2hk/\delta)}.\label{proof:final02}
\end{align}
Then take a union bound choosing $k=4$ (because the union bound is taken for $4$ equations, \autoref{proof:final011} and \autoref{proof:final02} for the first and the second terms in \autoref{proof:final00} respectively.  Subsequent values of $k$ are determined with a similar method.), with probability $1-\delta$ we have
\begin{align}
    \left| z_\vx(\alpha)\right|<4\sqrt{2}\alpha(1-\alpha)\sigma_\vv\sigma_\mU\log(8h/\delta)\sqrt{\sum_{i=1}^h \Vert \mM_{\mU,i:}\odot \vx \Vert_2^2}.
\end{align}
Then integrate it on the region $\sX$.  With probability $1-\delta$, we have
\begin{align}
    \left| z(\alpha)\right|&\leq 4\sqrt{2}\alpha(1-\alpha)\sigma_\vv\sigma_\mU\log(8h/\delta)b\sqrt{\frac{d}{d+2}}\sqrt{h-\frac{\#\{\mM_\vv=0\}}{d}}\label{proof:final0}\\
    &\leq \sqrt{2} \sigma_\vv\sigma_\mU\log(8h/\delta)b \sqrt{h-\frac{\#\{\mM_\vv=0\}}{d}}\\
    &=\sqrt{2} \sigma_\vv\sigma_\mU\log(8h/\delta)b \sqrt{h}\sqrt{1-\rho_\mU}.
\end{align}

 \autoref{proof:final0} is due to fact that the integration $\frac{1}{|\sX|}\int_\sX \sqrt{\sum_{i=1}^h\Vert \mM_{\mU,i:}\odot \vx\Vert_2^2}   d\vx$ satisfies 
 \begin{align}
   \frac{1}{|\sX|}\int_\sX \sqrt{\sum_{i=1}^h\Vert \mM_{\mU,i:}\odot \vx\Vert_2^2}   d\vx  &\le \sqrt{(\frac{1}{|\sX|}\int_\sX \sum_{i=1}^h\Vert \mM_{\mU,i:}\odot \vx\Vert_2^2   d\vx )(\frac{1}{|\sX|}\int_\sX d\vx)}\label{proof:final11}\\
   &= \sqrt{\frac{1}{|\sX|}\int_\sX \#\{\mM_{\mU}=1\} \evx_i^2   d\vx} \label{proof:final12}\\
   &= \sqrt{\frac{\#\{\mM_{\mU}=1\}}{d}\frac{1}{|\sX|}\int_\sX  \Vert \vx \Vert_2^2   d\vx }\label{proof:final13}\\
   &= \sqrt{(h-\frac{\#\{\mM_{\mU}=0\}}{d})\frac{db^2}{d+2}},\label{proof:final14}
 \end{align}
where \autoref{proof:final11} is due to Cauchy-Schwarz inequality of integration, \autoref{proof:final12} and \autoref{proof:final13} is due to the symmetry of different components of $\vx$ and \autoref{proof:final14} is due to the integration $\frac{1}{|\sX|}\int_\sX \Vert \vx \Vert_2^k d\vx=\frac{db^k}{d+k}, k\in \mathbb{Z}$.

\textbf{2)} For the first derivative, we have
\begin{equation} \label{proof:equ4}
\left|\frac{dz_{\vx}(\alpha)}{d\alpha}\right| \!\leq\!  \left|(\vv \!-\! \vv')^{\top}\sigma(g_{\alpha}(\vx)) \right| \!+\! \left| (\alpha\vv \!+\! (1 - \alpha)\vv')^{\top}\sigma'(g_{\alpha}(\vx))\right|\!+\!|v^\top\sigma(\mU\vx)-{v'}^\top\sigma(\mU'\vx)|.
\end{equation}

\textbf{i)} For the \textbf{concentration upper bound} of the first term of \autoref{proof:equ4}, we use the Hoeffding's inequality for elements of $\vv-\vv'$  and notice that $\evv_i-\evv_i'=0$ when $\emM_{\vv,i}=0$, with probability $1-\frac{\delta}{k}$
\begin{align}
     \left|(\vv - \vv')^{\top}\sigma(g_{\alpha}(\vx)) \right| &\leq \sigma_\vv\sqrt{4\log(2k/\delta)}\Vert\mM_\vv\odot \sigma(g_{\alpha}(\vx))\Vert_2\\
     &\leq \sigma_\vv\sqrt{4\log(2k/\delta)}\Vert\mM_\vv\odot g_{\alpha}(\vx)\Vert_2\label{concentration2}\\
     &\leq \sigma_\vv\sqrt{4\log(2k/\delta)} (\alpha\Vert  \mM_\vv\odot\mU \vx\Vert_2+(1-\alpha)\Vert\mM_\vv\odot\mU' \vx \Vert_2).
\end{align}
\autoref{concentration2} is due to the property of ReLU activation function that $|\sigma(x)|<|x|$.  The Hoffding's inequality is used again for each row $i$ of matrix $\mU$ and $\mU'$ with $\emM_{\vv,i}=1$, and after taking a union bound, we have the following inequality with probability $1-\frac{\delta}{k}$,
\begin{align}
    \Vert \mM_{\vv}\odot \mU \vx\Vert_2 &= \sqrt{\sum_{\emM_{\vv,i}=1} |\mU_{i,:} \vx|^2}\\
    &\leq\sigma_\mU\sqrt{2(h-\#\{\mM_{\vv}=0\})\log(2hk/\delta)}\Vert \vx \Vert_2. \label{concentration1}
\end{align}
$\Vert \mM_{\vv}\odot \mU' \vx\Vert_2$ can be calculated similarly to \autoref{concentration1}.  Then after taking a union bound, with $1- \frac{\delta}{k} $ the first term is bounded as 
\begin{align}
     \left|(\vv - \vv')^{\top}\sigma(g_{\alpha}(\vx)) \right| &\leq 2\sqrt{2}\sqrt{h-\#\{\mM_\vv=0\}}\sigma_\vv\sigma_\mU\log(6hk/\delta)\Vert \vx \Vert_2.
\end{align}

\textbf{ii)} For the \textbf{concentration upper bound} of the second term of \autoref{proof:equ4}, we use the Hoeffeding's inequality for each element of $\vv$ and $\vv'$ and take a union bound, with probability $1-\frac{\delta}{k}$ we have the following inequality,
\begin{align}
    &\left| (\alpha\vv + (1 - \alpha)\vv')^{\top}\sigma'(g_{\alpha}(\vx))\right| \nonumber\\
    &= \left| (\alpha\vv + (1 - \alpha)\vv')^{\top}\sigma'(\vy)|_{\vy=g_{\alpha}(\vx)}\odot (\mU-\mU')\vx)\right|\label{proof:concentration25}\\
    &\leq \sqrt{\alpha^2 + (1-\alpha)^2}\sigma_\vv \sqrt{2log(2k/\delta)} \Vert \sigma'(\vy)|_{\vy=g_{\alpha}(\vx)}\odot(\mU-\mU')\vx) \Vert_2 \label{proof:concentration3}\\
    &\leq \sigma_\vv\sqrt{log(2k/\delta)}  \Vert (\mU-\mU')\vx \Vert_2.
\end{align}
\autoref{proof:concentration25} is due to the chain rule of differentiation and \autoref{proof:concentration3} is due to the fact that the property $|\sigma'(\cdot)|<1$ of the ReLU activation function.  The term $\Vert (\mU-\mU')\vx \Vert_2\le \sigma_\mU \sqrt{\sum_{i=1}^h \Vert \mM_{\mU,i:}\odot \vx \Vert_2^2}\sqrt{4\log(2hk/\delta)}$ is obtained in \autoref{proof:final02}.  Then with $1-\frac{\delta}{k}$ after taking a union bound, the second term is bounded as 
\begin{align}
    |(\alpha \vv+(1-\alpha)\vv')^\top\sigma'(g_\alpha(\vx))&\le 2\log(4hk/\delta)\sigma_\vv \sigma_\mU \sqrt{\sum_{i=1}^h\Vert \mM_{\mU,i:}\odot \vx\Vert_2^2}.
\end{align}

\textbf{iii)} For the \textbf{concentration upper bound} of the third term of \autoref{proof:equ4}, first write it as 
\begin{align}
    \left| \vv^\top \sigma(\mU\vx)-{\vv'}^\top \sigma(\mU'\vx)\right|&=\left| \vv^\top \sigma(\mU\vx)-\vv^\top \sigma(\mU'\vx)+\vv^\top \sigma(\mU'\vx)-{\vv'}^\top \sigma(\mU'\vx)\right|\\
    &\le \left| \vv^\top \left[\sigma(\mU\vx)-\sigma(\mU'\vx)\right]\right|+\left|(\vv -{\vv'})^\top \sigma(\mU'\vx)\right|.
\end{align}
Then we use the Hoeffeding's inequality for each element of $\vv$ and $\vv'$ and notice that $\evv_i-\evv_i'=0$ when $\emM_{\vv,i}=0$.  After taking a union bound, with probability $1-\frac{\delta}{k}$ we have the following inequality,
\begin{align}
    &\left| \vv^\top \sigma(\mU\vx)-{\vv'}^\top \sigma(\mU'\vx)\right|\nonumber\\
    &\le \sigma_\vv\sqrt{2\log(4k/\delta)} \Vert \sigma(\mU\vx)-\sigma(\mU'\vx) \Vert_2+\sigma_\vv\sqrt{4\log(4k/\delta)}\Vert\mM_\vv\odot \sigma(\mU'\vx)\Vert_2\\
    &\le \sigma_\vv\sqrt{2\log(4k/\delta)} \Vert (\mU-\mU')\vx) \Vert_2+\sigma_\vv\sqrt{4\log(4k/\delta)}\Vert\mM_\vv\odot \mU'\vx\Vert_2. \label{proof:final15}
\end{align}
\autoref{proof:final15} is due to the fact the ReLU activation function $\sigma(\cdot)$ satisfied the Lipschitz continuity condition with constant $1$ and $|\sigma(x)| \le |x|$.  The term $\Vert (\mU-\mU')\vx) \Vert_2\leq \sigma_\mU \sqrt{\sum_{i=1}^h \Vert \mM_{\mU,i:}\odot \vx \Vert_2^2}\sqrt{4\log(2hk/\delta)}$ in \autoref{proof:final15} can be calculated as in \autoref{proof:final02} with probability $1-\frac{\delta}{k}$ and the term $\Vert\mM_\vv\odot \mU'\vx\Vert_2\leq \sigma_\mU\sqrt{2(h-\#\{\mM_{\vv}=0\})\log(2hk/\delta)}\Vert \vx \Vert_2$ can be caluclated as in \autoref{concentration1} with probability $1-\frac{\delta}{k}$.  Then take the union bound, with probability $1-\frac{\delta}{k}$ we have
\begin{align}
     &\left| \vv^\top \sigma(\mU\vx)-{\vv'}^\top \sigma(\mU'\vx)\right|\nonumber\\
     &\le \sigma_\vv\sigma_\mU\log(8kh/\delta)(2\sqrt{2}\sqrt{\sum_{i=1}^h \Vert \mM_{\mU,i:}\odot \vx \Vert_2^2}+2\sqrt{2}\sqrt{h-\#\{\mM_\vv=0\}}\Vert \vx \Vert_2).
\end{align}

In conjunction with analyses \textbf{i)},\textbf{ii)} and \textbf{iii)} and take a union bound choosing $k=3$, we have with probability $1-\delta$,
\begin{align}
    \left|\frac{dz_{\vx}(\alpha)}{d\alpha}\right| &\leq \sqrt{h-\#\{\mM_\vv=0\}}2\sqrt{2}\sigma_\vv\sigma_\mU\log(18h/\delta)\Vert \vx \Vert_2\nonumber\\
    &+2log(12h/\delta)\sigma_\vv\sigma_\mU \sqrt{\sum_{i=1}^h\Vert \mM_{\mU,i:}\odot \vx\Vert_2^2}\nonumber\\
    &+\sigma_\vv\sigma_\mU\log(24h/\delta)(2\sqrt{2}\sqrt{\sum_{i=1}^h \Vert \mM_{\mU,i:}\odot \vx \Vert_2^2}+2\sqrt{2}\sqrt{h-\#\{\mM_\vv=0\}}\Vert \vx \Vert_2)\\
    &\leq \sqrt{h-\#\{\mM_\vv=0\}}4\sqrt{2}\sigma_\vv\sigma_\mU\log(24h/\delta)\Vert \vx \Vert_2\nonumber\\
    &+4\sqrt{2}log(24h/\delta)\sigma_\vv\sigma_\mU \sqrt{\sum_{i=1}^h\Vert \mM_{\mU,i:}\odot \vx\Vert_2^2}.
\end{align}
Then integrate them on the region $\sX$.  With probability $1-\delta$ we have
\begin{align}
    \left|\frac{dz(\alpha)}{d\alpha}\right| &\leq \sqrt{h-\#\{\mM_\vv=0\}}4\sqrt{2}\sigma_\vv\sigma_\mU\log(24h/\delta)\frac{db}{d+1}\nonumber\\
&+4\sqrt{2}\sigma_\vv\sigma_\mU\log(24h/\delta)\sigma_\vv\sigma_\mU b\sqrt{\frac{d}{d+2}}\sqrt{h-\frac{\#\{\mM_\mU=0\}}{d}} \label{proof:final1}\\
    &\leq 4\sqrt{2} b\sigma_\vv\sigma_\mU \log{(24h/\delta)}(\sqrt{h-\#\{\mM_\vv=0\}}+\sqrt{h-\frac{\#\{\mM_\mU=0\}}{d}})\\
    &=4\sqrt{2} b\sigma_\vv\sigma_\mU \sqrt{h}\log{(24h/\delta)}(\sqrt{1-\rho_\vv}+\sqrt{1-\rho_\mM}).
\end{align}
\autoref{proof:final1} is due to the integration $\frac{1}{|\sX|}\int_\sX \Vert \vx \Vert_2 d\vx=\frac{db}{d+1}$ and $\frac{1}{|\sX|}\int_\sX \sqrt{\sum_{i=1}^h\Vert \mM_{\mU,i:}\odot \vx\Vert_2^2}   d\vx\leq \sqrt{(h-\frac{\#\{\mM_{\mU}=0\}}{d})\frac{db^2}{d+2}}$ from \autoref{proof:final14}

\textbf{3)} For the second derivative, we have
\begin{equation} \label{proof:equ5}
\left|\frac{d^2z_{\vx}(\alpha)}{d\alpha^2}\right| \leq  2\left|(\vv - \vv')^{\top}\sigma'(g_{\alpha}(\vx)) \right| + \left| (\alpha\vv + (1 - \alpha)\vv')^{\top}\sigma''(g_{\alpha}(\vx))\right|.
\end{equation}

\textbf{i)} For the \textbf{concentration upper bound} of the first term of \autoref{proof:equ5}, we use the Hoeffding's inequality for each element of $\vv-\vv'$  and notice that $\evv_i-\evv_i'=0$ when $\emM_{\vv,i}=0$, with probability $1-\frac{\delta}{k}$, we have
\begin{align}
     2\left|(\vv - \vv')^{\top}\sigma'(g_{\alpha}(\vx)) \right| 
     &= 2\left|(\vv - \vv')^{\top}\sigma'(\vy)|_{\vy=g_{\alpha}(\vx)}\odot (\mU-\mU')\vx \right| \label{proof:order2:equation1}\\
     &= 2\left|(\vv - \vv')^{\top}\mM_\vv\odot\sigma'(\vy)|_{\vy=g_{\alpha}(\vx)}\odot (\mU-\mU')\vx \right| \label{proof:order2:equation2}\\
     &\leq 4\sigma_\vv\sqrt{\log(2k/\delta)}\Vert \mM_\vv\odot \sigma'(\vy)|_{\vy=g_{\alpha}(\vx)}\odot (\mU-\mU')\vx\Vert_2 \label{proof:order2:equation3}\\
     &\leq 4\sigma_\vv\sqrt{\log(2k/\delta)}\Vert \mM_\vv\odot (\mU-\mU')\vx\Vert_2 \label{proof:order2:equation4}.
\end{align}

\autoref{proof:order2:equation1} is due to the chaine rule of differentiation, \autoref{proof:order2:equation2} is due to $\evv_i=\evv'_i$ when $\emM_{\vv,i}=0$,
\autoref{proof:order2:equation3} is due to Hoeffding's inequation and \autoref{proof:order2:equation4} is due to the property $|\sigma'(x)|<1$ of the ReLU activation function.  For the item $\Vert \mM_\vv\odot(\mU-\mU')\vx \Vert_2$, notice that $\emU_{ij}=\emU'_{ij}$ when $\emM_{\mU,ij}=0$ and take a union bound with probability $1-\frac{\delta}{k}$, we have
\begin{align}
    \Vert \mM_\vv \odot (\mU-\mU')\vx\Vert_2&\leq \sqrt{\sum_{\emM_{\vv,i}=1}|[\mM_{\mU,i:}\odot (\mU_{i,:}-\mU'_{i,:})]\vx|^2}\\
    &\leq \sqrt{\sum_{\emM_{\vv,i}=1}|(\mU_{i,:}-\mU'_{i,:})(\mM_{\mU,i:}\odot\vx)|^2}\\
    &\leq \sigma_\mU \sqrt{\sum_{\emM_{\vv,i}=1}\Vert \mM_{\mU,i:}\odot\vx \Vert_2^2} \sqrt{4\log(2hk/\delta)}.
\end{align}
Then with $1-\frac{\delta}{k}$ after taking a union bound, the first term is bounded as 
\begin{align}
    2\left|(\vv - \vv')^{\top}\sigma'(g_{\alpha}(\vx)) \right| \le 8 \sigma_\vv \sigma_\mU \log(4hk/\delta) \sqrt{\sum_{\emM_{\vv,i}=1}\Vert \mM_{\mU,i:}\odot\vx \Vert_2^2}.
\end{align}

\textbf{ii)} For the \textbf{concentration upper bound} of the second term of \autoref{proof:equ5}, note that property $\sigma''(x)=0$ of ReLU activation function, then 
\begin{align}
     &\left| (\alpha\vv + (1 - \alpha)\vv')^{\top}\sigma''(g_{\alpha}(\vx))\right|\nonumber\\
     &=\left| (\alpha\vv + (1 - \alpha)\vv')^{\top}\sigma''(\vy)|_{\vy=g_{\alpha}(\vx)}\odot(\mU-\mU')\vx \odot (\mU-\mU')\vx \right|\\
     &=0.
\end{align}
In conjunction with analyses \textbf{i)} and \textbf{ii)} and take a union bound choosing $k=1$, with probability $1-\delta$ we have 
\begin{align}
    \left|\frac{d^2z_{\vx}(\alpha)}{d\alpha^2}\right| &\leq 8 \sigma_\vv \sigma_\mU \log(4h/\delta) \sqrt{\sum_{\emM_{\vv,i}=1}\Vert \mM_{\mU,i:}\odot\vx \Vert_2^2}.
\end{align}
    
Then integrate them on the region $\sX$.  With probability $1-\delta$ we have
\begin{align}
    \left|\frac{d^2z(\alpha)}{d\alpha^2}\right| 
    &\leq 8\sigma_\vv \sigma_\mU \log(4h/\delta) \sqrt{(h-\frac{\max\{\#\{\mM_{\mU}=0\},d\#\{\mM_{\vv}=0\}\}}{d})\frac{db^2}{d+2}}\label{proof:final2}\\
    &\leq 8\sigma_\vv \sigma_\mU \log(4h/\delta) b\sqrt{(h-\frac{\max\{hd\rho_\mU,hd\rho_\vv\}}{d})}\\
    &\leq 8\sigma_\vv \sigma_\mU \log(4h/\delta) b\sqrt{h}\sqrt{(1-\max\{\rho_\mU,\rho_\vv\})}.
\end{align}

 \autoref{proof:final2} is due to the integration $\frac{1}{|\sX|}\int_\sX \sqrt{\sum_{i=1}^h\Vert \mM_{\mU,i:}\odot \vx\Vert_2^2}   d\vx$ satisfying
 \begin{align}
   \frac{1}{|\sX|}\int_\sX \sqrt{\sum_{\emM_{\vv,i}=1}\Vert \mM_{\mU,i:}\odot\vx \Vert_2^2}   d\vx  &\le \sqrt{(\frac{1}{|\sX|}\int_\sX \sum_{\emM_{\vv,i}=1}\Vert \mM_{\mU,i:}\odot\vx \Vert_2^2  d\vx )(\frac{1}{|\sX|}\int_\sX d\vx)}\label{proof:final21}\\
   &= \sqrt{\frac{1}{|\sX|}\int_\sX \#\{\mM_{\mU}\odot \mM_{\mV}=1\} x_i^2   d\vx} \label{proof:final22}\\
   &= \sqrt{\frac{\#\{\mM_{\mU} \odot \mM_{\mV}=1\}}{d}\frac{1}{|\sX|}\int_\sX  \Vert x \Vert_2^2   d\vx }\label{proof:final23}\\
   &\leq \sqrt{(h-\frac{\max\{\#\{\mM_{\mU}=0\},d\#\{\mM_{\vv}=0\}\}}{d})\frac{db^2}{d+2}}\label{proof:final24},
 \end{align}
where $\mM_{\mV}$ is the matrix whose each column is $\mM_\vv$. \autoref{proof:final21} is due to Cauchy-Schwarz inequality of integration, \autoref{proof:final22} and \autoref{proof:final23} is due to the symmetry of different components of $\vx$ and \autoref{proof:final24} is due to the integration $\frac{1}{|\sX|}\int_\sX \Vert \vx \Vert_2^2 d\vx=\frac{db^2}{d+2}$ and $\#\{\mM_{\mU} \odot \mM_{\mV}=1\}\leq\min\{\#\{\mM_{\mU}=1\},\#\{\mM_{\mV}=1\}\}=\min\{\#\{\mM_{\mU}=1\},d\#\{\mM_{\vv}=1\}\}$.
$\hfill \square$

\section{More Analysis and Results} \label{app_sec:more_results}


\subsection{More Results and Illustrations in Linear Mode Connectivity}\label{app_subsec:lmc}

\begin{figure}[htbp] 
    \centering
    \begin{subfigure}{0.7\textwidth}
        \centering
        \includegraphics[width=\linewidth]{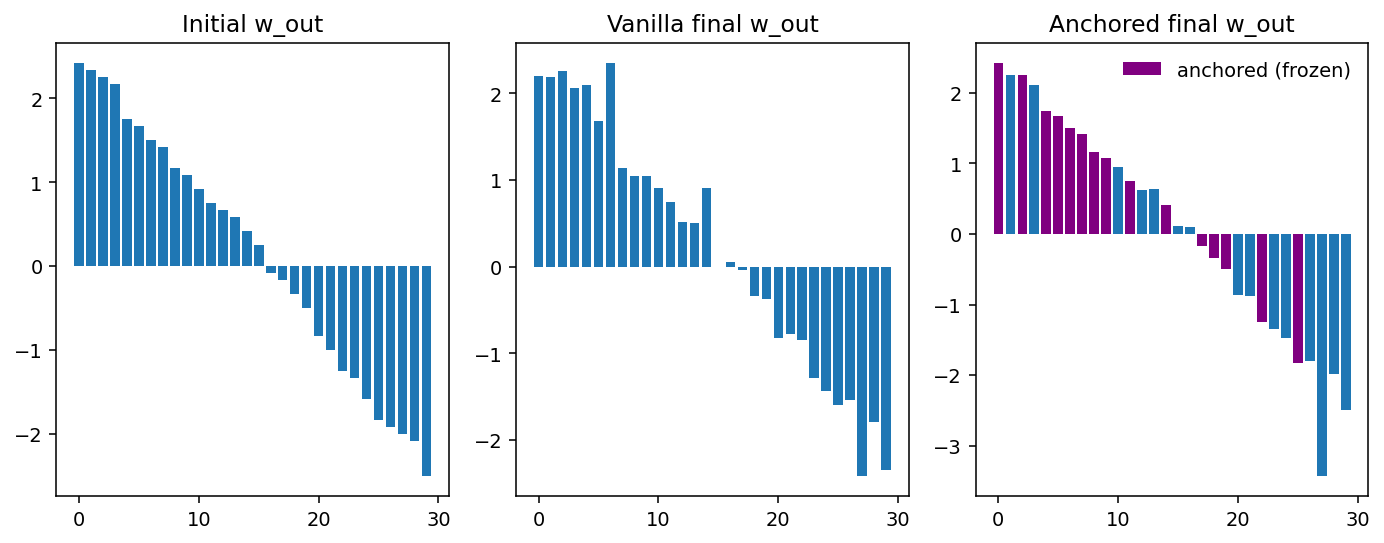}
        \caption{Trial 1}
        \label{fig:trial1}
    \end{subfigure}
    \hfill 
    \begin{subfigure}{0.7\textwidth}
        \centering
        \includegraphics[width=\linewidth]{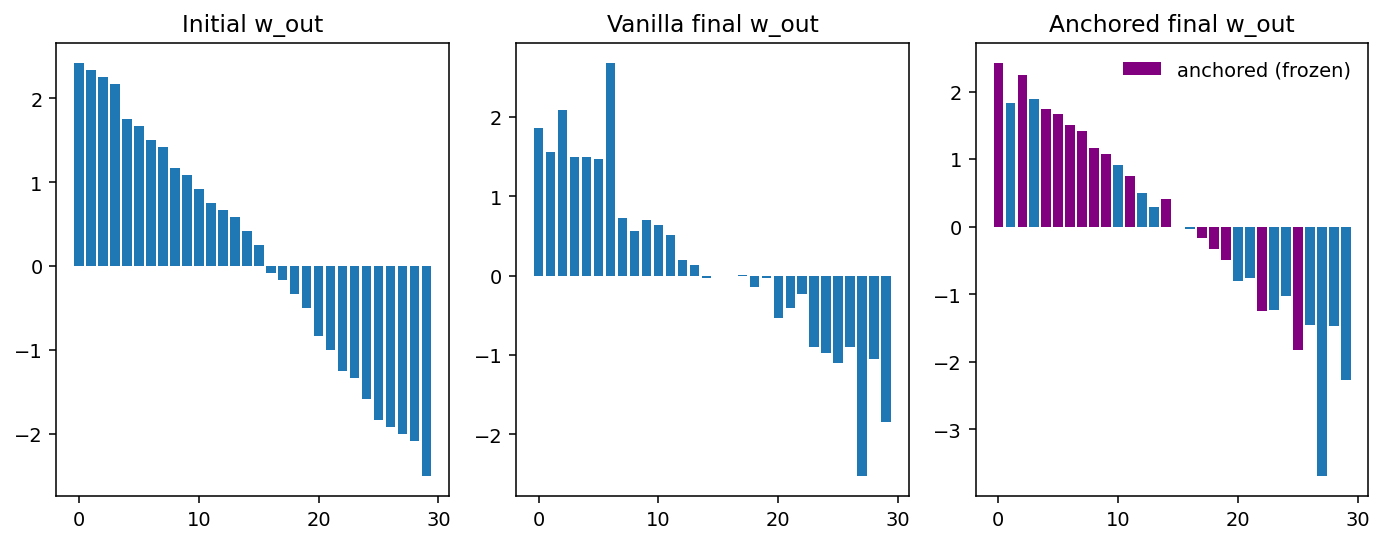}
        \caption{Trial 2}
        \label{fig:trial2}
    \end{subfigure}
    \caption{\textbf{Understanding permutation symmetry: position-wise output layer weight distributions.} \emph{Setting:} one-hidden-layer MLP with \(H{=}30\) on synthetic quadratic regression \(y=2x^2-1\) for \(x\in[-1,1]\); output-layer weights initialized from \(\mathrm{Unif}[-2.5,2.5]\) and sorted in descending order; train for 20 epochs. We compare vanilla training and TNA-PFN with anchor ratio \(\rho{=}0.5\) (anchored). \textbf{(a):} lower-noise, smaller learning rate (Trial~1). \textbf{(b):} higher-noise, larger learning rate (Trial~2). Purple bars indicate anchored (frozen) neurons. \emph{Findings:} vanilla training substantially reshuffles the order of neuron weights, while TNA-PFN preserves the initial ordering much better, including for non-anchored neurons. Ranking-change fractions: Trial~1—vanilla \(0.733\) vs.\ TNA-PFN \(0.233\); Trial~2—vanilla \(0.800\) vs.\ TNA-PFN \(0.467\).}
    \label{fig:vis_perm_sym_1}
\end{figure}

\begin{figure}
    \centering
    \includegraphics[width=0.4\linewidth]{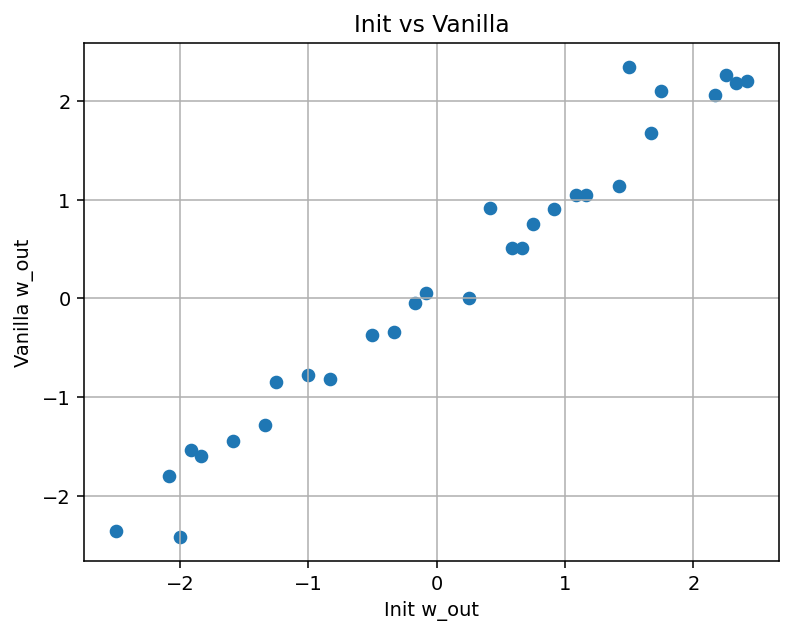}
    \includegraphics[width=0.4\linewidth]{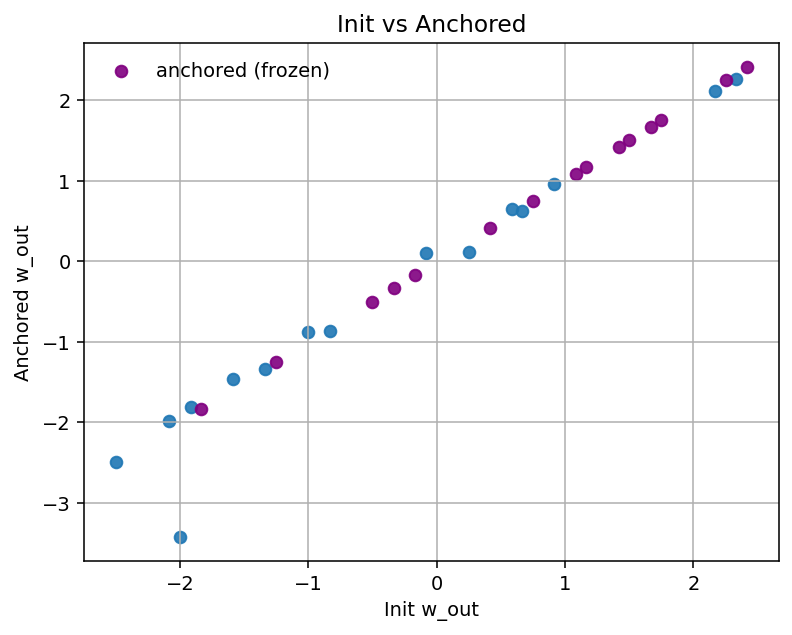}
    \caption{\textbf{Understanding permutation symmetry: visualization of initialized and final weights.} \emph{Setting:} same as Figure~2, trial 1. Left: vanilla training (initial weight on \(x\)-axis, final weight on \(y\)-axis). Right: TNA-PFN (anchored neurons highlighted in purple). \emph{Findings:} under vanilla training, points spread widely away from the diagonal, indicating large deviations from the initial ordering. Under TNA-PFN, points cluster near the diagonal even for the non-anchored neuron weights. }
    \label{fig:vis_perm_sym_2}
\end{figure}

\textbf{Permutation symmetry: an intuitive example.} To provide neuron-level evidence, we designed a demonstration example that showcases and visualizes permutation symmetry during standard training and the permutation asymmetry encouraged by TNA-PFN. Results are shown in \autoref{fig:vis_perm_sym_1} and \autoref{fig:vis_perm_sym_2}.

\begin{itemize}
\item \textbf{Task and model.} We use a one-hidden-layer MLP (\(H{=}30\) ReLU units) on a simple quadratic regression task \(y = 2x^2 - 1\) for \(x \in [-1,1]\). We split the data into 80\% training and 20\% validation. Output-layer weights are initialized i.i.d.\ from \(\mathrm{Unif}[-2.5, 2.5]\) (where weight norm distributions are more consistent with the final, so most likely, only the weight permutation will take place), and then \emph{sorted in descending order} so that neuron indices coincide with the initial rank (1 = largest weight).

\item \textbf{Training protocols.} We compare (i) vanilla training and (ii) our anchored training, TNA-PFN, with anchor ratio \(\rho = 0.5\). We train for 20 epochs under two conditions: \emph{Trial~1} (lower noise, smaller learning rate) and \emph{Trial~2} (higher noise, larger learning rate; same initialization).

\item \textbf{What we visualize.}  
  \begin{itemize}
  \item \emph{Position-wise bar plots} (Figure~\ref{fig:vis_perm_sym_1}): initial ordered weights, final weights after vanilla training, and final weights after TNA-PFN (anchored neurons shown in purple).  
  \item \emph{Scatter plots} (Figure~\ref{fig:vis_perm_sym_2}): initial weight on the \(x\)-axis, final weight on the \(y\)-axis. Left: vanilla; right: TNA-PFN (anchored neurons highlighted in purple). A tighter near-diagonal cloud means better preservation of the initial order.
  \end{itemize}

\item \textbf{Ranking-change metric.} To quantify permutation, we compute the \emph{ranking-change fraction}: after training, we re-rank neurons by their output layer weights and measure the fraction of indices whose ranks differ from the initialization. A value near 1 indicates strong permutation (heavy reshuffling); a smaller value means the learned solution preserves the initial ordering better.
  \begin{itemize}
  \item \emph{Trial~1 (lower noise)}: vanilla = \textbf{0.733}; TNA-PFN = \textbf{0.233}.  
  \item \emph{Trial~2 (higher noise)}: vanilla = \textbf{0.800}; TNA-PFN = \textbf{0.467}.
  \end{itemize}
    We note that although 50\% of neurons are anchored, theoretically, the final \emph{ranks} of \emph{all} neurons can still change if weights change dramatically. Anchored weights are fixed in value, but non-anchored weights can move above or below them; when ranks are recomputed globally, anchored neurons may shift in the ranking as others cross them. Hence, in principle, up to 100\% of positions can change. However, in Trial 1, TNA-PFN can fix 76.7\% weights without changing their rankings, whereas vanilla training causes 73.3\% neuron ranking changes, showing TNA-PFN's effectiveness in breaking permutation symmetries.

\item \textbf{Findings from the visualizations.}  
  \begin{itemize}
  \item \emph{Bar plots (Figure~\ref{fig:vis_perm_sym_1}):} vanilla training produces substantial reshuffling of weights; the initially smooth monotone profile breaks into a jagged pattern. With TNA-PFN, the final profile remains much closer to the initial ordering. The effect persists even for non-anchored neurons, indicating that anchors act as \emph{references} that stabilize the entire layer’s evolution. The stabilizing effect is stronger in Trial~1 (lower noise) but remains clear in Trial~2 (higher noise). Moreover, when comparing Trial~1 and Trial~2 side by side, we observe that the final distributions under vanilla training differ significantly across the two conditions, reflecting instability under noise and learning-rate changes. In contrast, TNA-PFN produces far more consistent weight profiles across trials, and even the unfrozen neurons show limited variation. This stability across conditions indicates that permutation symmetry is effectively broken by TNA-PFN.
  \item \emph{Scatter plots (Figure~\ref{fig:vis_perm_sym_2}):} vanilla training shows broad spread away from the diagonal (large deviations from the initial ordering). Under TNA-PFN, points concentrate near the diagonal; anchored neurons sit exactly on vertical lines at their initial values, and non-anchored neurons also exhibit smaller deviations, showing that TNA-PFN reduces permutation freedom \emph{without} freezing all degrees of freedom.
  \end{itemize}

\item \textbf{Interpretation.} The one-hidden-layer network is permutation-symmetric: permuting hidden units leaves the function unchanged. Vanilla SGD explores this symmetry freely; stochasticity (data order, noise, learning rate) makes units drift and swap ranks, leading to high ranking-change fractions and run-to-run misalignment. TNA-PFN partially breaks this symmetry by anchoring a subset of neurons. These anchors regularize gradients around stable reference directions, reducing index drift for both anchored and non-anchored neurons. In other words, TNA-PFN induces \emph{permutation asymmetry} during training—solutions become more index-stable and hence easier to fuse.

\item \textbf{Relation to our main claims.} The neuron-level analysis aligns with our barrier and fusion results: when units remain better aligned across runs, the loss/perplexity barriers drop and model fusion improves. Importantly, our constraint is \emph{mild}: unlike heavy pruning or strict PEFT, we do not remove most parameters or confine learning to a tiny subspace; we only fix a fraction of neurons to create stable landmarks. This is enough to reduce permutation symmetry, acting as a mild regularization that improves stability and sometimes generalization.
\end{itemize}

\begin{figure}
    \centering
    \includegraphics[width=0.8\linewidth]{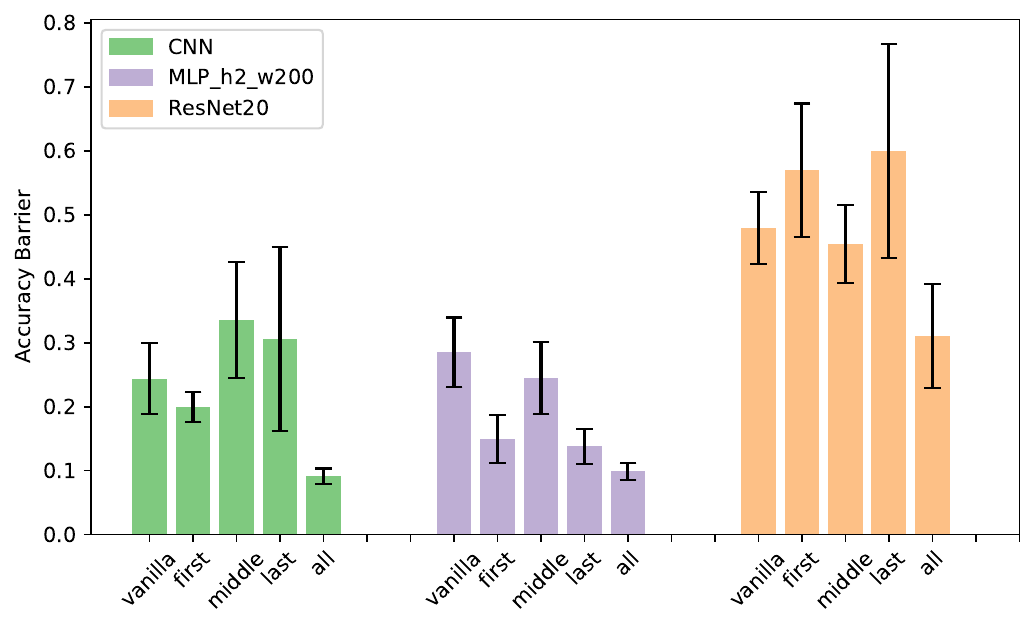}
    \caption{\textbf{Layer-wise analysis of TNA-PFN.} The dataset is CIFAR-10. ``vanilla'' refers to vanilla training. ``first''/``middle''/``last'' refers to only applying TNA-PFN to the first/middle/last layer. ``all'' refers to applying TNA-PFN to all layers (vanilla TNA-PFN). For CNN, the first layer is the convolution layer Conv2d(3, 32, 3), the middle layer is the convolution layer Conv2d(64, 64, 3), and the last layer is the fully connected layer Linear(64, 10) for classification; 
    for MLP\_h2\_w200, the first layer is the fully connected layer Linear(32*32*3, 200), the middle layer is the fully connected layer Linear(200, 200), and the last layer is the fully connected layer Linear(200, 10) for classification; 
    for ResNet20, the first layer is the convolution layer Conv2d(3, 16, kernel\_size=3, stride=1, padding=1, bias=False), the middle layer is the middle block, and the last layer is the fully connected layer Linear(64*block.expansion, 10) for classification; }
    \label{fig:lmc:layer_wise_analysis}
\end{figure}

\begin{figure}
    \centering
\includegraphics[width=0.45\linewidth]{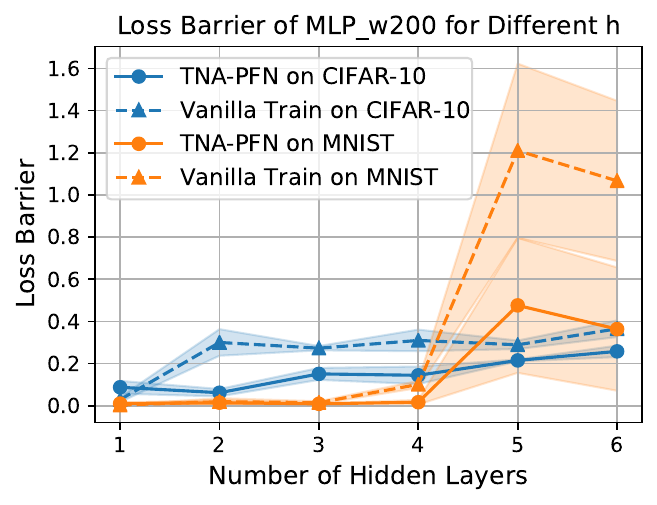}
\includegraphics[width=0.45\linewidth]{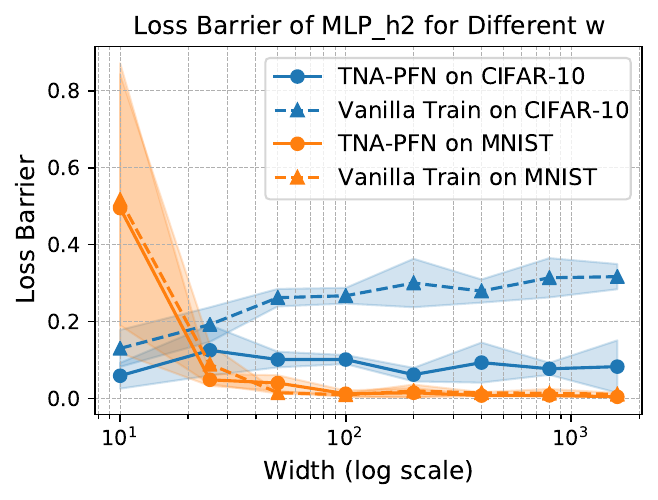}
\includegraphics[width=0.45\linewidth]{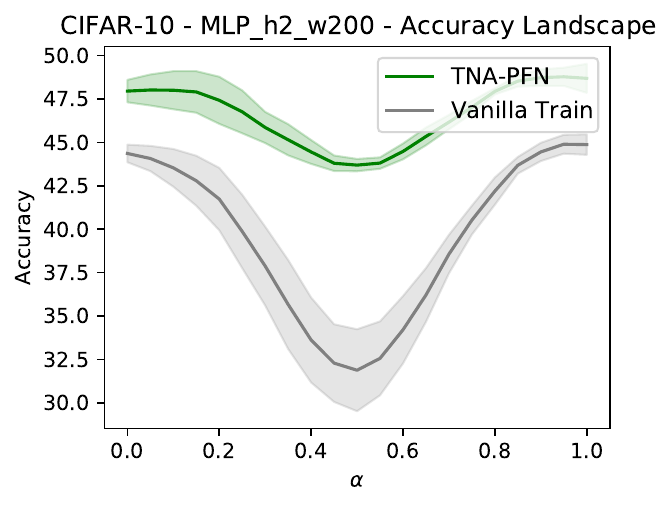}
\includegraphics[width=0.45\linewidth]{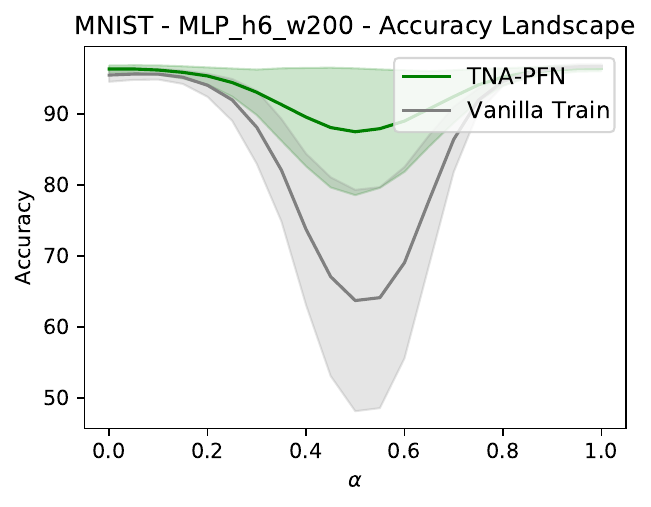}
\includegraphics[width=0.45\linewidth]{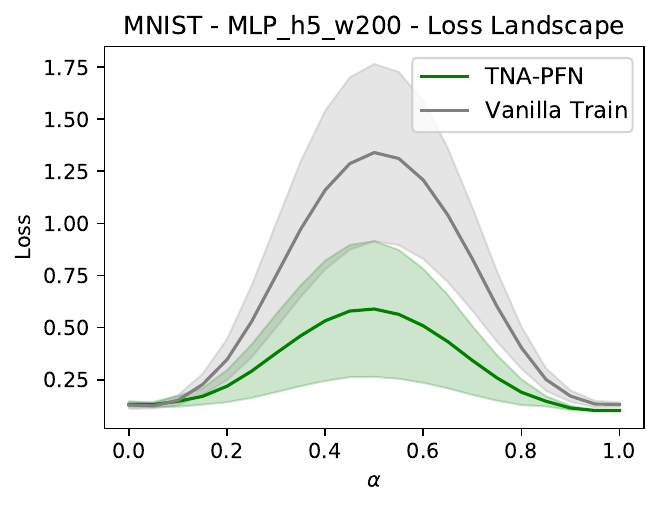}
\includegraphics[width=0.45\linewidth]{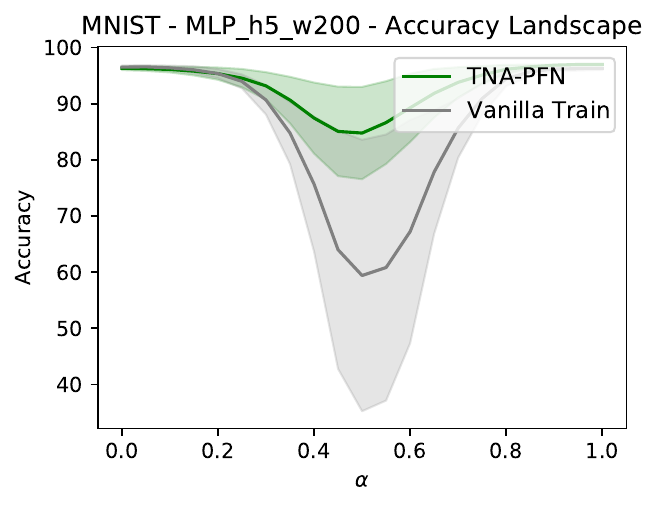}
    \caption{\textbf{Upper two: Loss barriers of MLP under different hidden layers ($h$) and widths ($w$). Middle two and Lower two: Accuracy and loss landscapes of MLPs.}}
    \label{fig:lmc:mlp_barrier_landscape2}
\end{figure}

\begin{figure}
    \centering
\includegraphics[width=0.45\linewidth]{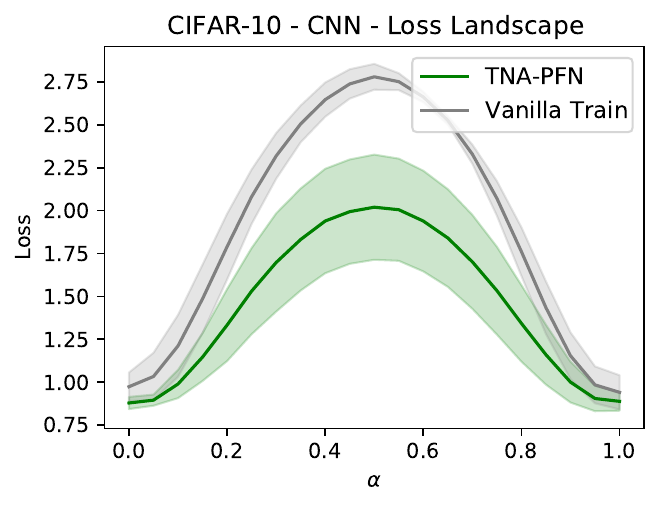}
\includegraphics[width=0.45\linewidth]{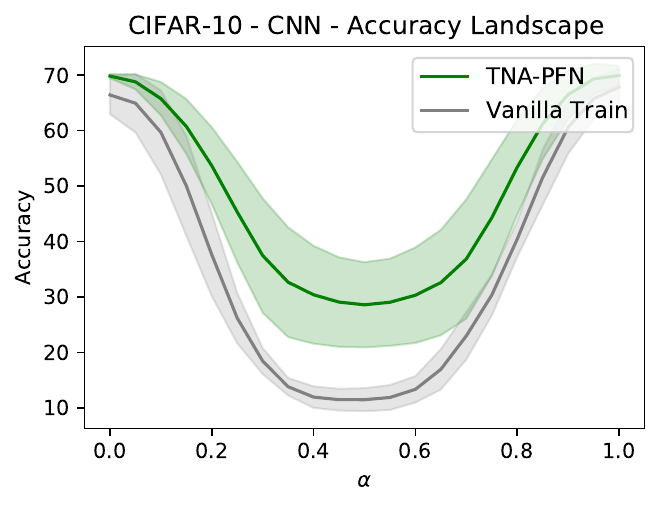}
\includegraphics[width=0.45\linewidth]{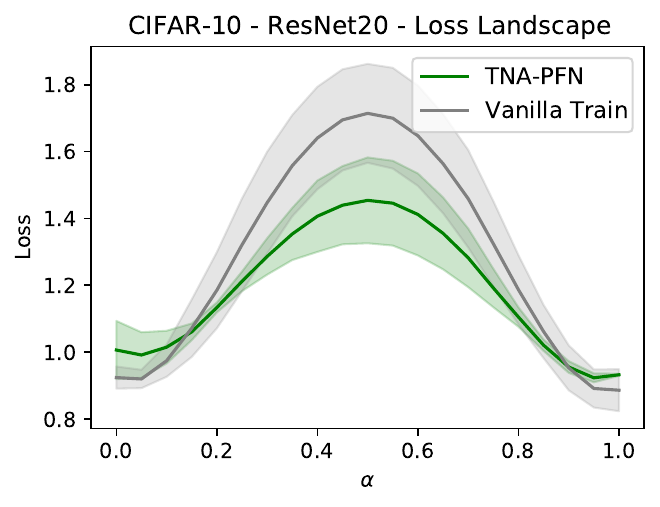}
\includegraphics[width=0.45\linewidth]{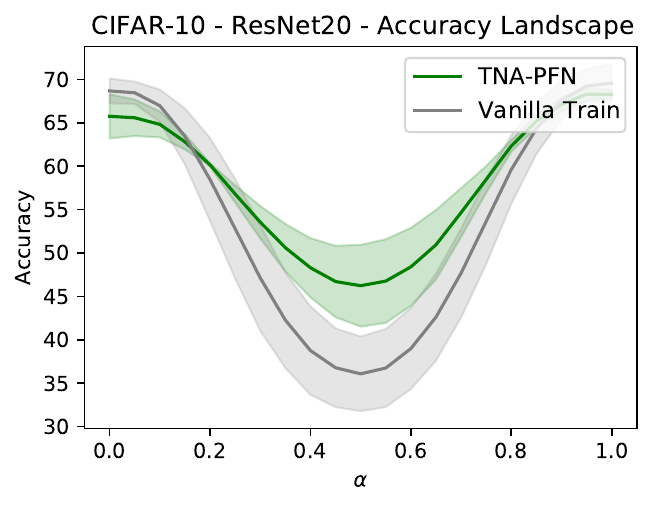}
    \caption{\textbf{Upper two: Loss and accuracy landscapes of CNN. Lower two: Loss and accuracy landscapes of ResNet20.}}
    \label{fig:lmc:model_archs_landscape}
\end{figure}

\begin{figure}
    \centering
\includegraphics[width=0.35\linewidth]{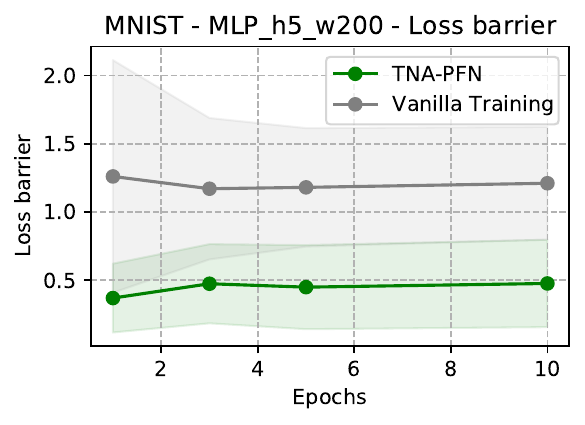}
\includegraphics[width=0.35\linewidth]{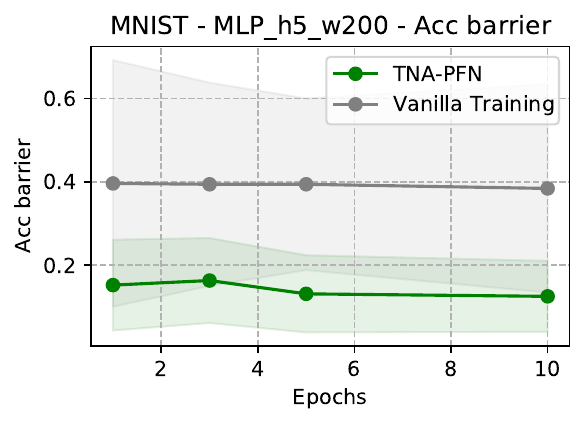}
\includegraphics[width=0.35\linewidth]{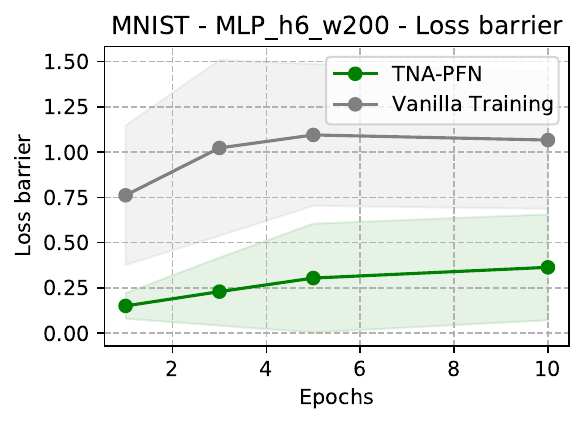}
\includegraphics[width=0.35\linewidth]{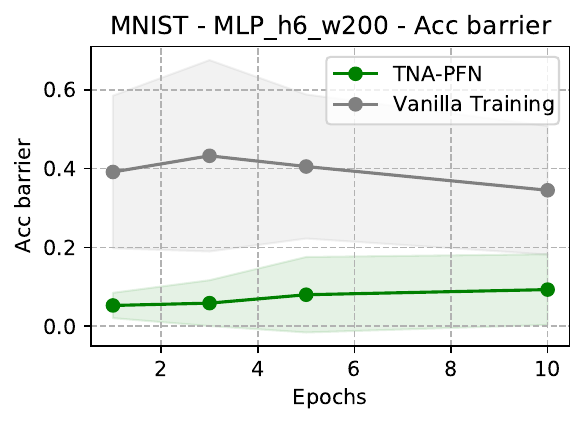}

\includegraphics[width=0.35\linewidth]{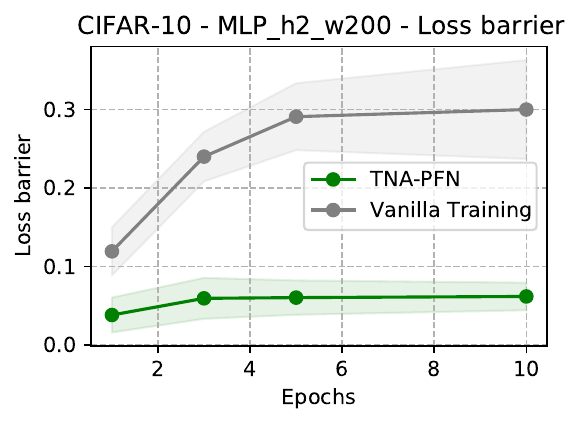}
\includegraphics[width=0.35\linewidth]{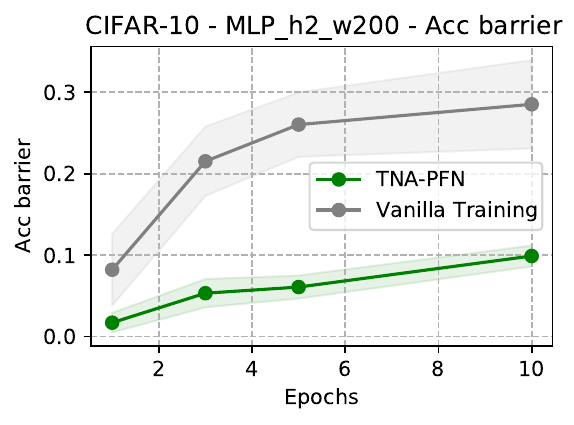}
\includegraphics[width=0.35\linewidth]{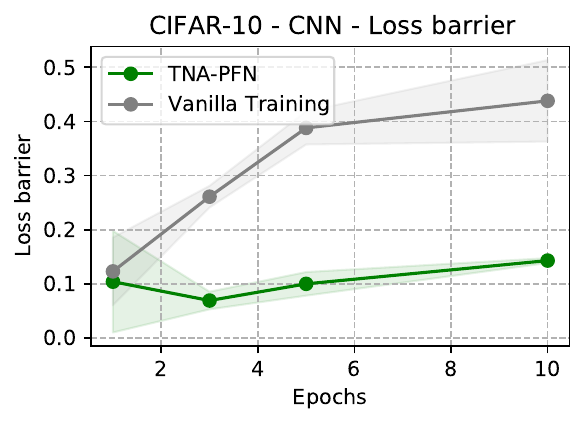}
\includegraphics[width=0.35\linewidth]{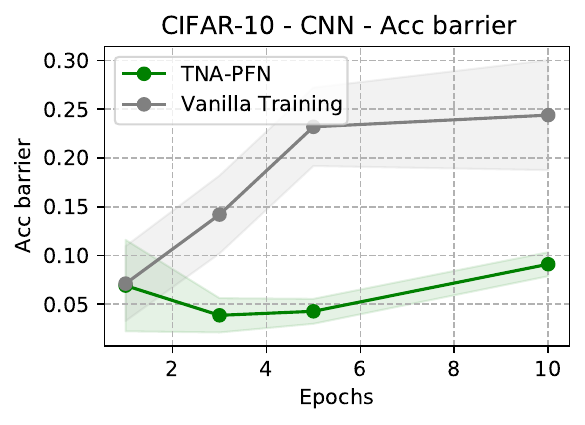}
\includegraphics[width=0.35\linewidth]{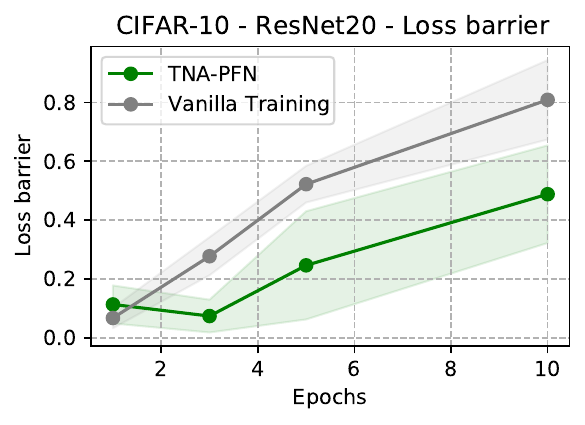}
\includegraphics[width=0.35\linewidth]{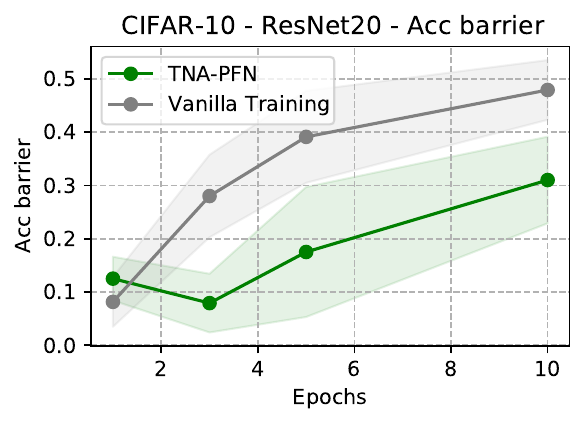}
\vspace{-0.3cm}
    \caption{\textbf{Barrier changes during training for different datasets and models.}}
    \label{fig:lmc:barrier_wrt_epoch}
\end{figure}

\begin{figure}
  \centering
    \includegraphics[width=0.5\linewidth]{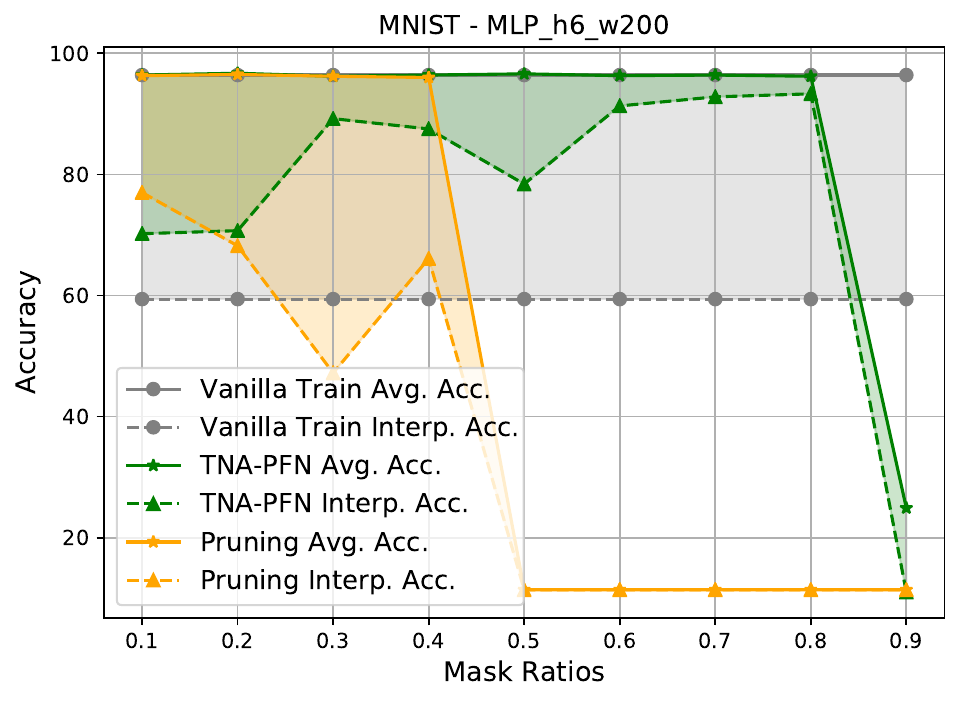}
    \includegraphics[width=0.5\linewidth]{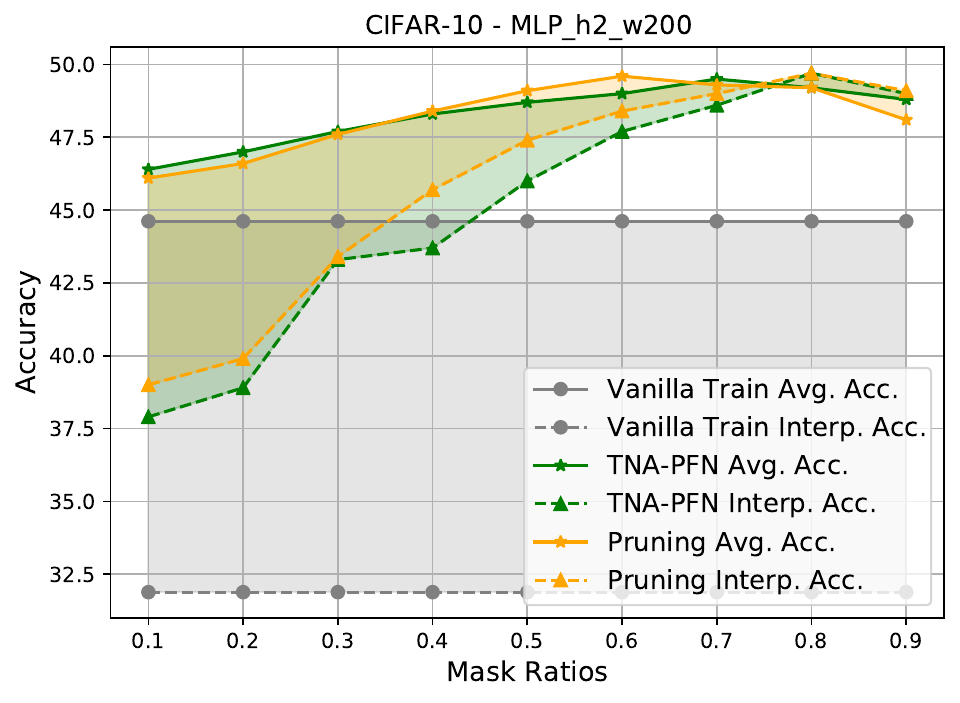}
    \includegraphics[width=0.5\linewidth]{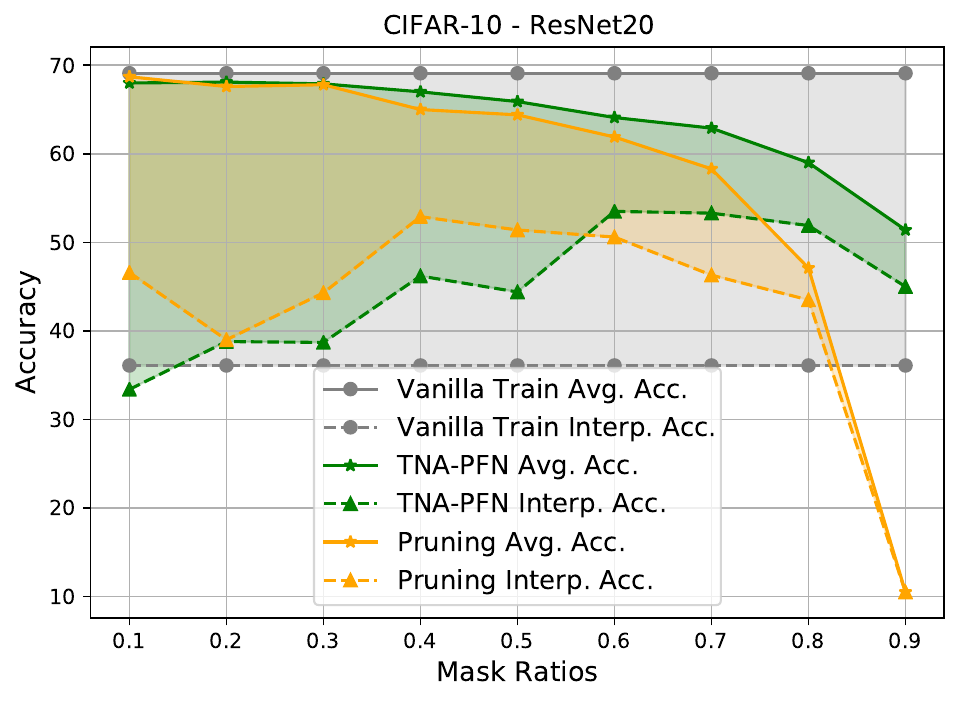}
    \caption{\textbf{More results about pruning and TNA-PFN under different mask ratios.} The shadow areas mean the accuracy barriers.}
    \label{fig:lmc:pruning_mask_ratios2}
\end{figure}

\begin{table}[h]
\centering
    \caption{\textbf{Linear mode connectivity on    Tiny ImageNet.} The $\rho$ for CNN is 0.4 and the $\rho$ for ResNet18 is 0.3. The learning rate is 0.08.}
    \vspace{-0.17cm} 
\resizebox{0.55\linewidth}{!}{
    \begin{tabular}{@{}l|c|c|c}
    \toprule
    \textbf{Models} &\textbf{Metrics} &\textbf{Vanilla Train} &\textbf{TNA-PFN} \\
    \midrule
    \multirow{4}{*}{CNN} 
    &Avg. Acc. &9.85±0.3 &11.4±0.6 \\
    &Interp. Acc. &1.4±0.2 &2.91±0.9 \\
    \cmidrule{2-4}
    &Acc. Barrier &0.86± 0.03 &0.75±0.07\ (\textcolor{red}{12.8\%$\downarrow$}) \\
    &Loss Barrier &0.84±0.08 &0.75±0.09 \ (\textcolor{red}{10.4\%$\downarrow$}) \\
    \midrule
    \multirow{4}{*}{ResNet20} 
    &Avg. Acc. &31.8±0.3 &31.6±0.4 \\
    &Interp. Acc. &6.86±1.8 &12.5±2.1 \\
    \cmidrule{2-4}
    &Acc. Barrier &0.78±0.06 &0.60±0.07\ (\textcolor{red}{23\%$\downarrow$}) \\
    &Loss Barrier &1.6±0.2 &1.2±0.09\ (\textcolor{red}{22.2\%$\downarrow$}) \\
    \bottomrule
    \end{tabular}}
    \label{tab:lmc:tinyimagenet}
\end{table}

\textbf{Results on large-scale dataset.} We conduct experiments on Tiny ImageNet~\cite{tinyimagenet}, a subset of ImageNet~\cite{deng2009imagenet}, containing 100000 images of 200 classes (500 for each class) downsized to 64×64 colored images. Each class has 500 training images, 50 validation images, and 50 test images. The result is shown in \autoref{tab:lmc:tinyimagenet} of this appendix. It can be seen that under large-scale datasets, TNA-PFN also can reduce the barriers in the linear mode connectivity.

\textbf{Layer-wise analysis.} We conduct a layer-wise analysis of TNA-PFN to see which layer matters most in improving LMC in \autoref{fig:lmc:layer_wise_analysis} of this appendix, and different model architecture poses different results. For simple CNN, only applying neuron fixing in the first layer (convolution) will improve LMC, and partially fixing weights in the middle (convolution) and the last (fully connected) layers will cause barrier increases. For MLP\_h2\_w200, we observe that independently fixing one layer will all cause barrier reductions, and the performance is more dominant when fixing the first and the last layers; jointly fixing all layers (``all'') will have the lowest barrier. For ResNet20, it is revealed that only fixing the middle layers (the middle block) will cause barrier degradation. 

\textbf{Extensions of Figure 4.} We provide more illustrations about the loss and accuracy barriers and landscapes in \autoref{fig:lmc:mlp_barrier_landscape2} of this appendix. 


\begin{figure}
\centering
    \includegraphics[width=0.5\linewidth]{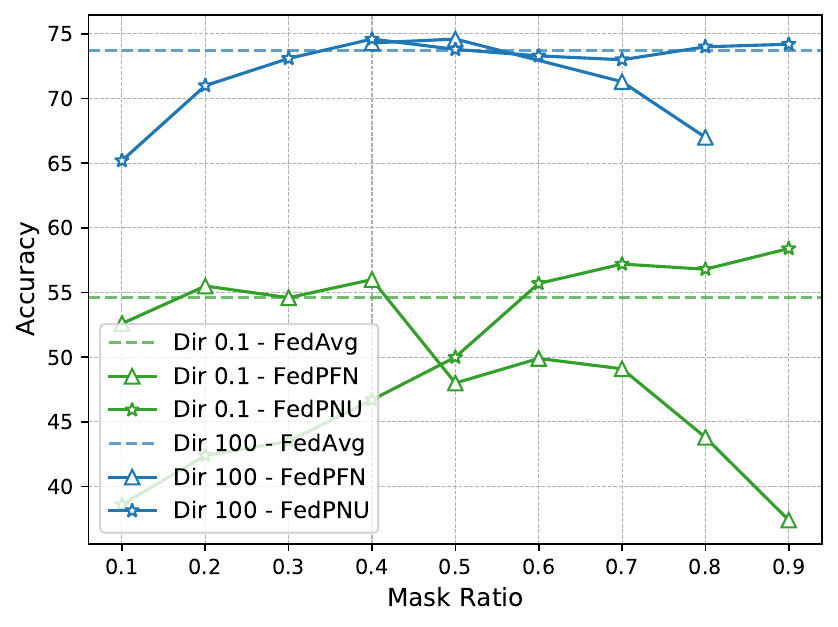}
    \caption{\textbf{Performances of FedPFN and FedPNU under different mask ratios.} CIFAR-10, ResNet20, and $E=3$.}
    \label{fig:fl:mask_ratios2}
\end{figure}

\textbf{Extensions of Figure 5.} We provide illustrations about the loss and accuracy barriers of the Figure 5 results in \autoref{fig:lmc:model_archs_landscape} of this appendix. It is obvious that TNA-PFN can lower the barriers in LMC.

\textbf{Loss and accuracy barriers w.r.t. epochs.} We demonstrate the barrier changes during training in \autoref{fig:lmc:barrier_wrt_epoch} of this appendix. It is shown that barriers increase during training, revealing that two independent networks diverge in parameter space. TNA-PFN has slower barrier-increasing rates than vanilla training.

\textbf{Extensions of Figure 2.} We provide more results about pruning and TNA-PFN under different mask ratios in \autoref{fig:lmc:pruning_mask_ratios2} of this appendix. Interestingly, for CNN, pruning and TNA-PFN improve both the accuracy and connectivity and the improvements go up along with the ratio increasing. On the other side, we observe an obvious accuracy-connectivity tradeoff for ResNet20 and it is more severe for pruning. Also, considering the layer-wise evaluation for ResNet in \autoref{fig:lmc:layer_wise_analysis} of this appendix, we reckon it is important to devise different mask strategies for the layers in ResNet and other deeper or more complex models.

\subsection{More Results and Illustrations in Federated Learning}\label{app_subsec:fl}

\textbf{Different local epochs.} We verify the TNA variants in FL under different local epochs in Figure 8. We find that the improvements are also strong when there are more local updates. It is observed that FedPNU is more robust regarding local epochs, and this is because it learns in the complementary subspaces progressively, reducing the negative effects of subspaces on accuracy. Similar reasons are also for why FedPNU is robust when the mask ratio is as high as 0.9 in Figure 9.

\textbf{The effects of mask ratios for FedPFN and FedPNU.} From Figure 9, it is shown that FedPFN benefits under smaller subspaces (higher mask ratios) but falls short when the subspace is too small (ratio $\rho=0.9$); whereas FedPNU is robust across all mask ratios due to its progressive learning.

\textbf{Extensions of Figure 9.} We show the performances of FedPFN and FedPNU under different mask ratios for ResNet20 in \autoref{fig:fl:mask_ratios2} of this appendix. The results indicate that FedPFN is sensitive to the mask ratio while FedPNU is more robust. FedPNU reaches higher performances under higher mask ratios ($\rho \in [0.8, 0.9]$).

\begin{figure}
  \centering
    \includegraphics[width=0.47\linewidth]{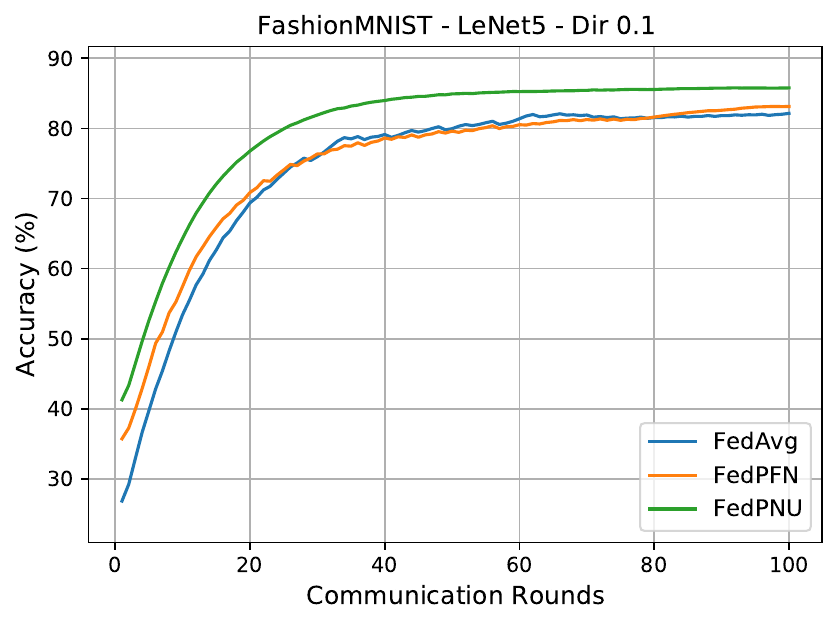}
    \includegraphics[width=0.47\linewidth]{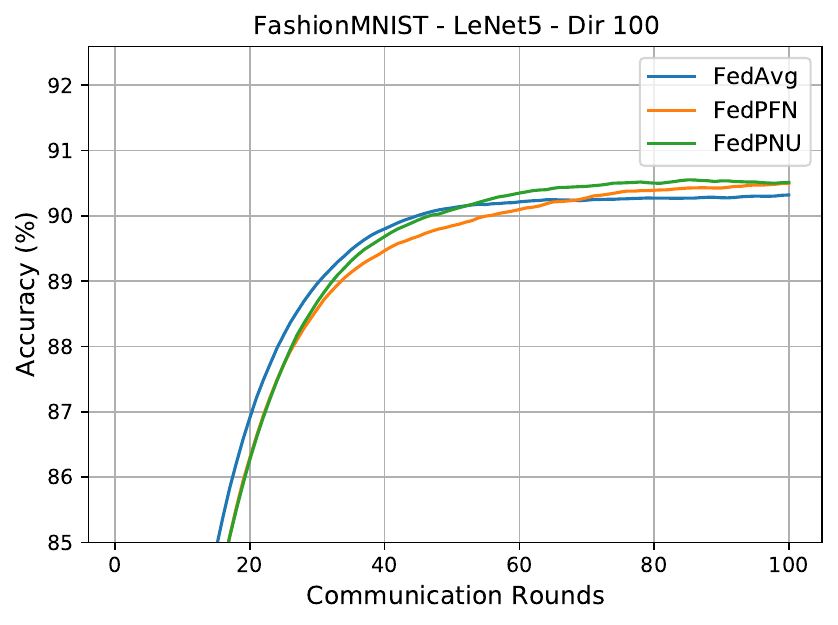}
    \includegraphics[width=0.47\linewidth]{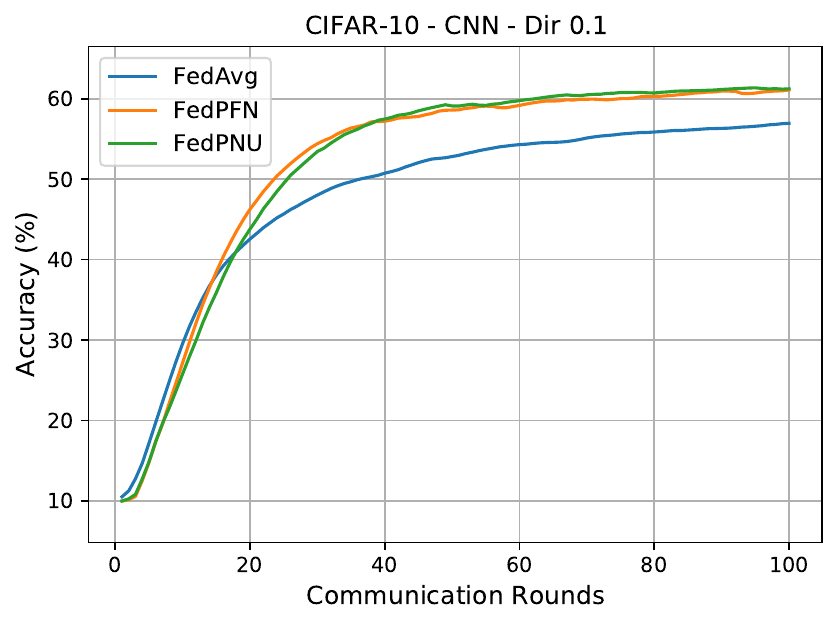}
    \includegraphics[width=0.47\linewidth]{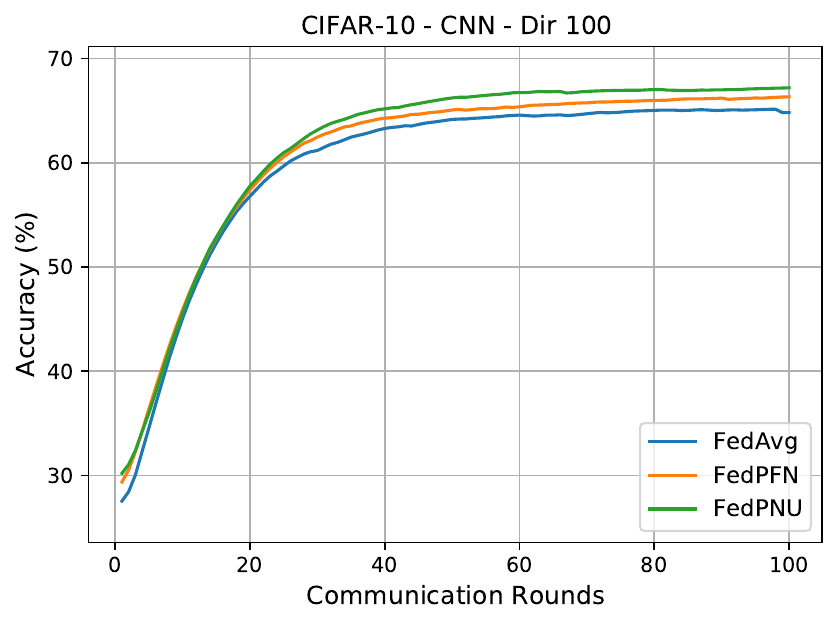}
    \includegraphics[width=0.47\linewidth]{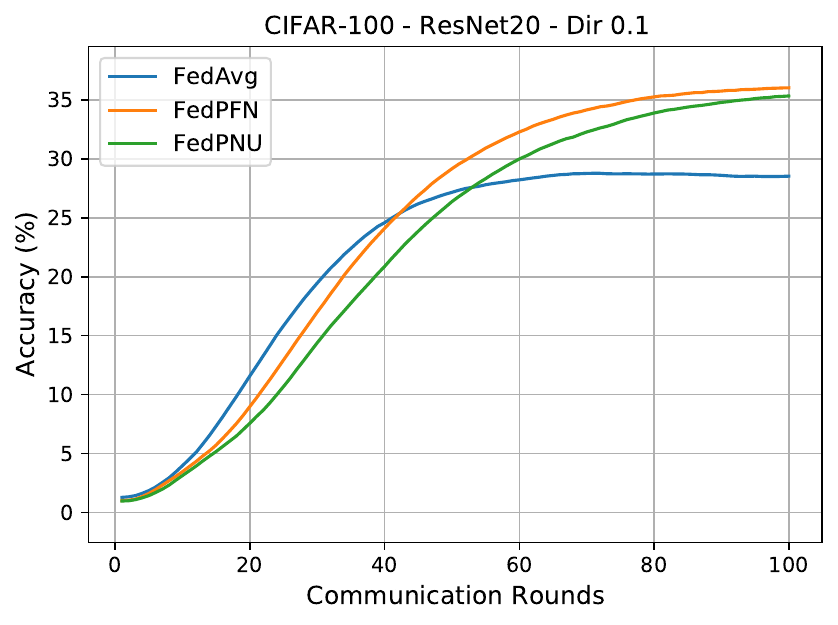}
    \includegraphics[width=0.47\linewidth]{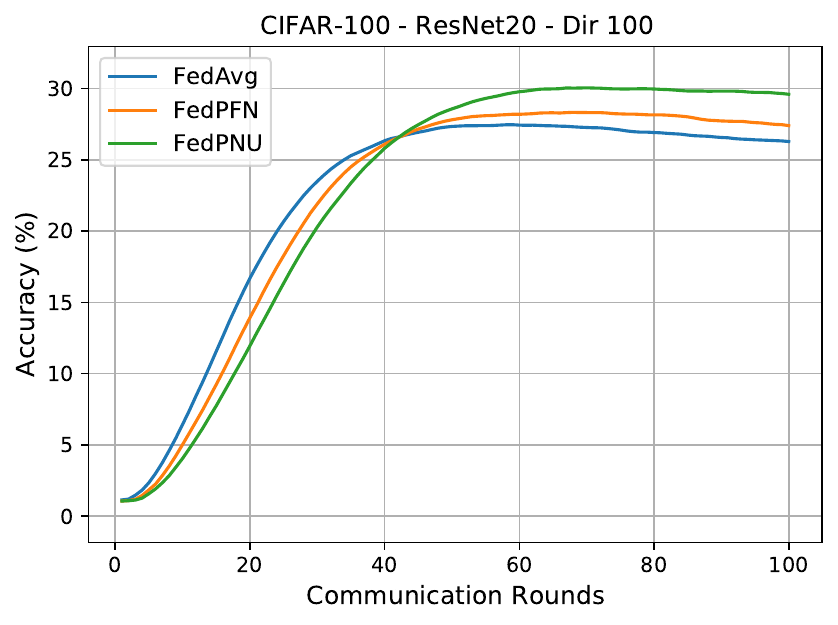}
    \caption{\textbf{Test accuracy curves of the federated learning methods.}}
    \label{fig:fl:acc_curves}
\end{figure}

\textbf{Illustrations of the learning curves.} We present the test accuracy curves of FedAvg, FedPFN, and FedPNU in \autoref{fig:fl:acc_curves} of this appendix. Our methods show dominant advantages over FedAvg in both IID and non-IID settings, especially for the more complex datasets, CIFAR-10 and CIFAR-100.

\subsection{Model soup analysis}\label{app_subsec:model_soup}

We show the model soup results of pruning and TNA-PFN under different mask ratios in Figure 6. Interestingly, for TNA-PFN, we find when the mask ratios increase, the individual accuracies also increase, while the greedy soup accuracy may drop under a large $\rho$. For pruning, the generalization is terrible since pruning destroys the intrinsic network representations of the pretrained models.

\end{document}

%% file: math_commands.tex
\usepackage{amsmath,amsfonts,bm}

\def\eqref#1{equation~\ref{#1}}

\def\1{\bm{1}}

\def\va{{\bm{a}}}

\def\vv{{\bm{v}}}

\def\vx{{\bm{x}}}
\def\vy{{\bm{y}}}

\def\evv{{v}}

\def\evx{{x}}

\def\mM{{\bm{M}}}

\def\mU{{\bm{U}}}
\def\mV{{\bm{V}}}

\DeclareMathAlphabet{\mathsfit}{\encodingdefault}{\sfdefault}{m}{sl}
\SetMathAlphabet{\mathsfit}{bold}{\encodingdefault}{\sfdefault}{bx}{n}

\def\sR{{\mathbb{R}}}

\def\sX{{\mathbb{X}}}

\def\emM{{M}}

\def\emU{{U}}

\newcommand{\vecx}{\mathbf{x}}
